\documentclass[letterpaper]{article} 
\usepackage{aaai24}  
\usepackage{times}  
\usepackage{helvet}  
\usepackage{courier}  
\usepackage[hyphens]{url}  
\usepackage{graphicx} 
\usepackage{natbib}  
\usepackage{caption} 
\usepackage{algorithm}
\usepackage{algpseudocode}
\usepackage{amssymb}
\usepackage{amsmath}
\usepackage{amsthm}
\usepackage{graphicx}
\usepackage{subcaption}
\usepackage{enumitem}
\usepackage{xcolor}
\usepackage{soul}

\newtheorem{axiom}{Axiom}
\theoremstyle{plain}
\newtheorem{definition}{Definition}
\theoremstyle{plain}
\newtheorem{theorem}{Theorem}
\theoremstyle{plain}
\newcommand{\dmax}{D_n}
\newcommand{\phiopt}{\phi^n}
\newcommand{\squishlisttwo}{
    \begin{list}{$\bullet$}{ 
        \setlength{\itemsep}{1pt}
        \setlength{\parsep}{0pt}
        \setlength{\topsep}{0pt}
        \setlength{\partopsep}{0pt}
        \setlength{\leftmargin}{1em}
        \setlength{\labelwidth}{1.5em}
        \setlength{\labelsep}{0.5em}}}
\newcommand{\squishend}{
    \end{list}}
\newcommand{\indentsquishlisttwo}{
    \begin{list}{$\bullet$}{ 
        \setlength{\itemsep}{1pt}
        \setlength{\parsep}{0pt}
        \setlength{\topsep}{0pt}
        \setlength{\partopsep}{0pt}
        \setlength{\leftmargin}{2em}
        \setlength{\labelwidth}{1.5em}
        \setlength{\labelsep}{0.5em}}}
\newcommand{\indentsquishend}{
    \end{list}}

\title{DeRDaVa: Deletion-Robust Data Valuation for Machine Learning}
\author {
    Xiao Tian\textsuperscript{\rm 1,\rm 2},
    Rachael Hwee Ling Sim\textsuperscript{\rm 1},
    Jue Fan\textsuperscript{\rm 1,\rm 2},
    Bryan Kian Hsiang Low\textsuperscript{\rm 1}
}
\affiliations {
    \textsuperscript{\rm 1} Department of Computer Science, National University of Singapore \\
    \textsuperscript{\rm 2} Department of Mathematics, National University of Singapore \\
    \{xiao.tian, rachael.sim, jue.fan\}@u.nus.edu, lowkh@comp.nus.edu.sg
}

\begin{document}

\maketitle

\begin{abstract}
Data valuation is concerned with determining a fair valuation of data from data sources to compensate them or to identify training examples that are the most or least useful for predictions. With the rising interest in personal data ownership and data protection regulations, model owners will likely have to fulfil more data deletion requests. This raises issues that have not been addressed by existing works: \textit{Are the data valuation scores still fair with deletions? Must the scores be expensively recomputed?} The answer is no. To avoid recomputations, we propose using our data valuation framework DeRDaVa upfront for valuing each data source's contribution to preserving robust model performance after anticipated data deletions. DeRDaVa can be efficiently approximated and will assign higher values to data that are more useful or less likely to be deleted. We further generalize DeRDaVa to Risk-DeRDaVa to cater to risk-averse/seeking model owners who are concerned with the worst/best-cases model utility. We also empirically demonstrate the practicality of our solutions.
\end{abstract}

\section{Introduction}\label{sec:introduction}

Data is essential to building machine learning (ML) models with high predictive performance and model utility. Model owners source for data directly from their customers or from collaborators in collaborative machine learning \citep{nguyen2022federated}. For example, a credit card company can train an accurate ML model to predict the probability of default based on consumers' income and payment history data \citep{tsai2010credit}. As another example, a healthcare firm can train an ML model to predict the progression of diabetes based on patients' health data \citep{tcfod2019sharing}. As the quality of data contributed by multiple data sources may vary widely, several works \citep{fan2022improving, tay2022incentivizing, xu2021gradient} have recognized the importance of \textit{data valuation} to help model owners compensate data sources fairly, or identify data that are most or least useful for predictions. 

\begin{figure}
    \centering
    \includegraphics[width=\linewidth]{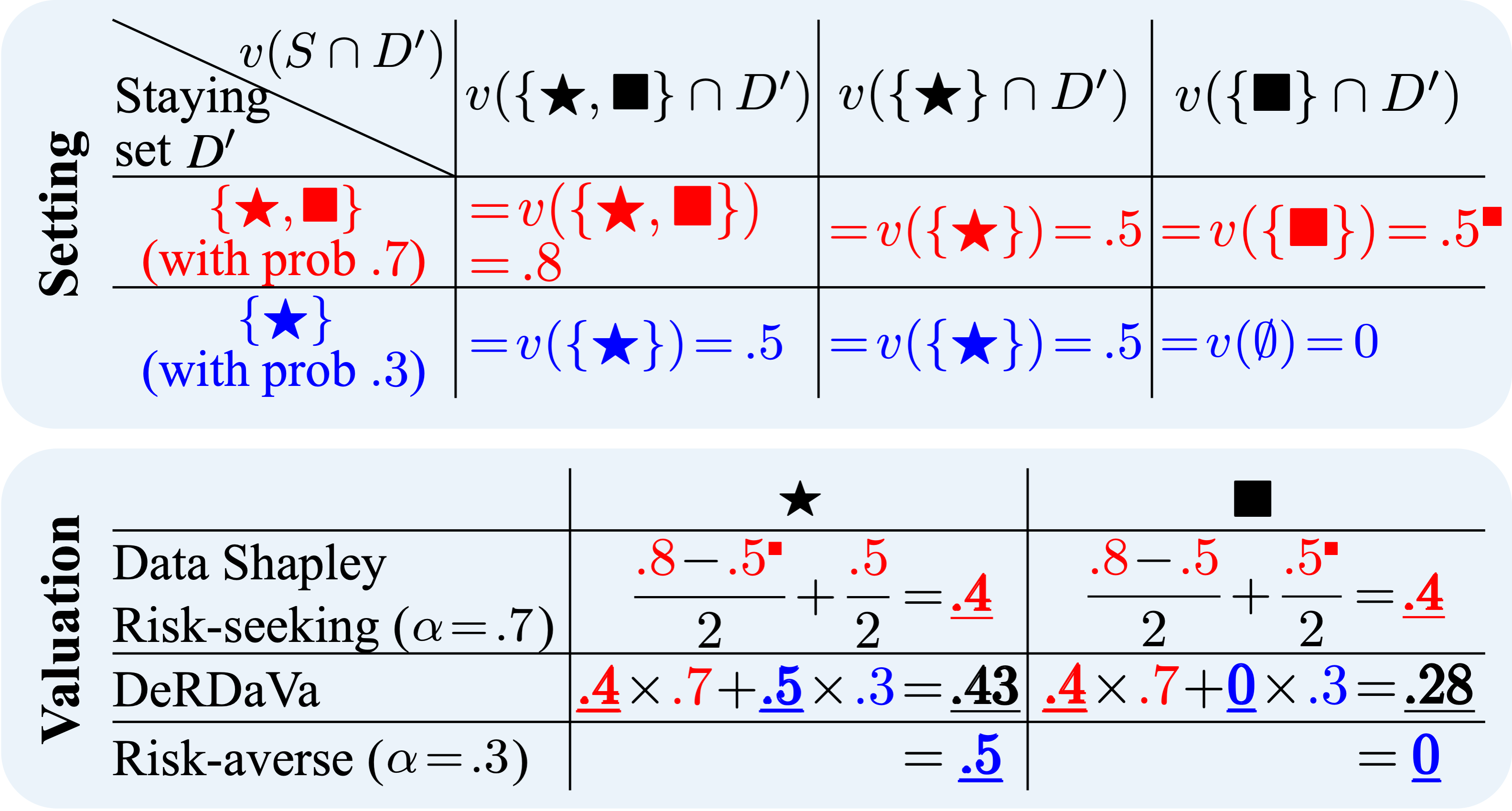}
        \caption{Comparison of Data Shapley vs. the deletion-robustified DeRDaVa and Risk-DeRDaVa scores with Shapley prior for a game with $2$ \emph{interchangeable} sources. $\bigstar$ and $\blacksquare$ stay with probability $1$ and $.7$ respectively. 
        $\bigstar$ and $\blacksquare$ have equal Data Shapley score but $\bigstar$ has higher DeRDaVa and Risk-averse DeRDaVa scores. This is because Data Shapley (Eq.~(\ref{eqn:semivalue})) considers only the initial support set $\{\bigstar, \blacksquare\}$ while DeRDaVa (Eq.~(\ref{eqn:urdava})) and Risk-averse DeRDaVa (Eq.~(\ref{eqn:risk-urdava})) also consider the worst-case support set $\{\bigstar\}$. Further explanation is included in App.~\ref{appendix:numerical-comparison}.}
    \label{fig:example}
\end{figure}

\textit{Data valuation} studies how much data is worth and proposes how rewards associated with the ML model can be \textbf{fairly} allocated to each data source \cite{sim2022survey}. Several existing data valuation techniques \citep{ghorbani2019data, jia2019towards, yang2019federated} have adopted the use of \textit{semivalues} from cooperative game theory. Recent works \citep{sim2020collaborative, tay2022incentivizing, zhang2021incentive} have also developed various reward allocation schemes based on semivalues. The core intuition behind semivalues is that a data source should be fairly valued relative to other data sources (i.e., based on the averaged model utility improvement it contributes to each sub-coalition of other data sources). 
Moreover, these works have justified the use of semivalues by fairness axioms such as \textit{Interchangeability} --- assigning the same valuation score to two data sources $d_i$ and $d_j$ with the same model utility improvement to every sub-coalition (e.g., $\bigstar$ and $\blacksquare$ in Fig.~\ref{fig:example}).

\textit{Data deletion} refers to the deletion of \textbf{data and their impact} from trained ML models. Due to the recent introduction of data protection legislation, such as General Data Protection Regulation (GDPR) in the European Union and California Consumer Privacy Act (CCPA) in the United States, data deletions are expected to occur more frequently. These laws legally assert that data are properties of their owners and data owners have the \textit{right to be forgotten} \citep{shastri2019seven}. This mandates model owners to \textbf{delete the data and their impact} from trained models \textit{without undue delay} \citep{magdziarczyk2019right} upon request or after a stipulated time \citep{ong2018data}. While these regulations have led to intensive research on \textit{machine unlearning} \citep{bourtoule2021machine, chen2021machine, sekhari2021remember} to efficiently remove the impact of deleted data from trained models, to our knowledge, no work has considered the impact of data deletions on data valuation.

It is our view that data deletions challenge existing semivalue-based data valuation techniques. The existing (pre-deletion) semivalue may not preserve the fairness axioms after data deletions. For example, if data sources $d_i$ and $d_j$ only make the same model utility improvement to every sub-coalitions after deletions, they may have been assigned different pre-deletion semivalues (which violates the Interchangeability axiom) (see App. \ref{appendix:semivalue-axioms}). At first glance, one may think of recomputing the semivalue after every deletion to address the challenge. However, the recomputations are computationally expensive and data owners (and legislators) may not tolerate the uncertainty and fluctuations in valuation (e.g., having to return monetary compensation). Thus, we propose new \textit{deletion-robustified fairness axioms} and a more proactive approach that anticipates and accounts for data deletions: the model owner should 
assign higher \textbf{deletion-robust data valuation (DeRDaVa)} scores upfront to data sources with higher probability of staying (and thus contribute to \emph{deletion-robustness}, i.e., robustly preserving model performance after deletions).
We show that DeRDaVa
\textbf{satisfies our deletion-robustified fairness axioms} (Sec.~\ref{subsec:random-cooperative-game-and-robustified-fairness-axioms}) and can be efficiently approximated (Sec.~\ref{subsec:urdava-and-its-efficient-approximation}). Lastly, to cater to model owners who are only concerned with the expectation of \textbf{worst-cases model utility} (i.e., a \textit{risk-averse} model owner) or best-cases model utility (i.e., a \textit{risk-seeking} model owner), we generalize DeRDaVa to \textbf{Risk-DeRDaVa} (Sec.~\ref{subsec:risk}). In Sec.~\ref{sec:experiments}, we empirically demonstrate and compare the behaviour of DeRDaVa with semivalues after data deletions on real-world datasets.

\section{Background and Related Works}\label{sec:background-and-related-works}

\subsection{Semivalue-Based Data Valuation} \label{subsec:semivalue-based-data-valuation}

Machine learning can be viewed as a \textit{cooperative game} among multiple data sources in order to gain the highest model performance, where each data source can be a single data point or a smaller dataset. Semivalue \citep{dubey1981value} is a concept from cooperative game theory (CGT) that measures the contribution of each data source in such a cooperative game. To formalize this, we define such a cooperative game as a pair $\langle D, v \rangle$, where the \textit{support set} $\textstyle D = \bigcup_{i = 1}^n \{d_i\}$ represents the set of $n$ data sources $d_i$ indexed by $i$ in the collaboration and the \textit{model utility function} $v: \mathcal{P}(D) \to \mathbb{R}$ maps each coalition of data sources in the power set of $D$ to its utility. Specifically, the utility $v(S)$ of a coalition $S$ of data sources can represent the prediction performance (e.g., validation accuracy) of model trained with data from $S$. For example, $v(\{d_0, d_1, d_2\}) = 0.9$ may represent that we use data from the three data sources $d_0$, $d_1$ and $d_2$ to train a model and obtain an accuracy of $0.9$ when evaluated on a validation set.

Let $G^n$ represent the set of all cooperative games with respective support set $D$ sized $n$. To measure the contribution of each data source, we want to find an $n$-sources \textit{data valuation function} $\phi^n: G^n \to \mathbb{R}^n$ that assigns each data source $d_i$ a real-valued valuation score $\phi_i^n(\langle D, v \rangle)$ abbreviated as $\phi_i^n(v)$. To ensure the \textbf{fairness} of data valuation functions, a common approach in CGT is \textit{axiomatization}, where a list of axioms is provided to be fulfilled by data valuation functions. There are four important axioms that are commonly agreed to be fair \citep{covert2021explaining, ridaoui2018axiomatisation, sim2022survey}: \textbf{Linearity}, \textbf{Dummy Player}, \textbf{Interchangeability} and \textbf{Monotonicity} (refer to App.~\ref{appendix:semivalue-axioms} for their respective definition and connection to fairness). Here, we define \textit{semivalue}, a unique form of data valuation functions that fulfils the four axioms concurrently below\footnote{In Sec.~\ref{subsec:random-cooperative-game-and-robustified-fairness-axioms}, we will explain how the axioms should be robustified with anticipated data deletions.}:
\begin{definition}\label{def:semivalue}
    \textup{[Semivalue \citep{dubey1981value}]} An $n$-sources data valuation function $\phi^n: G^n \to \mathbb{R}^n$ is called a \textbf{semivalue} if the valuation score  $\phi_i^n(v)$ assigned to any data source $d_i \in D$ satisfies
    \begin{equation}\label{eqn:semivalue} \textstyle
         \phi_i^n(v) = \sum\limits_{S \subseteq D \setminus \{d_i\}} w_{|S|} \cdot {\mathrm{MaC}_v(d_i | S)} / {\binom{n - 1}{|S|}},
    \end{equation}
    where $w_{|S|} \geq 0$ is a \textbf{weighting term} associated with all coalitions $S$ of size $s = |S|$, satisfying $\textstyle \sum_{s = 0}^{n - 1} w_s = 1$; $\mathrm{MaC}_v(d_i | S) = v(S \cup \{d_i\}) - v(S)$ is the \textbf{marginal contribution} of data source $d_i$ to coalition $S$ under model utility function $v$, representing the additional model utility brought by $d_i$ to $S$ measured by $v$.
\end{definition}
\noindent
\textit{Interpretation.} 
Semivalues can be interpreted as a weighted sum of marginal contribution of $d_i$ to each coalition $S$. Moreover, since a support set with $n$ data sources has $\binom{n - 1}{s}$ coalitions sized $s$ excluding data source $d_i$, Eq. (\ref{eqn:semivalue}) can be re-interpreted as the expectation of the average marginal contribution of $d_i$ to coalitions sized $0, 1, \cdots, n - 1$ over some categorical distribution $W_\mathbf{s}$, where $W_\mathbf{s}(\mathbf{s} = s) = w_s$. This offers model owners freedom to place more focus on larger or smaller coalitions. For example, Leave-One-Out (LOO) only considers $d_i$'s marginal contribution to the largest possible coalition excluding $d_i$ and 
sets $w_{n-1} = 1$ and other $w_s = 0$. 
Beta Shapley \citep{kwon2021beta} sets $W_\mathbf{s}$ to be a beta-binomial distribution with two parameters $\alpha$ and $\beta$ such that model owners can put more weights on smaller coalitions by setting a larger $\alpha$ and on larger coalitions by setting a larger $\beta$.

\subsection{Data Deletion and Machine Unlearning}\label{subsec:data-deletion-and-machine-unlearning}

Due to new data protection regulations, model owners must delete data sources' data from their datasets and erase their impact from their trained models upon request. 
Machine unlearning works \cite{nguyen2022survey} have studied how to erase data effectively and efficiently and proposed model-agnostic methods (such as decremental learning \cite{ginart2019making} and differential privacy \cite{gupta2021adaptive}), model-intrinsic methods (for linear models \cite{izzo2021approximate} and Bayesian models \cite{nguyen2020variational}), and data-driven methods (such as data partitioning \cite{bourtoule2021machine} and data augmentation \cite{huang2021unlearnable}).
Our work complements machine unlearning: model owners foresee future data source deletions (and changes in model performance), and thus require a new data valuation approach to value a data source based on its contribution to model performance both before and after anticipated data deletions.

In our problem setting, the main challenge is \textbf{how we can adapt and extend the (C1) fairness axioms and (C2) concept of semivalues to cases where data deletion occurs}. Assume that the model owner and data sources in the collaboration decide to use the data valuation function $\phi^n$ for valuation when there is no deletion.
Our solution should be derived from this jointly-agreed semivalue $\phi^n$ and satisfy some deletion-robustified fairness axioms\footnote{App.~\ref{appendix:scaled-semivalue} discusses why DeRDaVa is superior to the simpler alternative of multiplying the pre-deletion semivalue score of each data source $d_i$ with its staying probability.}.

\section{Methodology}
In Sec.~\ref{subsec:random-cooperative-game-and-robustified-fairness-axioms}, we define a \textit{random cooperative game} to model the situation where data sources may be deleted and the corresponding deletion-robustified fairness axioms to address (C1). In Sec.~\ref{subsec:npo}, we describe the \emph{null-player-out} consistency property to extend the jointly-agreed semivalue $\phiopt$ to the sub-games after data deletions to address (C2). In Sec.~\ref{subsec:urdava-and-its-efficient-approximation}, we define our deletion-robust data valuation technique, DeRDaVa, based on our solutions to (C1) and (C2) and discuss its efficient approximation via sampling. Lastly, in Sec.~\ref{subsec:risk}, we describe a generalization of DeRDaVa for risk-averse or risk-seeking model owners.

\subsection{Random Cooperative Game and Deletion-Robustified Fairness Axioms}\label{subsec:random-cooperative-game-and-robustified-fairness-axioms}

Let $\dmax$ denote the initial set of $n$ data sources before data deletions. In Sec.~\ref{sec:background-and-related-works}, we model the problem as a cooperative game $\langle \dmax, v \rangle$ with data valuation function $\phiopt$. When data deletion occurs, the support set $D_n$ shrinks to a smaller set $D' \subseteq \dmax$ but the same model utility function $v$ still maps any subset of the new support set $D'$ to its utility. In this section, we further consider a \textit{random cooperative game} $\langle \mathbf{D}, v \rangle$ to account for deletions. The  \textit{random staying set}  $\mathbf{D}$ is a subset of $\dmax$ and follows some probability distribution $P_\mathbf{D}$ (e.g., in Fig.~\ref{fig:example}, $P_\mathbf{D}$ is a categorical distribution with parameters $.7$ and $.3$). 

In practice, the model owner sets $P_\mathbf{D}$~by 
\squishlisttwo
    \item estimating the \textit{independent} probability $\text{Pr}[\mathbb{I}_{d_i} = 1]$ each data source $d_i$ stays in the collaboration from their surveys/histories, where $\mathbb{I}_{d_i}$ is an indicator variable which evaluates to $1$ when $d_i$ stays (not delete) and $0$ otherwise;
    \item weighing the emphasis of having only $|D'|$ data sources staying out of $\dmax$ (e.g., if the model owner intends to recompute the valuation scores when there are $\geq 30$ deletions, it should place higher probability on larger $D'$ with $|D'| > n - 30$);
    \item using the normalized ``reputation'' score of data source $d_i$ or subset $D'$ (i.e., how unlikely $d_i$ or $D'$ is malicious and deletion-worthy in upcoming data audits)\footnote{App. \ref{appendix:P_D} further discusses how to set $P_{\mathbf{D}}$.}.
\squishend

Instead of expensively recomputing semivalues every time a deletion happens,
we seek a deletion-robust data valuation function $\tau$ that acts on the random cooperative game $\langle \mathbf{D}, v \rangle$. The function assigns each data source $d_i$ a fair valuation score $\tau_i(v)$ that accounts for anticipated deletions once/upfront. 
To ensure the fairness of $\tau$, we examine and ``robustify'' each of the previously mentioned axioms Linearity, Dummy Player, Interchangeability and Monotonicity with minimal changes such that the robustified versions are desirable in our problem setting. 

The Linearity axiom is a very important requirement for any cooperative game and valuation scheme \citep{shapley1953value} since it provides a way to formally analyze games with linear algebra. Moreover, it ensures that if the marginal contribution of data source $d_i$ to each coalition $S$ doubles, then the valuation score assigned to $d_i$ shall also double; if a data source brings zero marginal contribution to all coalitions, then its valuation score shall be zero. This property is clearly still desirable in our problem setting:
\begin{axiom}\label{ax:robust-linearity}
    \textup{[Robust Linearity]} Given a random support set $\mathbf{D} \subseteq \dmax$ and any two model utility functions $v$ and $w$, a fair data valuation function $\tau$ shall satisfy 
    \begin{equation}
        \forall d_i \in \dmax \quad  [\tau_i(v) + \tau_i(w) = \tau_i(v + w)].
    \end{equation}
\end{axiom}

The Dummy Player axiom defines a specific type of data source called \textit{dummy player}, whose marginal contribution is always equal to its own utility (i.e., ${\forall S \subseteq \dmax \setminus \{d_i\}} \quad  [\text{MaC}_v(d_i | S) = v(\{d_i\})]$). The axiom states that the valuation score assigned to a dummy player shall be equal to its own utility since its marginal contribution is equal to its own utility in all cases. However, in our problem setting, consider two dummy players $\mathtt{DP}_i$ and $\mathtt{DP}_j$ with equal own utility, where $\mathtt{DP}_i$ always stays in the collaboration while $\mathtt{DP}_j$ stays or deletes with a $50$-$50$ chance. Although their contributions to pre-deletion model performance are the same, $\mathtt{DP}_i$ contributes more to model performance after anticipated data deletions (i.e., deletion-robustness) than $\mathtt{DP}_j$. Therefore, the original Dummy Player axiom is no longer desirable. Instead, our data valuation function should only reward a dummy player for cases where it stays in $\mathbf{D}$:
\begin{axiom}\label{ax:robust-dummy-player}
    \textup{[Robust Dummy Player]} A data source $\mathtt{DP}$ is called a \textbf{dummy player} if its marginal contribution is always equal to its own utility. For any dummy player $\mathtt{DP}$, a fair data valuation function $\tau$ shall satisfy
    \begin{equation}
    \tau_{\mathtt{DP}}(v) = \mathbb{E}_{\mathbf{D} \sim P_{\mathbf{D}}} \left[ v(\{\mathtt{DP}\}) \cdot \mathbb{I}[\mathtt{DP} \in \mathbf{D}] \right],
    \end{equation}
    where $\mathbb{I}[\mathtt{DP} \in \mathbf{D}]$ is an indicator variable that equates to $1$ when $\mathtt{DP}$ is present in $\mathbf{D}$ and vice versa.
\end{axiom}

The Interchangeability axiom defines that two data sources are \textit{interchangeable} if their marginal contributions to any coalition $S$ are always equal. It states that two interchangeable data sources shall receive the same valuation score. However, in a random cooperative game, two interchangeable data sources that have different probabilities of staying contribute differently to deletion-robustness, and their valuation scores should not be equal (e.g., $\bigstar$ and $\blacksquare$ in Fig.~\ref{fig:example}). Therefore, we add a further constraint to this axiom:
\begin{axiom}\label{ax:robust-interchangeability}
    \textup{[Robust Interchangeability]} Two data sources $d_i$ and $d_j$ are said to be \textbf{robustly interchangeable} ($d_i \cong d_j$) if their marginal contribution to any coalition $S \subseteq \dmax$ is always equal and their probability of staying alongside others sources $D' \subseteq \dmax \setminus \{d_i, d_j\}$ are also equal. The valuation scores assigned to any two robustly interchangeable data sources shall be equal:
    \begin{equation}
        d_i \cong d_j \Rightarrow \tau_i(v) = \tau_j(v).
    \end{equation}
\end{axiom}

Finally, the Monotonicity axiom states that if every data source makes a non-negative marginal contribution to every coalition (i.e., model utility function $v$ is monotone increasing), then their valuation scores shall be non-negative. This is still valid in a random cooperative game because if $v$ is monotone increasing, then the marginal contribution of any data source is at least $0$ even if it quits the collaboration. Therefore, we keep the original version of this axiom:
\begin{axiom}\label{ax:robust-monotonicity}
    \textup{[Robust Monotonicity]} If model utility function $v$ is monotone increasing, then the valuation score assigned to any data source shall be non-negative:
    \begin{align}
        &\ \forall S, T \subseteq \dmax  \quad  [S \subseteq T \Rightarrow v(S) \leq v(T)] \nonumber \\
         \Rightarrow &\ \forall d_i \in \dmax  \quad  [ \tau_i(v) \geq 0].
    \end{align}
\end{axiom}

With the formalization of Axioms \ref{ax:robust-linearity} to
\ref{ax:robust-monotonicity}, we proceed to find a solution that satisfies these axioms.

\subsection{Null-Player-Out (NPO) Consistency and Extension}\label{subsec:npo}
After data deletions, the number of data sources will be $<n$. Thus, we can no longer directly apply the $n$-sources data valuation function $\phi^n$. Instead,  
we need to derive a sequence of semivalues $\Phi = \langle \phi^k : k = 1, 2, \cdots, n \rangle$ to value every game with support set sized $k$ ranging from $1$ to $n$. 
We address this challenge by considering a post-deletion model utility function $\nu$ which maps each coalition of data sources to the model utility of the staying subset of $D'$, e.g., $\nu(\dmax) = v(D')$.  In the cooperative game $\langle \dmax, \nu \rangle$,  any deleted data source is a \textit{null player} \citep{van2007null}:
\begin{definition}
    \textup{[Null player]} A data source in a cooperative game $\langle D, \nu \rangle$ is said to be a \textbf{null player} ($\mathtt{NP}$) if its marginal contribution to every coalition $S$ is always equal to zero, i.e.,
    \begin{equation}
        \forall S \subseteq D \setminus \{\mathtt{NP}\} \quad  [\mathrm{MaC}_\nu(\mathtt{NP} | S) = 0].
    \end{equation}
\end{definition}
A null player $\mathtt{NP}$, e.g., an empty data source, should be assigned a valuation score $\phi_{\mathtt{NP}}^n(\nu)$ of zero.

Consider the case where some data sources have quit $\dmax$ and only a subset $D' \subset \dmax$ stays as the support set.
Intuitively, the value assigned by $\phi^n$ to an undeleted data source $d_i$ in the cooperative game $\langle \dmax, \nu \rangle$ (using the post-deletion model utility function) should be the same as its value assigned by $\phi^{|D'|}$ in the cooperative game $\langle D', v \rangle$ (as though $\dmax \setminus D'$ never joined the collaboration) (\textdaggerdbl).
This condition requires us to select the sequence of semivalues $\Phi = \langle \phi^k : k = 1, \cdots, n-1 \rangle$ to be \textit{NPO-consistent}:
\begin{definition}
    \textup{[NPO-consistency]} Consider a set of data sources $D_n = \bigcup_{i = 1}^n \{d_i\}$ and a natural number $k \leq n$ such that only data sources in $D_k = \{d_i : 1 \leq i \leq k\}$ are \textbf{non}-null players in the cooperative game $\langle \dmax, \nu \rangle$. 
    Two semivalues $\phi^n: G^n \rightarrow \mathbb{R}^n$ and $\phi^k: G^k \rightarrow \mathbb{R}^k$ are \textbf{NPO-consistent} if the presence of null players (e.g., empty data sources) do not affect the values of the non-null players (who contribute valid datasets):
    \begin{equation}
        \forall d_i \in D_k \quad  [\phi^n_i(\nu) = \phi^k_i(\nu)]
    \end{equation}
    holds for all such $D_n$ and $D_k$ with fixed $n$ and $k$. Moreover, a sequence of semivalues $\Phi$ is NPO-consistent if every pair of semivalues in $\Phi$ is NPO-consistent.
\end{definition}
Note that the null players $\dmax \setminus D_k$ in $\langle \dmax, \nu \rangle$ correspond to deleted data sources.
As data sources in $D_k$ stays undeleted, $v(S) = \nu(S)$ for all $S \subseteq D_k$ and the values $\phi^k_i(\nu) = \phi^k_i(v)$. The NPO-consistent property guarantees that $\phi^n_i(\nu)$ equals $\phi^k_i(v)$, thus, achieving (\textdaggerdbl).

We then construct a sequence of semivalues $\Phi = \langle \phi^k : k = 1, 2, \cdots, n \rangle$ that is NPO-consistent with the following NPO-extension process:
\begin{theorem} \label{thm:npo-extension}
    \textup{[NPO-extension]} Every semivalue $\phi^n: G^n \to \mathbb{R}^n$ can be \textbf{uniquely} extended to a sequence of semivalues $\Phi = \langle \phi^k : k = 1, 2, \cdots, n \rangle$ that is NPO-consistent through the following unified \textbf{NPO-extension} process: \begin{enumerate}[leftmargin=*,itemsep=5pt,topsep=0pt,parsep=0pt,partopsep=0pt]
        \item\label{enum:npo-extension-step-1} From the weighting term $w_s$ in Definition \ref{def:semivalue}, calculate the quantity $w^n_s = w_s / \binom{n - 1}{s}$, which is sometimes referred to as \textbf{weighting coefficient} \citep{carreras2000note}.
        \item\label{enum:npo-extension-step-2} Calculate the weighting coefficients $w^{n-1}_s$ of $\phi^{n-1}$ using the formula $w^{n-1}_s = w^n_s + w^n_{s + 1}$. We can therefore construct $\phi^{n-1}$ by setting the weighting term $w_s$ in $\phi^{n - 1}$ to be $w^{n-1}_s \cdot \binom{n - 2}{s}$ for each $s = 0, 1, \cdots, n - 2$.
        \item Repeat Steps \ref{enum:npo-extension-step-1} and \ref{enum:npo-extension-step-2} until we have constructed every semivalue in $\Phi$.
    \end{enumerate}
\end{theorem}

\noindent
\textit{Intuition behind NPO-extension.}
Consider the cooperative game $\langle D, \nu \rangle$ with $|D| = n$ with one null player $\mathtt{NP}$. The marginal contribution of any non-null data source $d_i \neq \mathtt{NP}$ to any coalition $S$ without the null player (i.e., $\mathtt{NP} \notin S$) is always equal to its marginal contribution to $S \cup \{\mathtt{NP}\}$ (i.e., 
$\mathrm{MaC}_\nu(d_i | S) = \mathrm{MaC}_\nu(d_i | S \cup \{\mathtt{NP}\})$.
For $\phi^{n - 1}$ to be NPO-consistent with $\phi^n$, the total weights on 
$\mathrm{MaC}_\nu(d_i | S)$ must be equal in $\phi^{n-1}$ and $\phi^{n}$.
Hence, the weighting coefficient $w^{n-1}_s$ must equal the sum of the weighting coefficients of the above two marginal contribution terms ($w^n_s + w^n_{s+1}$) \citep{domenech2016some}. 
In App. \ref{appendix:proof-npo-extension}, we formally prove the validity and uniqueness of the result in Theorem \ref{thm:npo-extension} and the NPO-consistent property of $\Phi$ defined using common semivalues such as Data Shapley \citep{ghorbani2019data, jia2019towards}, Beta Shapley \cite{kwon2021beta} and Data Banzhaf \citep{wang2023data}.

\subsection{DeRDaVa and Its Efficient Approximation} \label{subsec:urdava-and-its-efficient-approximation}

Let $\Phi = \{\phi^k : k = 1, 2, \cdots n\}$ be the sequence of semivalues derived from $\phiopt$ using NPO-extension. Each data source's contribution to model performance and deletion-robustness can be regarded as an aggregate of its contribution (measured by $\phi^{|D'|} \in \Phi$) to every possible staying set $D' \subseteq \dmax$. Hence, we take the expectation of valuation scores $\phi_i^{|\mathbf{D}|} (v)$ over the probability distribution $P_{\mathbf{D}}$ and regard $d_i$ as making $0$ contribution when $d_i \notin D'$. This leads to the formal definition of DeRDaVa:
\begin{definition} \label{def:urdava}
    \textup{[DeRDaVa]} Let $\langle \mathbf{D}, v \rangle$ be a random cooperative game with random support set $\mathbf{D}$ over some probability distribution $P_{\mathbf{D}}$ where the maximal support set $\dmax$ contains $n$ data sources. Suppose also that $\phiopt: G^n \to \mathbb{R}^n$ is the jointly-agreed semivalue and $\Phi = \{\phi^k : k = 1, 2, \cdots n\}$ is the sequence of semivalues derived from $\phiopt$ using NPO-extension. The \textbf{DeRDaVa score with $\pmb\phiopt$ prior} assigned to data source $d_i$, $\tau_i(v)$, is given by
    \begin{align}
        \label{eqn:urdava}\textstyle
            \tau_i(v) & = \mathbb{E}_{\mathbf{D} \sim P_{\mathbf{D}}} \left[\mathbb{I}[d_i \in \mathbf{D}] \cdot \phi_i^{|\mathbf{D}|} (v)\right] \nonumber \\
            & = \sum_{D' \subseteq \dmax} \bigg(P_{\mathbf{D}}(\mathbf{D} = D') \cdot \mathbb{I}[d_i \in D'] \cdot \nonumber \\
            & \phantom{{} = {}}
            \sum_{S \subseteq D' \setminus \{d_i\}} w^{|D'|}_{|S|} \cdot \mathrm{MaC}_{v}(d_i | S)\bigg),
    \end{align}
    where $\mathbb{I}[d_i \in \mathbf{D}]$ is an indicator variable that equates to $1$ only when $d_i$ stays present, $\phi^{|\mathbf{D}|} \in \Phi$ is the semivalue from NPO-extension, and $w^{|D'|}_{|S|}$ is the weighting coefficient to coalition $S$ in $\phi^{|D'|}$ defined in Theorem \ref{thm:npo-extension}.
\end{definition}

The fairness and uniqueness of DeRDaVa are guaranteed with the following theorem:
\begin{theorem} \label{thm:urdava}
    \textup{[Fairness and uniqueness of DeRDaVa]} Given a random cooperative game $\langle \mathbf{D}, v \rangle$ with the same notations $P_{\mathbf{D}}$, $\dmax$, $n = |\dmax|$ and $\phiopt: G^n \to \mathbb{R}^n$ as in Definition \ref{def:urdava}, the DeRDaVa function $\tau$ defined in Definition \ref{def:urdava} uniquely satisfies Axioms \ref{ax:robust-linearity}, \ref{ax:robust-dummy-player}, \ref{ax:robust-interchangeability} and \ref{ax:robust-monotonicity} defined in Sec.~\ref{subsec:random-cooperative-game-and-robustified-fairness-axioms}.
\end{theorem}

In App.~\ref{appendix:proof-urdava}, we prove the (F) fairness and (U) uniqueness of DeRDaVa. (F) involves verifying that DeRDaVa satisfies our four robustified axioms. Let $V(\cdot)$ denote the \textit{random utility function} that maps each coalition $S \subseteq \dmax$ to the random utility after deletions, i.e., $V(S) = v(S \cap \mathbf{D})$.
(U) involves identifying the \textit{static dual} 
game $\left\langle \dmax, \mathbb{E}[V(\cdot)] \right\rangle$ to our random cooperative game $\langle \mathbf{D}, v \rangle$ and proving that the original axioms of semivalues for the static dual game are equivalent to the four robustified axioms for any random cooperative game. The uniqueness of DeRDaVa then follows from the uniqueness of semivalues.

From Eq. (\ref{eqn:urdava}), we need to consider every possible pair $\langle S, D' \rangle$ of subset $S$ and staying set $D'$ such that $S \cup \{d_i\} \subseteq D' \subseteq \dmax$. Each data source $d_j \neq d_i$ has exactly three \textit{states}: (i) It is in neither $S$ nor $D'$ (State 0); (ii) It is in $D'$ but not in $S$ (State 1); (iii) It is in both $S$ and $D'$ (State 2). Thus, the total number of unique state combinations or pairs $\langle S, D' \rangle$ needed to exactly compute DeRDaVa scores is $O\left(3^n\right)$. 
Exact computation is intractable when the number of data sources is large. Thus, model owners should
efficiently approximate DeRDaVa scores based on Monte-Carlo sampling
and additionally use our \textbf{012-MCMC algorithm} when it is hard to estimate/sample from $P_\mathbf{D}$.

\paragraph{Monte-Carlo Sampling} DeRDaVa scores can be alternatively viewed as the expectation of marginal contribution $\mathrm{MaC}_v(d_i | S)$ over some distribution of staying set $D'$ (i.e., $P_{\mathbf{D}}$) and coalition $S$. Therefore, it is natural to use Monte-Carlo sampling when it is tractable to sample from $P_{\mathbf{D}}$ directly. For example, for the special case where data sources decide to stay/delete independently, we sample whether each data source stays to determine staying set $D'$, the size $s$ of coalition $S$ (using the weighting coefficients) and lastly, $s$ data sources out of $D'$. In App.~{\ref{appendix:mc-sampling}}, we prove that the number of samples needed to approximate DeRDaVa with $(\delta, \epsilon)$-error is $O(\frac{2r^2 n}{\epsilon^2}\log \frac{2n}{\delta})$, where $r$ is the range of model uility function $v$. 

\paragraph{012-MCMC} When direct Monte-Carlo sampling is hard, we propose to repeatedly sample the state for each source in $\dmax \setminus \{d_i\}$ from the uniform distribution over $\{0, 1, 2\}$. For the $t$-th sample, we construct $S_{(t)} = \{d_j : \mathtt{state}(d_j)=2 \}$ and $D'_{(t)} = \{d_j : \mathtt{state}(d_j) \neq 0 \} \cup \{d_i\}$. Thus, we enforce $S_{(t)} \cup \{d_i\} \subseteq D'_{(t)}$. The DeRDaVa score of source $d_i$, $\tau_i(v)$, is then approximated by importance sampling and taking the average of $T$ samples:
\begin{equation} \label{eqn:urdava-sampling}\textstyle
    \frac{1}{T} \sum\limits_{t=1}^T \left(\frac{P_\mathbf{D}\left(\mathbf{D} = D'_{(t)} \right)}{1/3^{(n-1)}} \cdot w^{|D'_{(t)}|}_{|S_{(t)}|} \cdot \mathrm{MaC}_{v}(d_i | S_{(t)}) 
    \right)\ . 
\end{equation}
By using importance sampling, we avert computing $P_\mathbf{D}$ for every subset. Instead, we use the known probability $\textstyle P_\mathbf{D}\left(\mathbf{D} = D'_{(t)} \right)$ for each sampled staying set $D'_{(t)}$. 

In practice, we construct $M$ parallel Markov chains of samples and use the Gelman-Rubin statistic \citep{gelman1992inference} to assess the convergence of the approximation. The threshold for the Gelman-Rubin statistic is usually set around $1.1$ \citep{vats2021revisiting}, but we set it to $\leq 1.005$ for higher approximation precision.
The time complexity of our 012-MCMC algorithm depends on the number of samples generated which is significantly smaller than $O(3^n)$. The justification and pseudocode for 012-MCMC algorithm are included in App. \ref{appendix:012-mcmc}.

\subsection{Risk-DeRDaVa: A Variant for Different Risk Attitudes}\label{subsec:risk}

In Sec.~\ref{subsec:urdava-and-its-efficient-approximation}, the DeRDaVa scores are equivalent to the values assigned by $\phi^n$ on the static dual game $\langle \dmax, \mathbb{E}[V(\cdot)] \rangle$. The model owner considers each data source's marginal contribution to the \textbf{expected} utility of each coalition $S$. The model owner is \emph{risk-neutral} and indifferent between (R1) a constant random utility function $V(\cdot)$ or (R2) a varying $V(\cdot)$ with a worse worst-case (with equal expected values).

\begin{figure}
  \centering
  \begin{subfigure}[b]{0.32\linewidth}
    \centering
    \includegraphics[width=\textwidth]{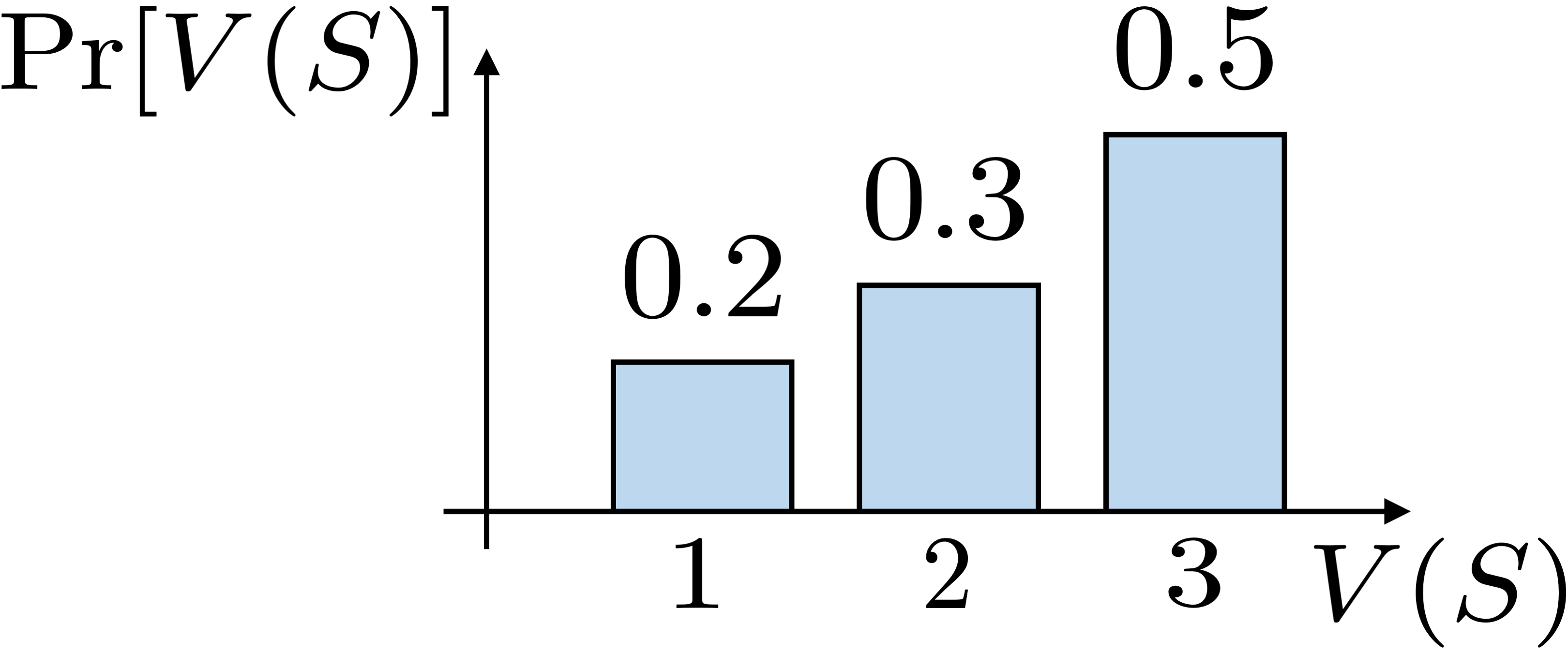}
    \caption{Risk-neutral.}\label{img:risk-neutral}
  \end{subfigure}
  \begin{subfigure}[b]{0.32\linewidth}
    \centering
    \includegraphics[width=\textwidth]{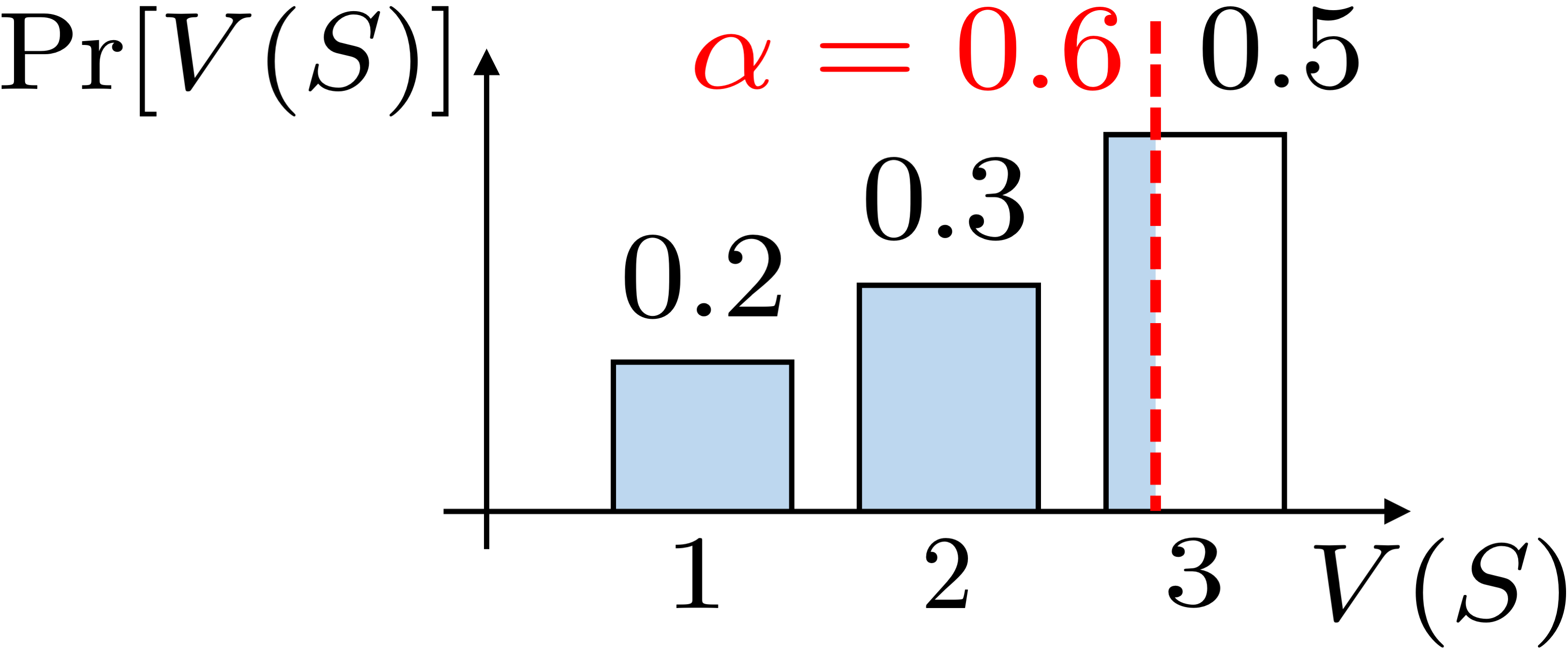}
    \caption{Risk-averse.}\label{img:risk-averse}
    \end{subfigure}
  \begin{subfigure}[b]{0.32\linewidth}
    \centering
    \includegraphics[width=\textwidth]{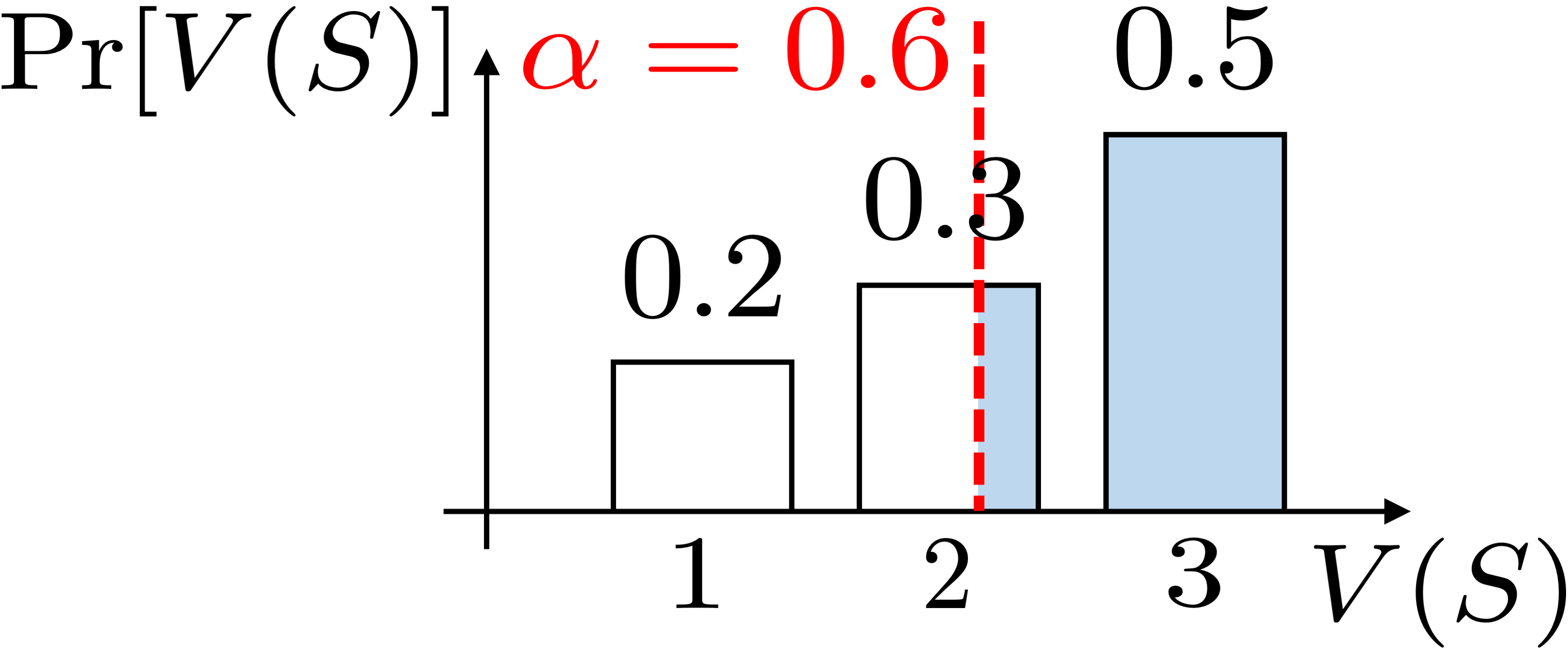}
    \caption{Risk-seeking.}\label{img:risk-seeking}
  \end{subfigure}
  \caption{Model owners with different risk attitudes will map the random utility function $V(S)$ evaluated at coalition $S$ to a deterministic value differently.
  The risk-neutral owner (\ref{img:risk-neutral}) takes expectation (blue) over all possible utilities.
  A risk-averse (\ref{img:risk-averse})/
  risk-seeking (\ref{img:risk-seeking}) owner takes expectation over the lower/worst $0.6$-tail and upper/best $0.6$-tail respectively.}
  \label{fig:c-cvar-illustration}
\end{figure}

In practice, model owners may strictly prefer R1 or R2. \emph{Risk-averse} model owners would prefer R1 with a higher worst-case model utility and thus highly value data sources that have higher marginal contributions to the worst-case model utility.
In contrast, \emph{risk-seeking} model owners would prefer R2 with a higher best-case model utility and thus value those data sources that have higher marginal contributions to the best-case model utility. 
In Fig.~\ref{fig:c-cvar-illustration}, we illustrate and contrast how the risk-neutral, risk-averse and risk-seeking model owners will transform the random utility function $V(\cdot)$ to a static dual game.

To quantify risks, we use discrete Conditional Value-at-Risk (CVaR) \citep{uryasev2010var} to define Coalitional Conditional Value-at-Risk (C-CVaR) in our problem setting. There are two types of C-CVaR, namely \textit{Risk-Averse C-CVaR} (C-CVaR$^-$) and \textit{Risk-Seeking C-CVaR} (C-CVaR$^+$), which corresponds to the expectation within the lower tail (Fig.~\ref{img:risk-averse}) and upper tail (Fig.~\ref{img:risk-seeking}) respectively. For example, consider the $V(S)$ given in Fig.~\ref{img:risk-averse}. The C-CVaR$^-$ at level $\alpha = 0.6$ is the expectation of $V(S)$ in the blue shaded region (i.e., the lower $60\%$ tail): $\textup{C-CVaR}^-_{0.6}[V(S)] = \frac{0.2}{0.6} \times 1 + \frac{0.3}{0.6} \times 2 + \frac{0.1}{0.6} \times 3 = \frac{11}{6}$. A formal definition of C-CVaR$^-$ and C-CVaR$^+$ is included in App.~\ref{appendix:non-additivity-of-risk-urdava}. Therefore, to provide a suitable solution for risk-averse/seeking model owners, we consider the static prior game $\left\langle \dmax, \textup{C-CVaR}^{\mp}_\alpha [V(\cdot)] \right\rangle$ whose data valuation function is $\phiopt$ and define Risk-DeRDaVa:
\begin{definition}
    \textup{[Risk-DeRDaVa]} Given a random cooperative game $\langle \mathbf{D}, v \rangle$ with the same notations $P_{\mathbf{D}}$, $\dmax$, $n = |\dmax|$ and $\phiopt: G^n \to \mathbb{R}^n$ as in Definition \ref{def:urdava}, first define for any coalition $S \subseteq \dmax$ the \textbf{random utility function} $V(S) = v(S \cap \mathbf{D})$.
    Let $v_{\textup{risk}}(S) = \textup{C-CVaR}^{\mp}_\alpha [V(S)]$. The \textbf{Risk-DeRDaVa score with $\pmb\phiopt$ prior at level $\pmb\alpha$} for risk averse/seeking model owners is defined as 
    \begin{equation} \label{eqn:risk-urdava} \textstyle
        \rho_i(v) = \sum\limits_{S \subseteq \dmax \setminus \{d_i\}} w_{|S|} \cdot {\mathrm{MaC}_{v_{\mathrm{risk}}}(d_i | S)}/ {\binom{n - 1}{s}},
    \end{equation}
    where $w_{|S|}$ is the weighting term associated with all coalitions $S$ of size $s = |S|$ given by $\phiopt$.
\end{definition}

Note that $\alpha=1$ recovers the DeRDaVa scores.
In practice, it is more common for model owners to be risk-averse, so we default Risk-DeRDaVa to refer to the risk-averse version. 
As C-CVaR is non-additive (see App. \ref{appendix:non-additivity-of-risk-urdava}), we approximate Risk-DeRDaVa scores by sampling $S$ and using the Monte-Carlo CVaR algorithm \citep{hong2014monte}. 

\section{Experiments}\label{sec:experiments}

Our experiments use the following [model-dataset] combinations: [NB-CC] Naive Bayes trained on Credit Card \cite{yeh2009the}, [NB-Db] Naive Bayes trained on Diabetes \cite{carrion2022pima}, [NB-Wd] Naive Bayes trained on Wind \cite{vanschoren2014wind}, [SVM-Db] Support Vector Machine trained on Diabetes, and [LR-Pm] Logistic Regression trained on Phoneme \cite{grin2022phoneme}.
More experimental details are included in App. \ref{appendix:experimental-details}.

\subsection{Measure of Contribution to Model Performance and Deletion-Robustness}

\begin{figure} 
    \centering
    \begin{subfigure}[b]{0.4375\linewidth}
        \centering
        \includegraphics[width=\textwidth]{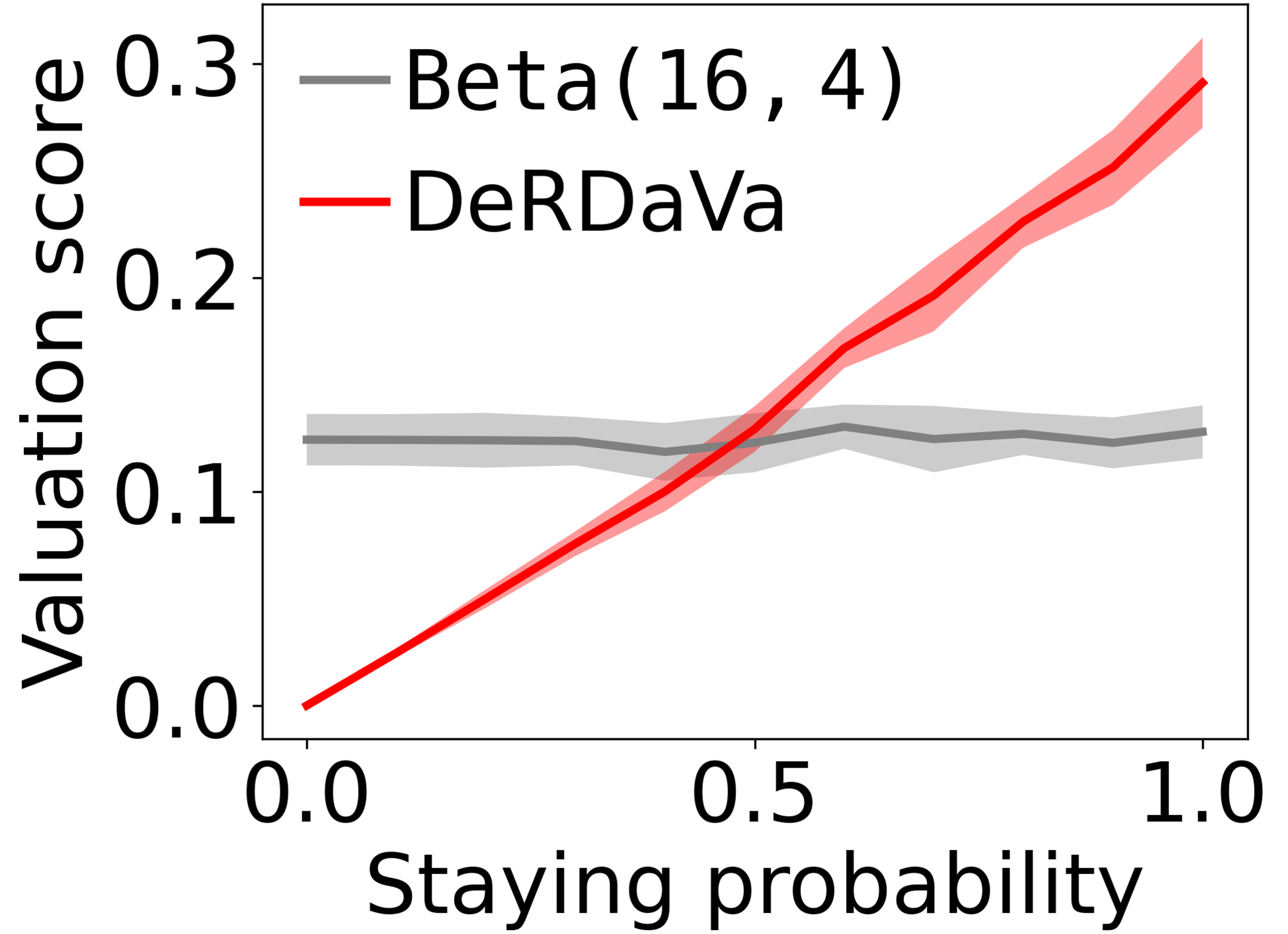}
        \caption{[SVM-Db].}\label{fig:sp-10-diabetes-svm-beta-16-4}
    \end{subfigure}
    \begin{subfigure}[b]{0.4375\linewidth}
        \centering
        \includegraphics[width=\textwidth]{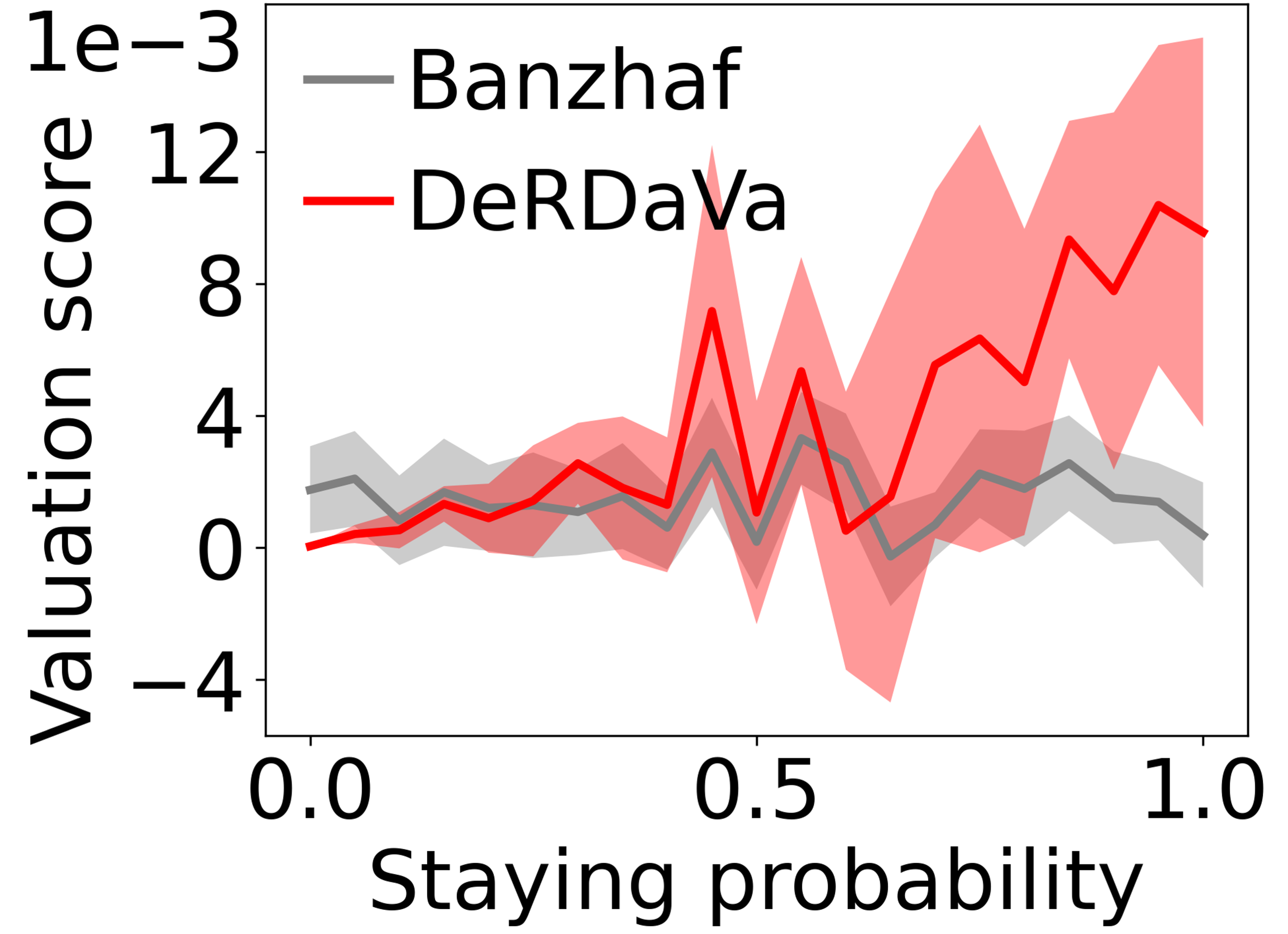}
        \caption{[NB-CC].}\label{fig:sp-20-creditcard-nb-banzhaf}
    \end{subfigure}
    \begin{subfigure}[b]{0.4375\linewidth}
        \centering
        \includegraphics[width=\textwidth]{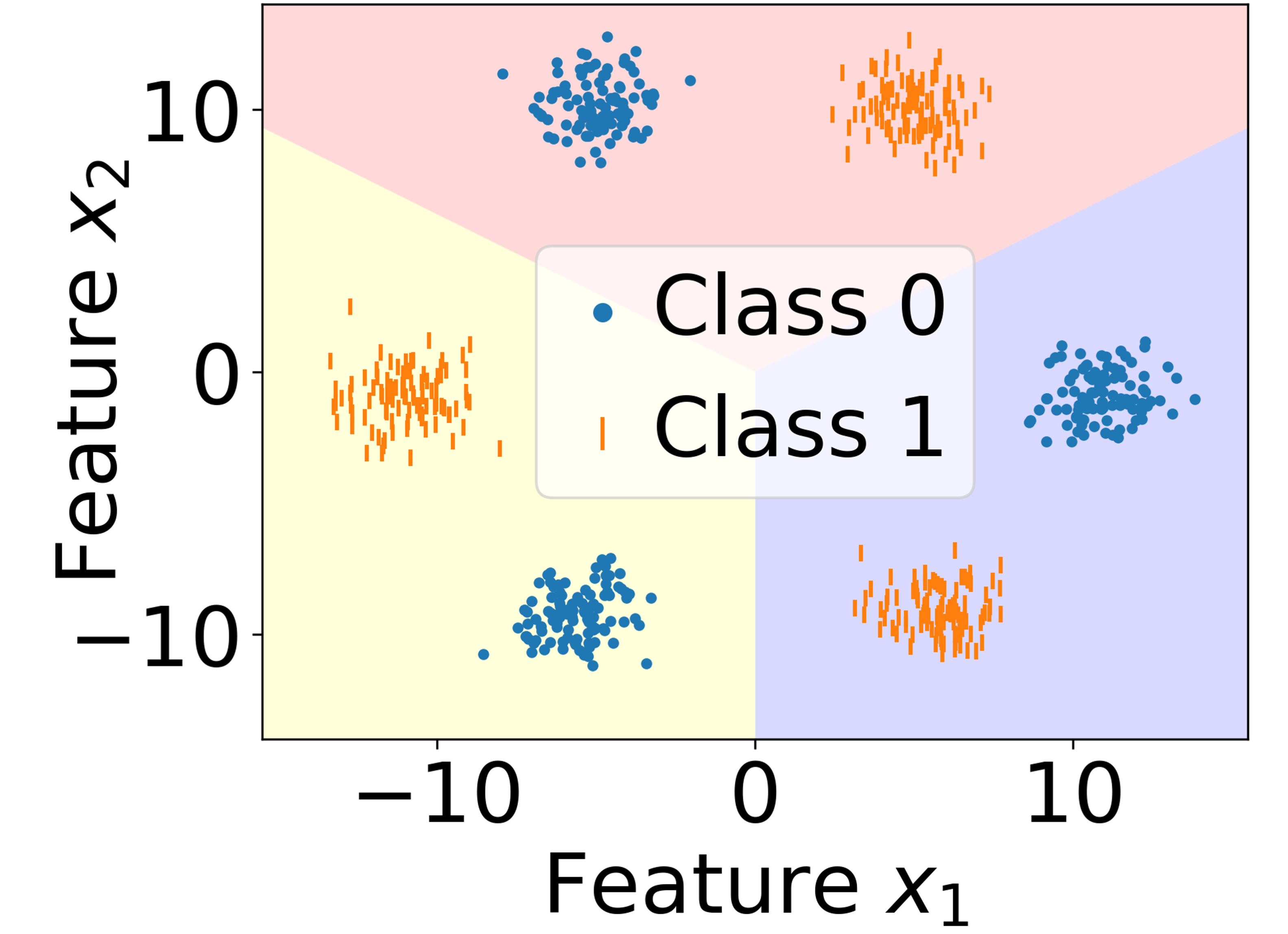}
        \caption{Synthetic dataset.}\label{fig:ds-synthetic-dataset-visualization}
    \end{subfigure}
    \begin{subfigure}[b]{0.4375\linewidth}
        \centering
        \includegraphics[width=\textwidth]{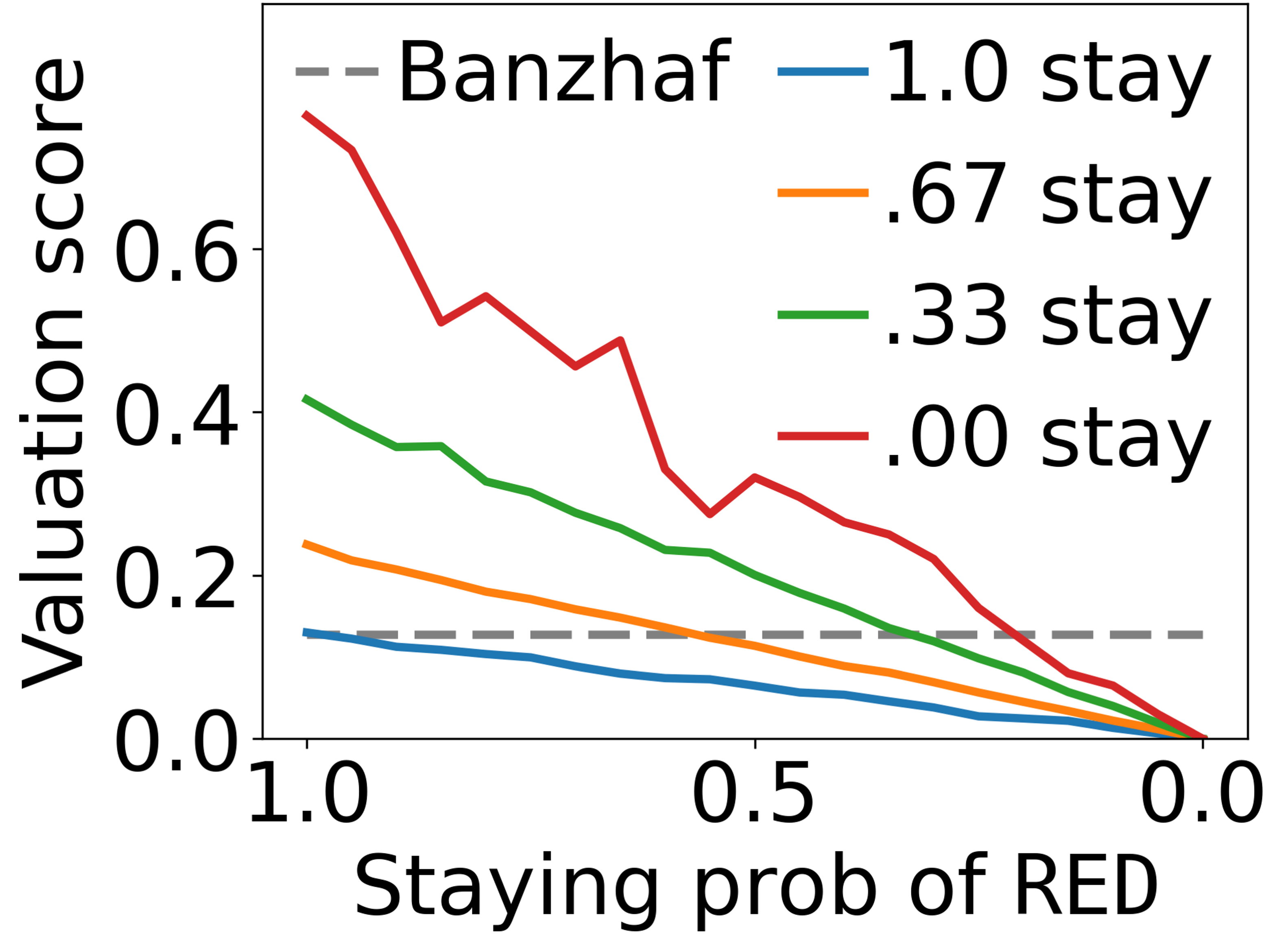}
        \caption{Data similarity.}\label{fig:ds-synthetic-dataset-result}
    \end{subfigure}
    \caption{DeRDaVa accounts for data deletions. (\ref{fig:sp-10-diabetes-svm-beta-16-4}) (11 data sources) and (\ref{fig:sp-20-creditcard-nb-banzhaf}) (21 data sources) show the effect of staying probability on DeRDaVa scores with Beta Shapley and Data Banzhaf prior; (\ref{fig:ds-synthetic-dataset-visualization}) and (\ref{fig:ds-synthetic-dataset-result}) show when DeRDaVa score of a redundant data source exceeds its Banzhaf score.}
    \label{fig:contribution-to-model-performance-and-model-robustness}
\end{figure}

DeRDaVa is designed to measure each data source's contribution to both model performance and deletion-robustness. By creating data sources with different contributions and comparing their DeRDaVa scores, we can verify the empirical behaviour of DeRDaVa. We analyse the three main factors that affect the contribution of data sources (\textbf{staying probability}, \textbf{data similarity} and \textbf{data quality}) below.

\paragraph{Staying Probability}
We repeat $50$ runs of creating data sources with equal number of randomly sampled training examples, assigning different independent staying probabilities and computing their semivalue and corresponding DeRDaVa scores. From Fig.~\ref{fig:sp-10-diabetes-svm-beta-16-4} and \ref{fig:sp-20-creditcard-nb-banzhaf}, we observe that data sources with higher staying probability receive higher DeRDaVa scores as they contribute more to model performance after anticipated deletions.

\paragraph{Data Similarity} We create a synthetic dataset (Fig.~\ref{fig:ds-synthetic-dataset-visualization}) with $4$ data sources. The yellow and blue regions are exclusively owned by $2$ different data sources while the red region is co-owned by $2$ data sources \texttt{RED} and \texttt{REDD}. Thus, \texttt{RED} and \texttt{REDD} data are highly similar. The model utility function is the accuracy of the trained $k$-Nearest Neighbours model. In Fig.~\ref{fig:ds-synthetic-dataset-result}, we observe that the \texttt{RED} is assigned a higher DeRDaVa score (plotted as solid lines) than Banzhaf score (plotted as a dashed line) when its staying probability is high and when other data sources do not stay with certainty. This aligns with our intuition that deletion-robust data valuation should favour \texttt{RED}, despite its redundancy in the presence of \texttt{REDD}, when \texttt{RED} is more likely to stay than others.

\paragraph{Data Quality} Data sources with poor data quality (e.g., with high noise level) make a low contribution to model performance regardless of data deletions. Similar to semivalues, DeRDaVa is also capable of reflecting data quality and thus can be applied to identify noisy data (see App. \ref{appendix:experimental-details}).

\subsection{Point Addition and Removal}

\begin{figure}
  \centering
  \begin{subfigure}[b]{0.44\linewidth}
    \centering
    \includegraphics[width=\textwidth]{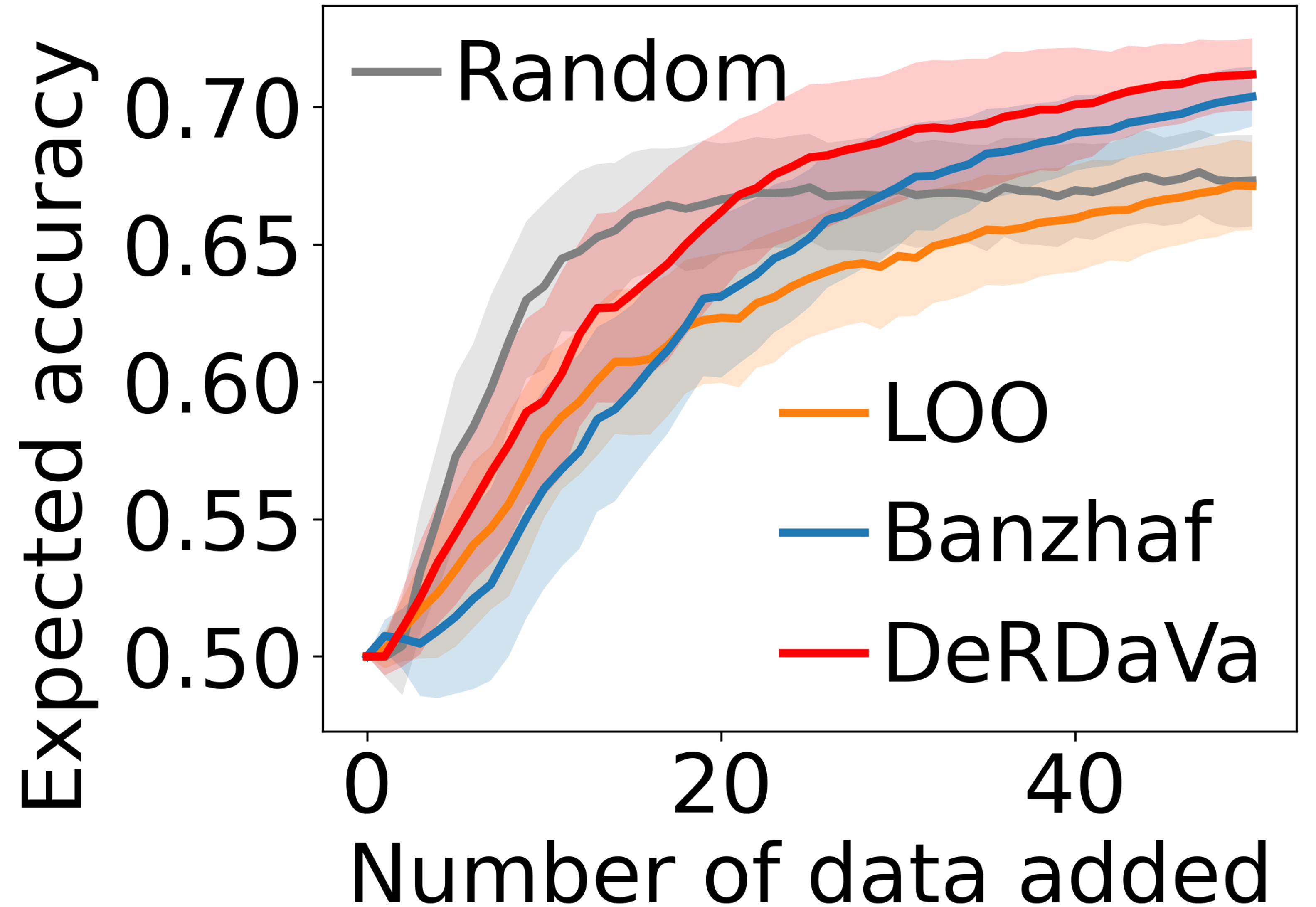}
    \caption{Add highest first.}\label{fig:par-add-highest-wind-nb-banzhaf}
  \end{subfigure}
  \begin{subfigure}[b]{0.44\linewidth}
    \centering
    \includegraphics[width=\textwidth]{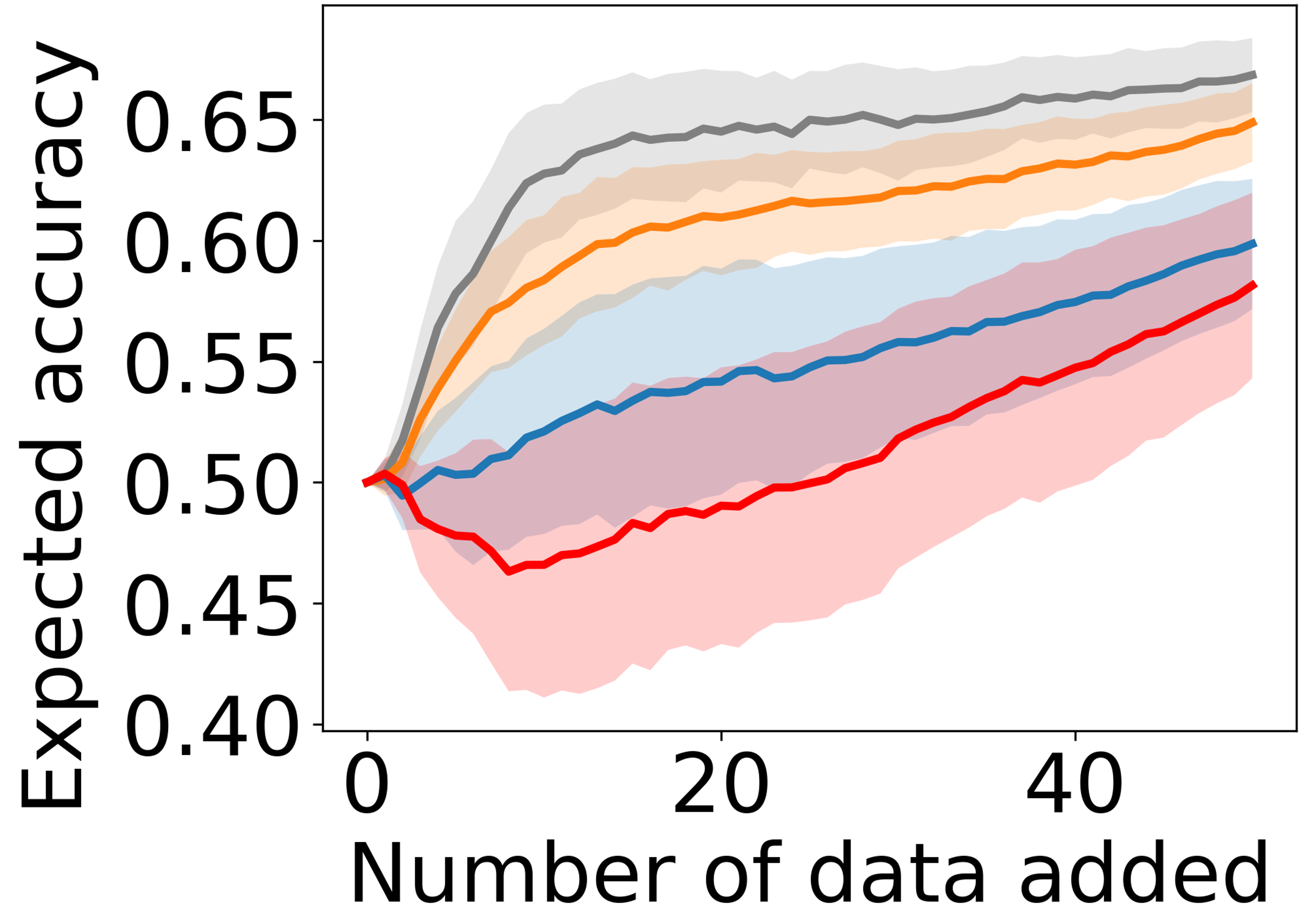}
    \caption{Add lowest first.}\label{fig:par-add-lowest-wind-nb-banzhaf}
  \end{subfigure}
  \begin{subfigure}[b]{0.44\linewidth}
    \centering
    \includegraphics[width=\textwidth]{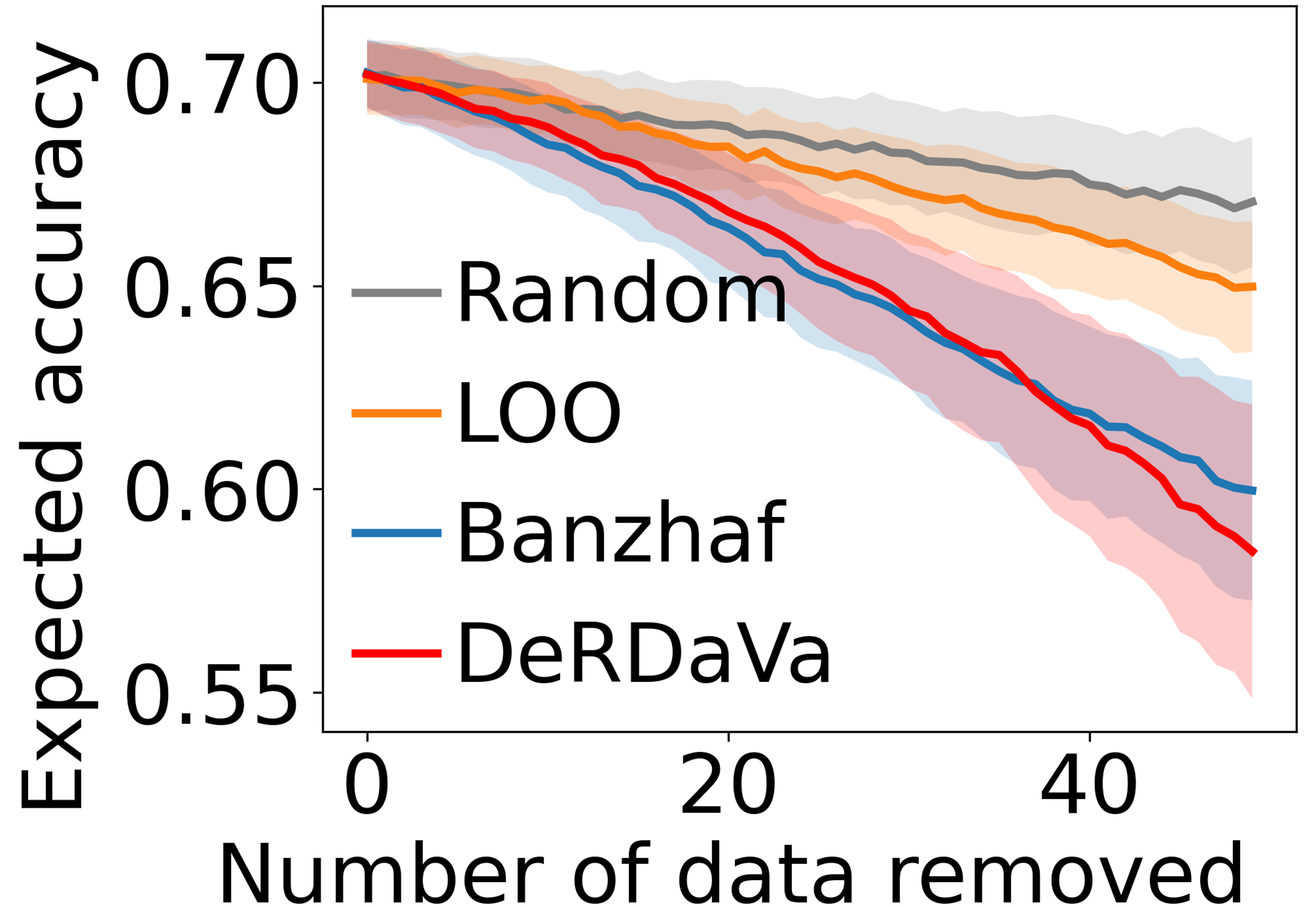}
    \caption{Remove highest first.}\label{fig:par-remove-highest-wind-nb-banzhaf}
  \end{subfigure}
  \begin{subfigure}[b]{0.44\linewidth}
    \centering
    \includegraphics[width=\textwidth]{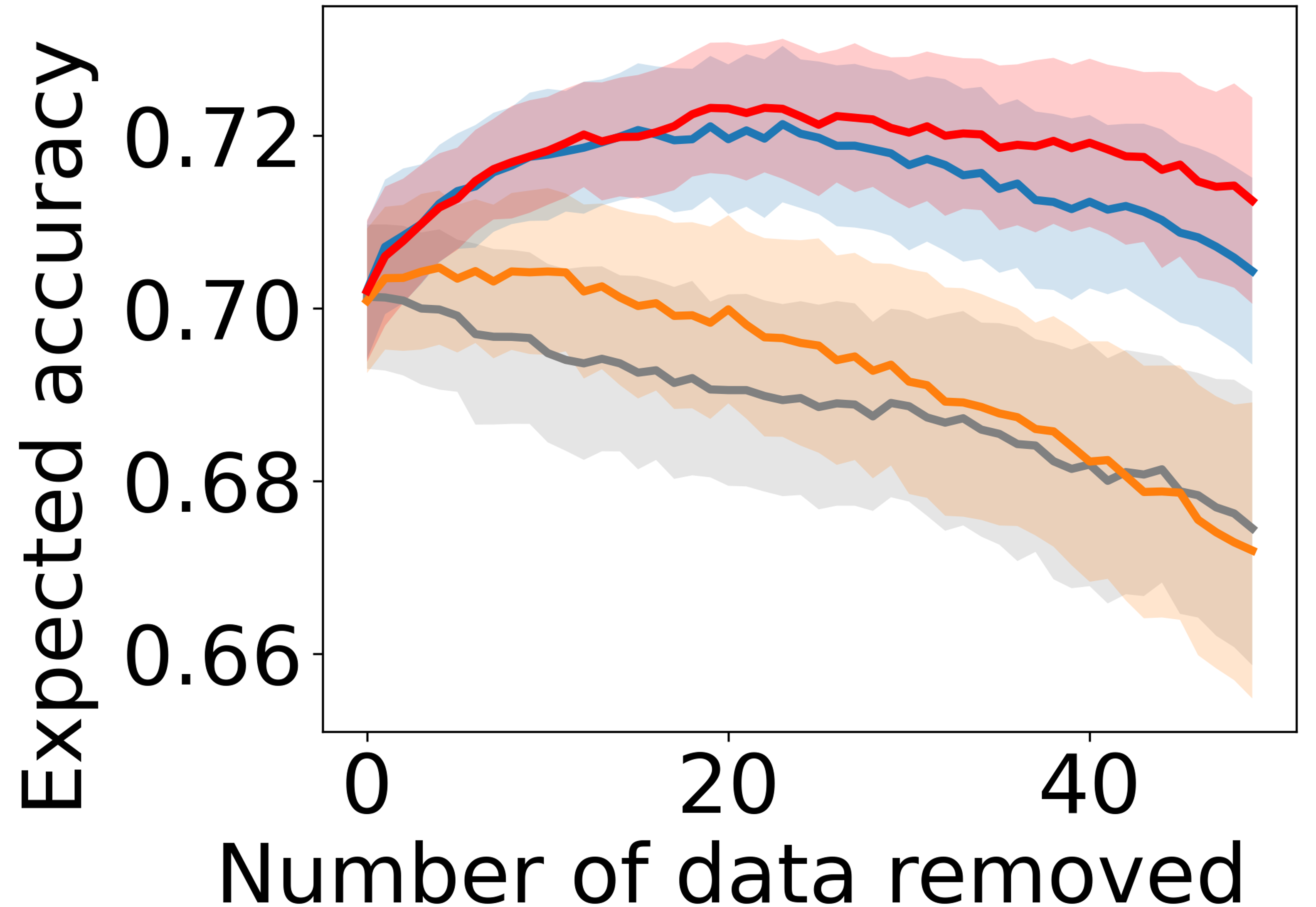}
    \caption{Remove lowest first.}\label{fig:par-remove-lowest-wind-nb-banzhaf}
  \end{subfigure}
  \caption{Point addition and removal experiments. All experiments are run using [NB-Wd], $100$ data sources and Data Banzhaf prior.}
  \label{fig:point-addition-removal}
\end{figure}

We perform point addition and removal experiments which are often used in evaluation of data valuation techniques \citep{ghorbani2019data, kwon2021beta} with an adaption to our setting --- we measure the \textbf{expected} model performance after data deletion instead. When data with the highest scores are added first (Fig.~\ref{fig:par-add-highest-wind-nb-banzhaf}), \texttt{Random} shows a rapid increase in expected model performance at the beginning as the training curve has not plateaued and almost every added point contributes a lot. However, after more additions, DeRDaVa with Banzhaf prior surpasses all others as its selected points contribute to preserving high model accuracy after anticipated deletions. When data with the lowest scores are added first (Fig.~\ref{fig:par-add-lowest-wind-nb-banzhaf}), DeRDaVa's expected accuracy drops since these data are harmful to both model performance and deletion-robustness. When data with the highest scores are removed first (Fig.~\ref{fig:par-remove-highest-wind-nb-banzhaf}), DeRDaVa exhibits a rapid decrease in expected model performance. This is because data sources that contribute more to deletion-robustness are removed. When data with the lowest scores are removed first (Fig.~\ref{fig:par-remove-lowest-wind-nb-banzhaf}), DeRDaVa demonstrates a rapid increase in expected model performance at the beginning and the slowest decrease later. This is because data which contributes to preserving higher model accuracy after anticipated deletions tend to have higher DeRDaVa scores and are not removed.

\begin{figure}
  \centering
  \begin{subfigure}[b]{0.44\linewidth}
    \centering
    \includegraphics[width=\textwidth]{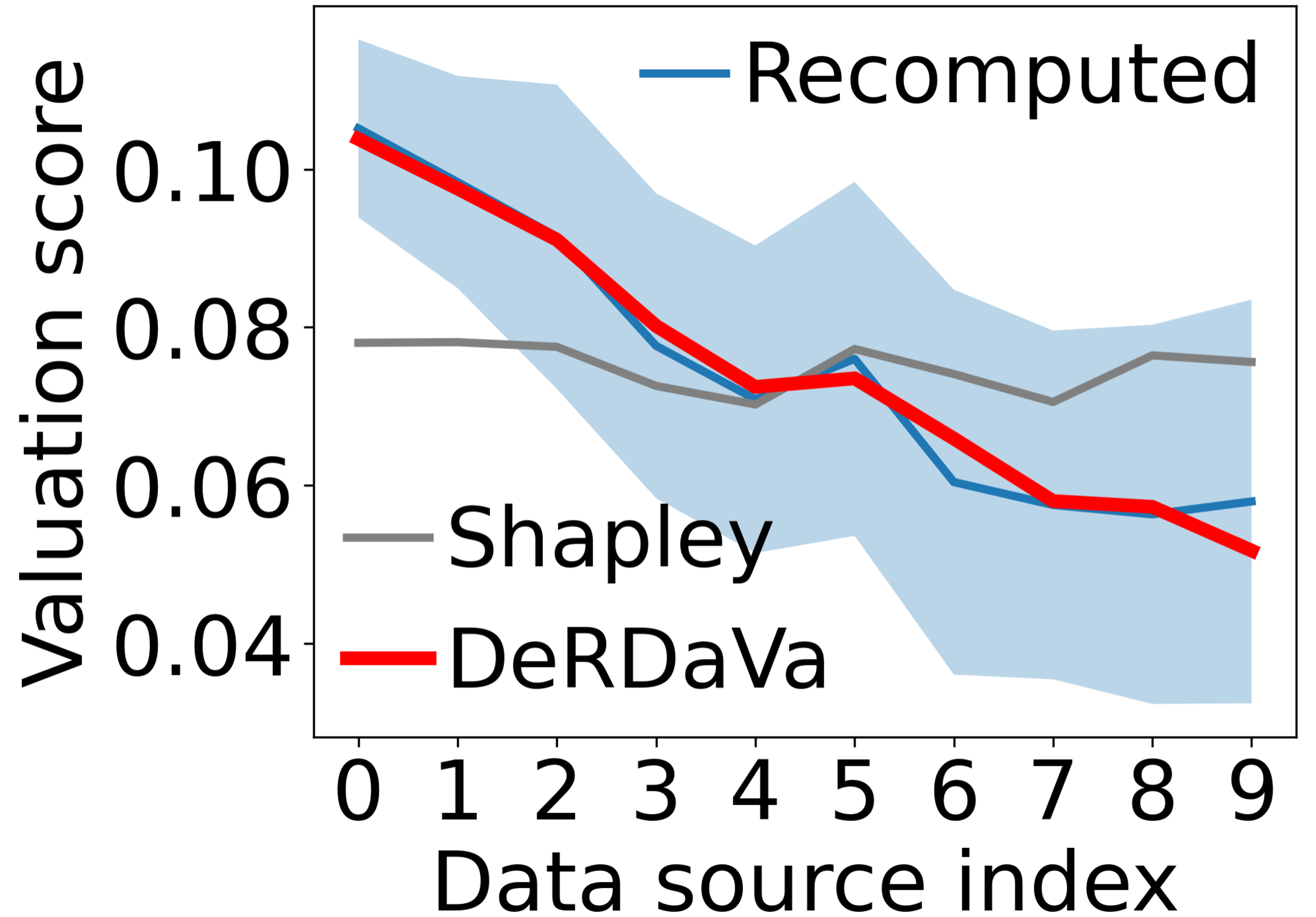}
    \caption{[LR-Pm].}\label{fig:dd-10-phoneme-logistic-shapley}
  \end{subfigure}
  \begin{subfigure}[b]{0.44\linewidth}
    \centering
    \includegraphics[width=\textwidth]{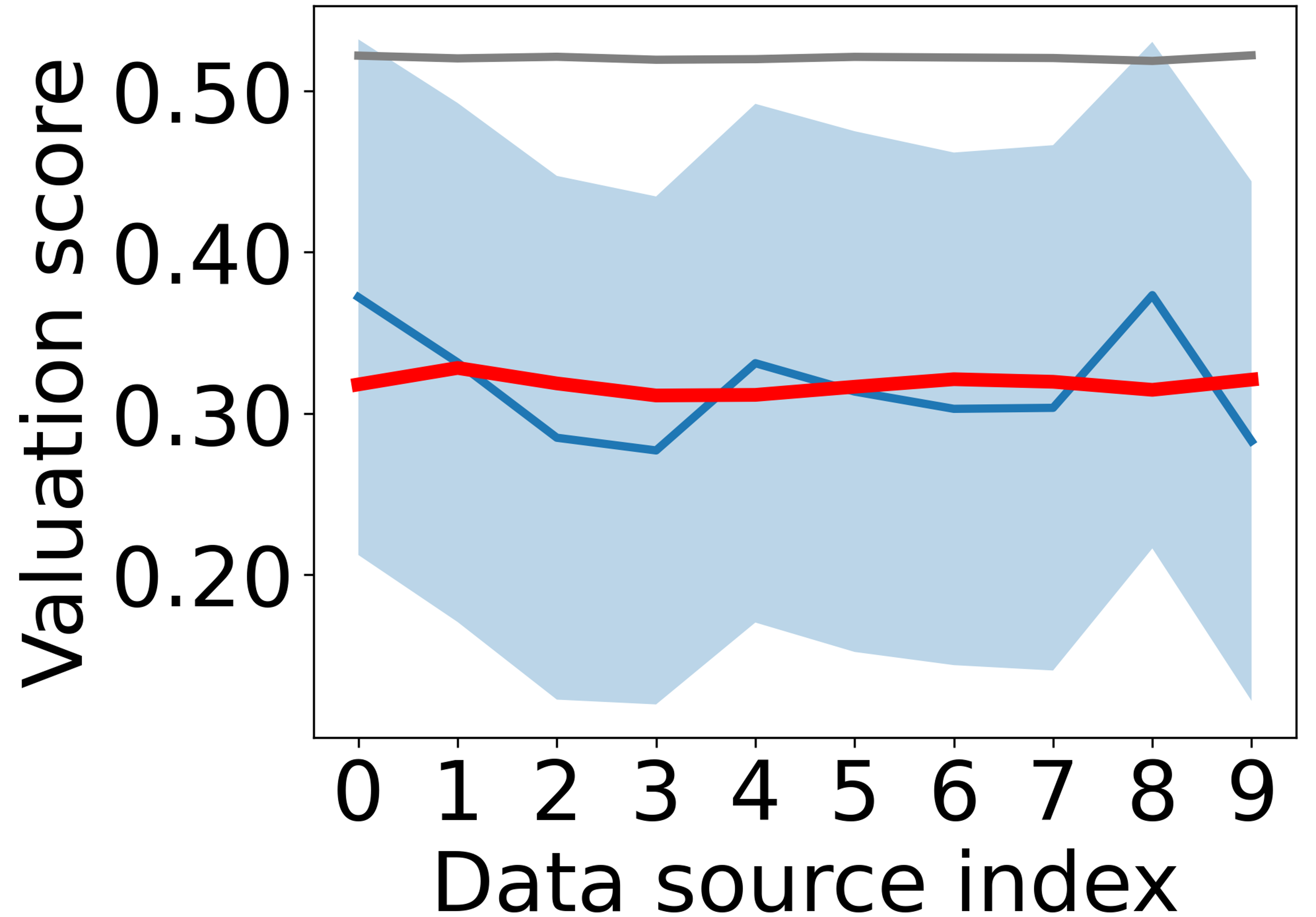}
    \caption{[NB-Wd].}\label{fig:dd-10-wind-nb-beta-16-1}
    \end{subfigure}
  \begin{subfigure}[b]{0.44\linewidth}
    \centering
    \includegraphics[width=\textwidth]{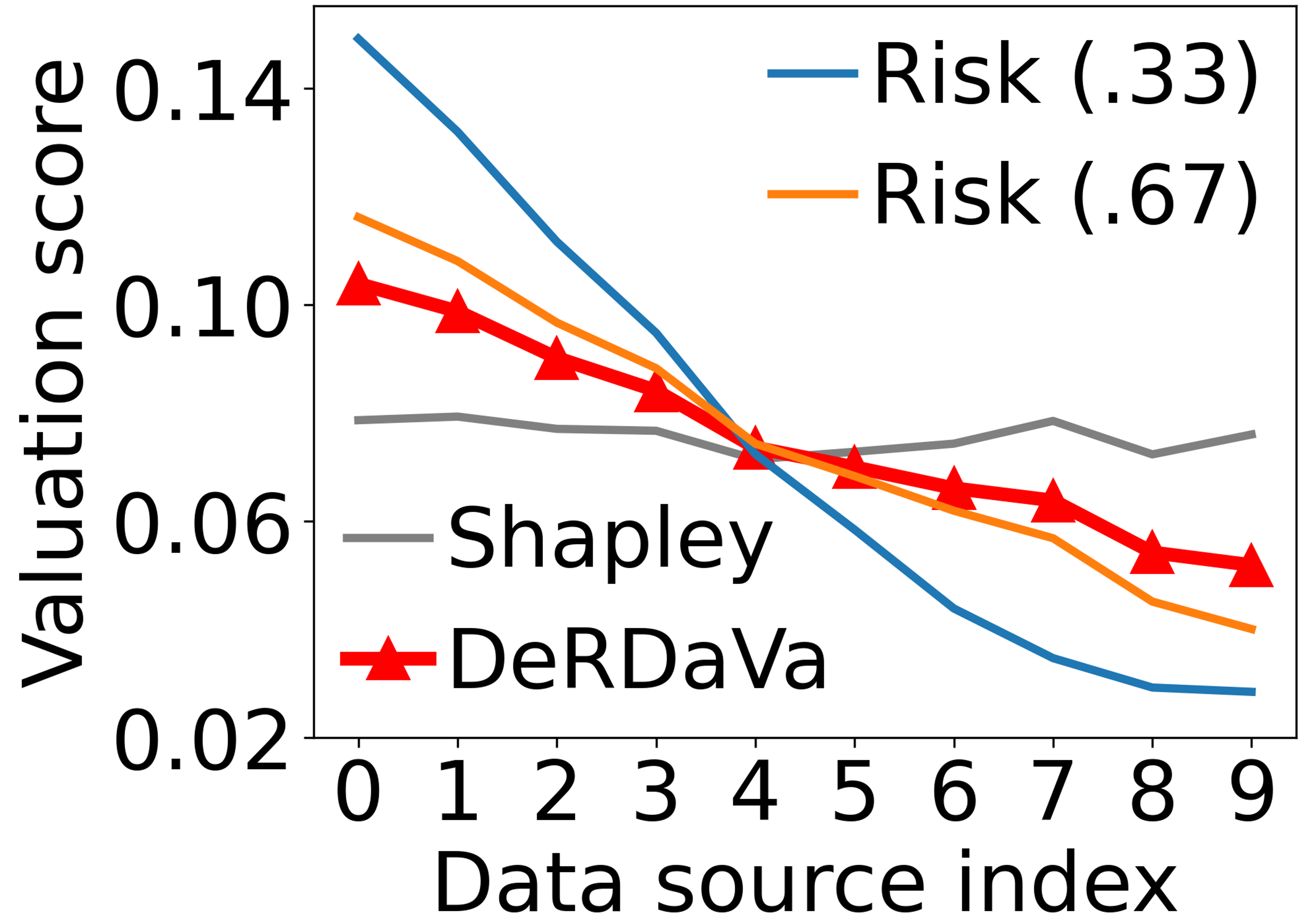}
    \caption{[LR-Pm].}\label{fig:rd-10-phoneme-logistic-shapley}
  \end{subfigure}
  \begin{subfigure}[b]{0.44\linewidth}
    \centering
    \includegraphics[width=\textwidth]{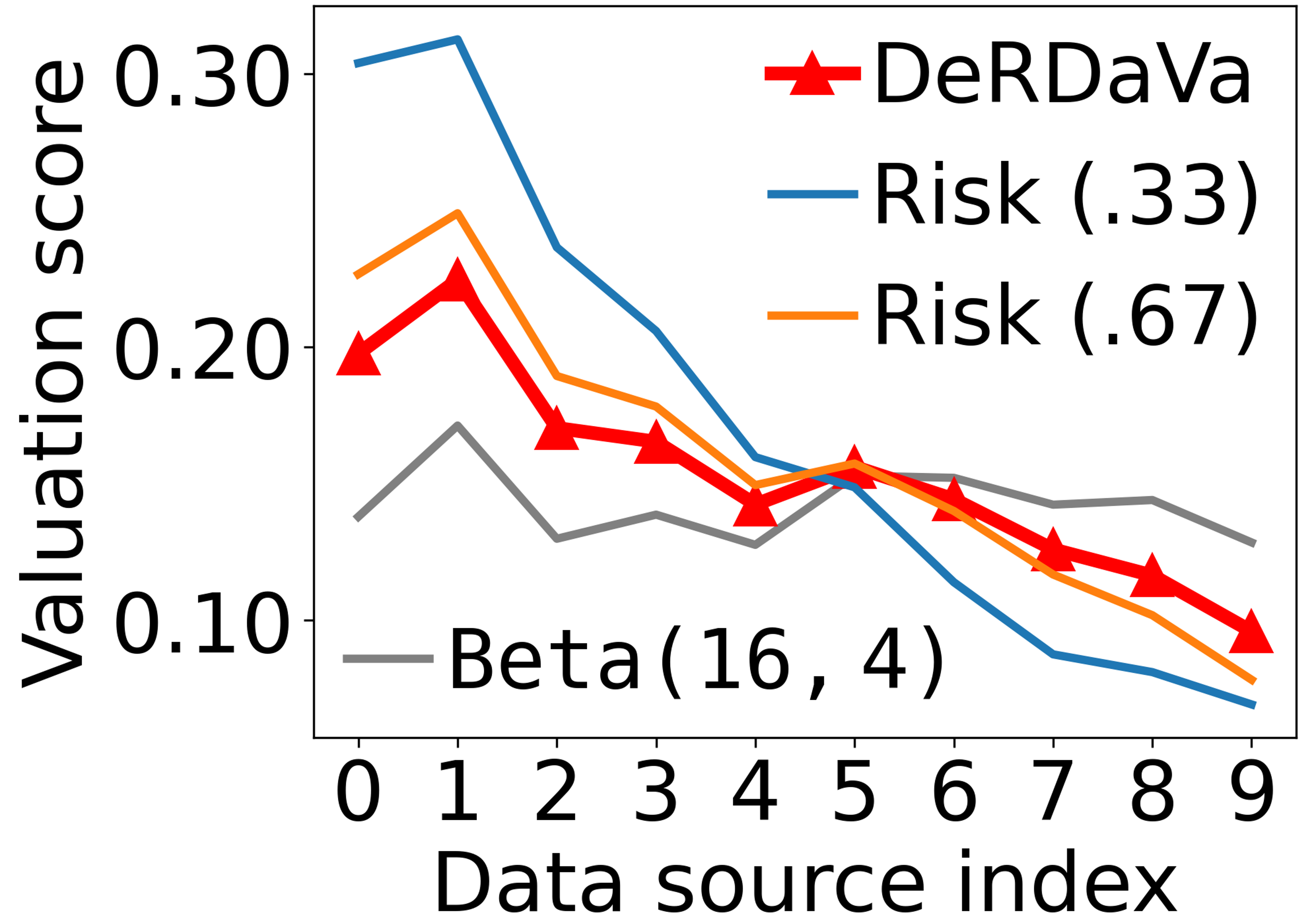}
    \caption{[NB-Db].}\label{fig:rd-10-diabetes-nb-beta-16-4}
  \end{subfigure}
  \caption{When data sources stay with independent (\ref{fig:dd-10-phoneme-logistic-shapley}) and dependent (\ref{fig:dd-10-wind-nb-beta-16-1}) probabilities, the recomputed semivalue scores of $10$ data sources always converge to DeRDaVa scores and deviate from pre-deletion scores; (\ref{fig:rd-10-phoneme-logistic-shapley}) and (\ref{fig:rd-10-diabetes-nb-beta-16-4}) compare Risk-DeRDaVa with DeRDaVa and semivalues.}
  \label{img:dd-rd}
\end{figure}

\subsection{Reflection of Long-Term Contribution}

Next, we simulate data deletions and recompute the semivalue scores to see how the contribution of a data source changes as data deletion occurs. We then compare these recomputed scores with the pre-deletion semivalue scores and DeRDaVa scores to investigate which represent the long-term contribution better. 
Fig.~\ref{fig:dd-10-phoneme-logistic-shapley} and \ref{fig:dd-10-wind-nb-beta-16-1} show that the average of the distribution of recomputed valuation scores is almost the same as the DeRDaVa scores but deviate significantly from the pre-deletion scores.
Moreover, the recomputed semivalues can vary widely (see shaded region) with different deletion outcomes. This aligns with our motivation to avert uncertainty and fluctuations in the valuation by efficiently computing DeRDaVa scores upfront.

\subsection{Empirical Behaviours of Risk-DeRDaVa}

In this section, we observe the empirical behaviours of Risk-DeRDaVa and investigate how the valuation scores change as we use C-CVaR$^-$ at different levels $\alpha$, which reflects model owners with different risk attitudes. We assign a predetermined independent staying probability to each data source, where data sources with smaller indices have higher staying probability. As shown in Fig.~\ref{fig:rd-10-phoneme-logistic-shapley} and \ref{fig:rd-10-diabetes-nb-beta-16-4}, Risk-DeRDaVa (risk-averse) assigns even higher scores to data sources with high staying probability and penalizes data sources that are likely to delete harder. 

\section{Conclusion and Discussion}\label{sec:conclusion}

In this paper, we propose a deletion-robust data valuation technique DeRDaVa and an efficient approximation algorithm to improve its practicality. We also introduce Risk-DeRDaVa for model owners with different risk attitudes. We have shown both theoretically and empirically that our proposed solutions have more desirable properties than existing works when data deletion occurs. 
Future work can consider other possible applications (e.g., heuristics of active learning) and address the limitations and negative social impacts raised in App.~\ref{appendix:limitations-and-social-impacts} such as approximating DeRDaVa scores more efficiently with guarantees, estimating the staying probabilities $P_{\mathbf{D}}$ more accurately and preventing intentional misreporting of staying probabilities $P_{\mathbf{D}}$ or data.

\section*{Acknowledgments}

This research/project is supported by the National Research Foundation Singapore and DSO National Laboratories under the AI Singapore Programme (AISG Award No: AISG$2$-RP-$2020$-$018$). 
The author would like to extend special thanks to Dr. Henry Wai Kit Chia for his valuable comments and suggestions on this research.

\bibliography{aaai24}

\clearpage
\appendix
\onecolumn

\section{Fairness Axioms of Semivalues} \label{appendix:semivalue-axioms}

This section uses the same notations as in Sec.~\ref{subsec:semivalue-based-data-valuation}. \citet{dubey1981value} compiles four important axioms that every \textbf{fair} data valuation function must satisfy, which are defined below:
\begin{axiom}
    \textup{[Linearity]} Given a cooperative game $\langle D, v \rangle$ and any two model utility functions $v$ and $w$, a fair data valuation function $\phi^n$ shall satisfy 
    \begin{equation}\textstyle
        \forall d_i \in D \quad  [\phi^n_i(v) + \phi^n_i(w) = \phi^n_i(v + w)].
    \end{equation}
\end{axiom}

\noindent\textit{Fairness intuition.} 
\indentsquishlisttwo
    \item Let $v_1, v_2, \cdots, v_m$ be the $m$ model utility functions that involve the same ML model evaluated on $m$ validation sets\footnote{Recall from Sec. \ref{subsec:semivalue-based-data-valuation} that a model utility function $v$ takes in a coalition $S$ of data sources, uses $S$'s data to train a given ML model, and returns the validation score of the trained ML model evaluated on a given validation set.} with \textbf{same} size but \textbf{different} data (i.e., the valuation scores given by each validation set have equal credibility). The Linearity axiom guarantees that if we combine the $m$ validation sets into one and perform data valuation with the grand validation set, the valuation score we obtain should be equal to the average of the $m$ valuation scores.
    \item Consider a data source $\mathtt{NP}$ with an empty dataset. The existence of $\mathtt{NP}$ makes no impact on the utility of any coalition $S$ (i.e., its \textit{marginal contribution}\footnote{Marginal contribution is defined in Definition \ref{def:semivalue}.} $\mathrm{MaC}_v(d_i | S)$ is always zero). The Linearity axiom guarantees that data sources like $\mathtt{NP}$ are always assigned a valuation score of $0$.
\indentsquishend

\begin{axiom}
    \textup{[Dummy Player]} A data source $\mathtt{DP}$ is called a \textbf{dummy player} if its marginal contribution is always equal to its own utility. For any dummy player $\mathtt{DP}$, a fair data valuation function $\phi^n$ shall satisfy
    \begin{equation}\textstyle
        \phi^n_{\mathtt{DP}}(v) = v(\{\mathtt{DP}\}).
    \end{equation}
\end{axiom}

\noindent \textit{Fairness intuition.} 
\indentsquishlisttwo
    \item The \emph{Milnor's condition} \cite{milnor1952reasonable} states that $\phi^n_{i}(v)$ should lie between data source $d_i$'s minimum and maximum marginal contribution (i.e., $\min_{S \subseteq \dmax\setminus \{d_i\}} \mathrm{MaC}_v(d_i | S) \leq \phi^n_{i}(v) \leq \max_{S \subseteq \dmax\setminus \{d_i\}} \mathrm{MaC}_v(d_i | S)$). This is because if a data source's valuation score is less than its minimum marginal contribution, it is clearly underrated, vice versa. Since a dummy player $\mathtt{DP}$'s marginal contribution is always $v(\{\mathtt{DP}\})$, its valuation score should also be $v(\{\mathtt{DP}\})$ by Milnor's condition. 
    \item Consider a \textit{modular} cooperative game (i.e., each data source is contributing \textbf{independently}: $\forall d_i \in D_n \quad \forall S \subseteq D_n \setminus d_i \quad [v(S \cup \{d_i\}) = v(S) + v(\{d_i\})]$). In such a game, the Dummy Player axiom guarantees that each data source is valued by its own worth.
\indentsquishend

\begin{axiom}
    \textup{[Interchangeability/Symmetry]} Two data sources $d_i$ and $d_j$ are said to be \textbf{interchangeable} ($d_i \cong d_j$) if their marginal contributions to any coalition $S \subseteq D$ are always equal. The valuation scores assigned to any two interchangeable data sources shall be equal:
    \begin{equation}\textstyle
        d_i \cong d_j \Rightarrow \phi^n_i(v) = \phi^n_j(v).
    \end{equation}
\end{axiom}

\noindent \textit{Fairness intuition.} 
\indentsquishlisttwo
    \item This axiom ensures that ``equal'' data sources are treated equally.
\indentsquishend

\begin{axiom}
    \textup{[Monotonicity]} If model utility function $v$ is monotone increasing, then the valuation score assigned to any data source shall be non-negative:
    \begin{equation}\textstyle
        \forall S, T \subseteq \dmax \quad  [S \subseteq T \Rightarrow v(S) \leq v(T)] \Rightarrow \forall d_i \in \dmax \quad  [\phi^n_i(v) \geq 0].
    \end{equation}
\end{axiom}

\noindent \textit{Fairness intuition.} 
\indentsquishlisttwo
    \item If every data source makes a non-negative contribution, they should each get a non-negative valuation (or reward for collaborating).
\indentsquishend

In our paper, some of the above axioms become \textbf{no longer desirable} in our problem setting (e.g., see Fig.~\ref{fig:example} in the main paper ($2$ data sources) and Fig. \ref{fig:example-3-sources} ($3$ data sources) in App.~\ref{appendix:numerical-comparison}) and we address this by \textbf{``robustifying''} some of them. Refer to Axioms \ref{ax:robust-linearity}, \ref{ax:robust-dummy-player}, \ref{ax:robust-interchangeability} and \ref{ax:robust-monotonicity} in Sec. \ref{subsec:random-cooperative-game-and-robustified-fairness-axioms} for details.

Besides the axioms mentioned above, semivalue also has other properties that make it desirable, for example the two \textit{Desirability relations} \citep{carreras2000note}:
\begin{equation} \label{eqn:desirability-relation-1}
    \forall S \subseteq D_n \setminus \{d_i, d_j\} \quad [\mathrm{MaC}_v(d_i | S) \geq \mathrm{MaC}_v(d_j | S)] \Rightarrow \phi^n_i(v) \geq \phi^n_j(v);
\end{equation}
\begin{equation} \label{eqn:desirability-relation-2}
    \forall S \subseteq D_n \setminus \{d_i\} \quad [\mathrm{MaC}_v(d_i | S) \geq \mathrm{MaC}_w(d_i | S)] \Rightarrow \phi^n_i(v) \geq \phi^n_i(w).
\end{equation}

\noindent \textit{Fairness intuition.} 
\indentsquishlisttwo
    \item The first desirability relation (Eq. (\ref{eqn:desirability-relation-1})) guarantees that if data source $d_i$ always contributes more to any coalition than data source $d_j$, then $d_i$ should be valued higher than $d_j$.
    \item The second desirability relation (Eq. (\ref{eqn:desirability-relation-2})) guarantees that if under model utility function $v$ a data source $d_i$ always contributes more than under model utility function $w$, then the valuation score given by $v$ should also be larger than $w$.
    \item As such, these desirability relations ensure that ``unequal'' sources are treated unequally in the right direction.
\indentsquishend

\citet{shapley1953value} raises another important axiom for $n$-person games, which is the Efficiency axiom:
\begin{axiom}
    \textup{[Efficiency]} The sum of valuation scores assigned to every data source should be equal to their total utility: \begin{equation}\textstyle
        \sum\limits_{i=1}^n \phi^n_i(v) = v(D).
    \end{equation}
\end{axiom}

This axiom is desirable when the total utility is transferable (e.g., monetary profits). In the context of machine learning, the utility of each coalition usually refers to the performance of the model trained using data from the coalition, and there is no need to transfer it. Therefore, many research on data valuation do not take this axiom into consideration. 

\subsection{Numeric Comparison of Semivalue, DeRDaVa and Risk-DeRDaVa Scores }\label{appendix:numerical-comparison}

In this section, we present two simple numerical examples that demonstrate why some of the fairness axioms become undesirable when data deletion occurs, and compare the scores assigned by semivalue (e.g., Data Shapley), DeRDaVa and Risk-DeRDaVa.

\paragraph{{Numerical Example With 2 Data Sources (Fig.~\ref{fig:example})}}

We consider a $2$-source ($\bigstar$ and $\blacksquare$) data valuation problem to highlight the difference between Data Shapley and DeRDaVa valuation scores. $\bigstar$ always stays in the collaboration while $\blacksquare$ stays with probability $.3$: \squishlisttwo
    \item Data Shapley ignores potential data deletions and applies Eq.~\eqref{eqn:semivalue} to the \textcolor{red}{game with $\{\bigstar, \blacksquare\}$} as the support set. Data Shapley assigns equal score to both data sources as their marginal contributions are always equal.
    \item In contrast, DeRDaVa uses Eq.~\eqref{eqn:urdava} and considers different staying sets after anticipated data deletions. DeRDaVa sums the probability of a staying set (e.g., $.7$ for both staying) multiplied by the corresponding Data Shapley scores (e.g., $.4$ as calculated for the \textcolor{red}{game with $\{\bigstar, \blacksquare\}$}). $\bigstar$ receives a higher DeRDaVa score since it is more likely to stay.
    \item Risk-averse/seeking DeRDaVa considers the worst/best-cases staying sets, respectively. For example, when $\alpha = 0.3$, Risk-averse DeRDaVa only considers the \textcolor{blue}{game with $\{\bigstar\}$} as the staying set\footnote{However, note that the worst/best-cases staying sets do not necessarily correspond to the same game.}.
\squishend

\begin{figure}[!ht]
    \centering
    \includegraphics[width=\linewidth]{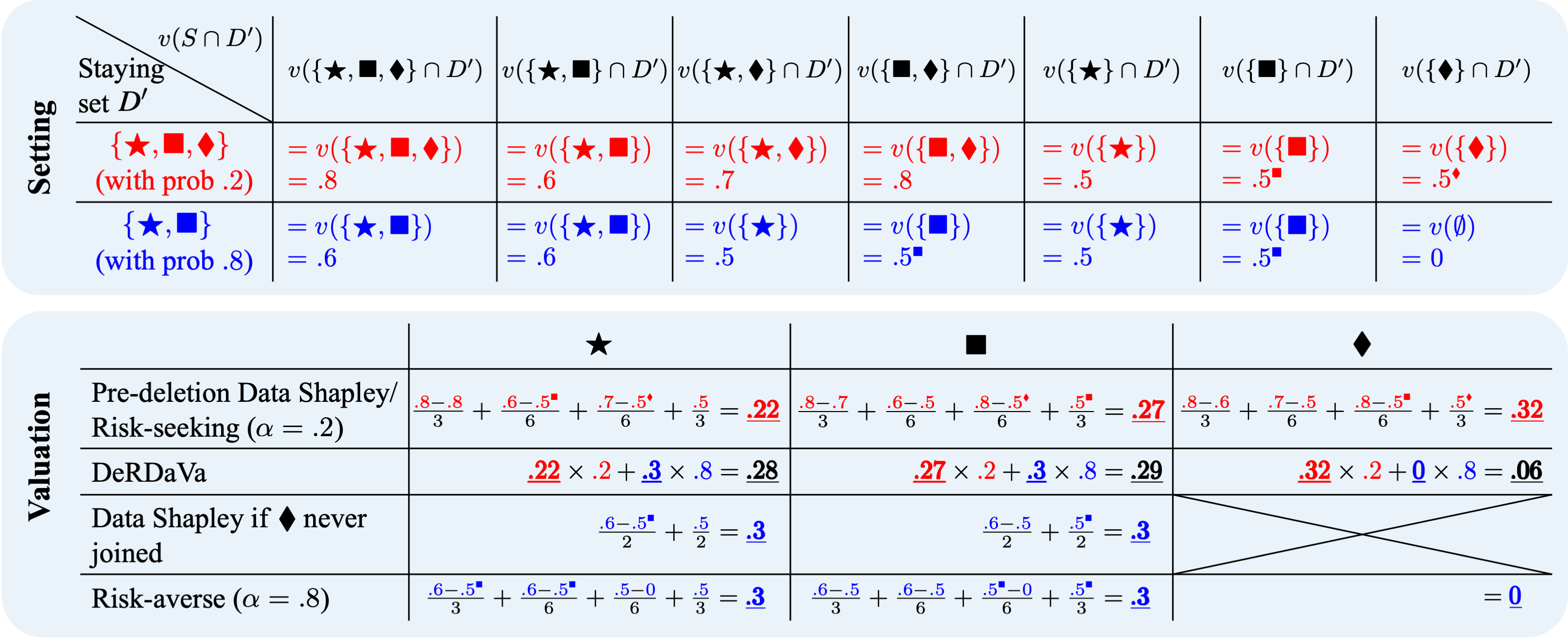}
    \caption{Another simple numerical comparison of Data Shapley vs. the deletion-robustified DeRDaVa and Risk-DeRDaVa scores (with Shapley prior). After the \textcolor{blue}{deletion of $\blacklozenge$},  $\bigstar$ and $\blacksquare$ are interchangeable players and should be assigned equal valuation score based on the Interchangeability axiom. However, they were assigned different pre-deletion Data Shapley scores due to the existence of $\blacklozenge$.
    In contrast, DeRDaVa anticipated deletions and assigned more similar scores.}
    \label{fig:example-3-sources}
\end{figure}

\paragraph{Numerical Example With 3 Data Sources (Fig.~\ref{fig:example-3-sources})}

We consider a $3$-source ($\bigstar$, $\blacksquare$ and $\blacklozenge$) data valuation problem to highlight the difference between Data Shapley and DeRDaVa valuation scores. $\bigstar$ and $\blacksquare$ always stay in the collaboration while $\blacklozenge$ stays with probability $.2$: \squishlisttwo
    \item Data Shapley ignores potential data deletions and applies Eq.~\eqref{eqn:semivalue} to the \textcolor{red}{game with $\{\bigstar, \blacksquare, \blacklozenge\}$} as the support set. Risk-seeking DeRDaVa with $\alpha=.2$ considers the best-case \textcolor{red}{game with $\{\bigstar, \blacksquare, \blacklozenge\}$} as the staying set, which gives the same scores to each data source as Data Shapley.
    \item In comparison, DeRDaVa uses Eq.~\eqref{eqn:urdava} and considers different staying sets after anticipated data deletions. DeRDaVa assigns higher values to $\bigstar$ and $\blacksquare$ than Data Shapley since they are more likely to stay.
    \item Risk-averse DeRDaVa considers the worst-case staying sets. For example, when $\alpha = 0.8$, Risk-averse DeRDaVa only considers the \textcolor{blue}{game with $\{\bigstar, \blacksquare\}$} as the staying set, and gives the same values as Data Shapley if $\blacklozenge$ never joined.
\squishend

\section{Proof of Theorem \ref{thm:npo-extension}} \label{appendix:proof-npo-extension}

Before starting the proof, we first give an example of how NPO-extension works. Let $\phi^5$ be the Shapley value for support set of size $5$, for which the weighting term $w_s \equiv \frac{1}{5}$ for any coalition size $s$. The tuple of weighting coefficients is thus \begin{equation}\textstyle
    \begin{split}
        \begin{pmatrix}
            w^5_0 & w^5_1 & w^5_2 & w^5_3 & w^5_4
        \end{pmatrix} & = \begin{pmatrix}
            \frac{1}{5} / \binom{4}{0} & \frac{1}{5} / \binom{4}{1} & \frac{1}{5} / \binom{4}{2} & \frac{1}{5} / \binom{4}{3} & \frac{1}{5} / \binom{4}{4}
        \end{pmatrix} \\ 
        & = \begin{pmatrix}
            \frac{1}{5} & \frac{1}{20} & \frac{1}{30} & \frac{1}{20} & \frac{1}{5}
        \end{pmatrix}.
    \end{split}    
\end{equation}

\begin{figure}[!ht]
    \centering
    \includegraphics[width=0.5\textwidth]{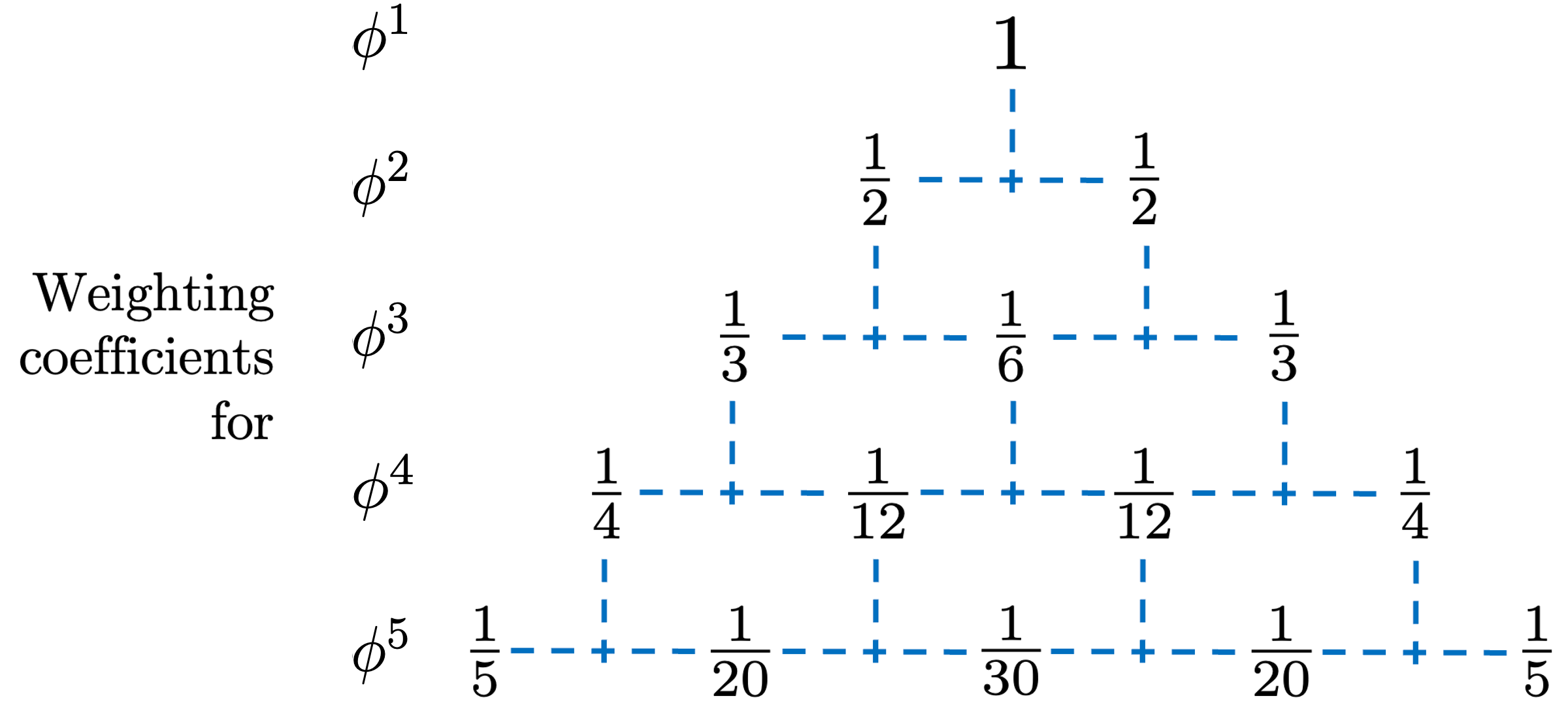}
    \caption{An example of NPO-extension from $\phi^5$.}
    \label{fig:npo-extension-example}
\end{figure}

The NPO-extension process to extend $\phi^5$ to the sequence $\Phi = \langle\phi^1, \phi^2, \phi^3, \phi^4, \phi^5\rangle$ is illustrated in Fig.~\ref{fig:npo-extension-example}. For example, in the extended semivalue for support set of size $3$, $\phi^3$, the tuple of weighting coefficients is \begin{equation}\textstyle
    \begin{pmatrix}
        w^3_0 & w^3_1 & w^3_2
    \end{pmatrix} = \begin{pmatrix}
        \frac{1}{3} & \frac{1}{6} & \frac{1}{3}
    \end{pmatrix}.  
\end{equation}

We first prove that each $\phi^k$ in the extended sequence $\Phi = \langle \phi^k : k = 1, 2, \cdots, n \rangle$ is indeed a semivalue. This can be done by induction starting from $\phi^n$. Suppose $\phi^k$ is a semivalue, by Definition \ref{def:semivalue} we have \begin{equation}\textstyle
    \sum\limits_{s = 0}^{k - 1} w^k_s \cdot \binom{k - 1}{s} = 1.
\end{equation}
To prove that $\phi^{k-1}$ is a semivalue, it suffices to show that \begin{equation}\textstyle
    \sum\limits_{s = 0}^{k - 2} w^{k-1}_s \cdot \binom{k-2}{s} = 1.
\end{equation}
We have \begin{equation}\textstyle
    \begin{split}
        \mathrm{LHS} & = \sum\limits_{s = 0}^{k - 2} \left(w^k_s + w^k_{s+1}\right) \cdot \binom{k-2}{s} \\
        & = \sum\limits_{s = 0}^{k - 2} w^k_s \cdot \binom{k-2}{s} + \sum\limits_{s = 1}^{k - 1} w^k_s \cdot \binom{k-2}{s - 1} \\
        & = \sum\limits_{s = 1}^{k - 2} w^k_s \cdot \left[\binom{k-2}{s} + \binom{k-2}{s - 1}\right] + w^k_0 + w^k_{k - 1} \\
        & = \sum\limits_{s = 1}^{k - 2} w^k_s \cdot \binom{k - 1}{s} + w^k_0 \cdot \binom{k - 1}{0} + w^k_{k-1} \cdot \binom{k-1}{k-1} \text{ (by Pascal's identity \citep{merris2003combinatorics})} \\
        & = \sum\limits_{s = 0}^{k - 1} w^k_s \cdot \binom{k - 1}{s} \\
        & = 1 = \mathrm{RHS}.
    \end{split}
\end{equation}

We shall then prove that $\Phi$ is NPO-consistent. \citet{derks1999null} states that if any two data valuation functions $\phi^k$ and $\phi^{k - 1}$ both satisfy the Linearity axiom and assign $0$ score to any null player, then they are NPO-consistent if and only if $w^{k - 1}_s = w^k_s + w^k_{s+1}$ for $s = 0, 1, \cdots, k - 2$. Since every $\phi^k \in \Phi$ is a semivalue, it satisfies the Linearity axiom and assigns $0$ score to any null player since the weighted sum of $0$ is $0$. From the procedures of NPO-extension, it is thus clear that $\Phi$ is NPO-consistent.

The uniqueness of $\Phi$ from any $\phi^n$ is a direct result of Recursion theorem \citep{kleene1938notation}. Therefore, we have completed our proof for Theorem \ref{thm:npo-extension}.

\section{Proof of Theorem \ref{thm:urdava}} \label{appendix:proof-urdava}

We first verify that the DeRDaVa function $\tau$ indeed satisfies Axioms \ref{ax:robust-linearity}, \ref{ax:robust-dummy-player}, \ref{ax:robust-interchangeability} and \ref{ax:robust-monotonicity}: \begin{enumerate}
    \item For any two model utility functions $v$ and $w$, we have \begin{equation}\textstyle
        \begin{split}
            \tau_i (v + w) & = \mathbb{E}_{\mathbf{D} \sim P_{\mathbf{D}}} \left[\mathbb{I}[d_i \in \mathbf{D}] \cdot \phi^{|\mathbf{D}|}_i (v + w)\right] \\
            & = \mathbb{E}_{\mathbf{D} \sim P_{\mathbf{D}}} \left[\mathbb{I}[d_i \in \mathbf{D}] \cdot \left(\phi^{|\mathbf{D}|}_i (v) + \phi^{|\mathbf{D}|}_i (w)\right)\right] \\
            & = \mathbb{E}_{\mathbf{D} \sim P_{\mathbf{D}}} \left[\mathbb{I}[d_i \in \mathbf{D}] \cdot \phi^{|\mathbf{D}|}_i (v)\right] + \\
            & \phantom{{} = {}} \mathbb{E}_{\mathbf{D} \sim P_{\mathbf{D}}} \left[\mathbb{I}[d_i \in \mathbf{D}] \cdot \phi^{|\mathbf{D}|}_i (w)\right] \\
            & = \tau_i (v) + \tau_i (w).
        \end{split}
    \end{equation}
    Therefore, Axiom \ref{ax:robust-linearity} (Robust Linearity) holds.
    \item For any dummy player $\mathtt{DP}$, we have \begin{equation}\textstyle
        \begin{split}
            \tau_{\mathtt{DP}}(v) & = \mathbb{E}_{\mathbf{D} \sim P_{\mathbf{D}}} \left[\mathbb{I}[\mathtt{DP} \in \mathbf{D}] \cdot \phi^{|\mathbf{D}|}_{\mathtt{DP}}(v)\right] \\
            & = \mathbb{E}_{\mathbf{D} \sim P_{\mathbf{D}}} \left[\mathbb{I}[\mathtt{DP} \in \mathbf{D}] \cdot v(\{\mathtt{DP}\})\right].
        \end{split}
    \end{equation}
    Therefore, Axiom \ref{ax:robust-dummy-player} (Robust Dummy Player) holds.
    \item For any two robustly interchangeable data sources $d_i \cong d_j$, we have for any $D'' \subseteq \dmax \setminus \{d_i, d_j\}$, \begin{equation}\textstyle
        P_{\mathbf{D}}(\mathbf{D} = D'' \cup \{d_i\}) = P_{\mathbf{D}}(\mathbf{D} = D'' \cup \{d_j\})
    \end{equation}
    based on the definition of robust interchangeability. By symmetry we have \begin{equation}\textstyle
        \tau_i(v) = \tau_j(v).
    \end{equation}
    Therefore, Axiom \ref{ax:robust-interchangeability} (Robust Interchangeability) holds.
    \item If model utility function $v$ is monotone increasing, then for any data source $d_i$ we have \begin{equation}\textstyle
        \tau_i(v) = \mathbb{E}_{\mathbf{D} \sim P_{\mathbf{D}}} \left[\mathbb{I}[d_i \in \mathbf{D}] \cdot \phi^{|\mathbf{D}|}_i (v)\right].
    \end{equation}
    Since $\mathbb{I}[d_i \in \mathbf{D}] \geq 0$ and $\phi^{|\mathbf{D}|}_i (v) \geq 0$ because of monotonicity of semivalues, the expectation $\tau_i(v)$ is also non-negative. Therefore, Axiom \ref{ax:robust-monotonicity} (Robust Monotonicity) holds.
\end{enumerate}

We shall then prove the uniqueness of $\tau$. Let $V(S) = v(S \cap \mathbf{D})$ be the \textit{random utility function} through which we can shift our uncertainty in support set $\mathbf{D}$ to uncertainty in utility function $V$. Let $\mathbb{E}[V(S)] = \mathbb{E}_{\mathbf{D} \sim P_{\mathbf{D}}}[v(S \cap \mathbf{D})]$, which represents the expected utility of coalition $S$. Consider the static cooperative game $\langle \dmax, \mathbb{E}[V(\cdot)] \rangle$. It is not hard to see that for any given $P_\mathbf{D}$, there exists a bijection between the random cooperative game $\langle \mathbf{D}, v \rangle$ and the static cooperative game $\langle \dmax, \mathbb{E}[V(\cdot)] \rangle$. We call $\langle \dmax, \mathbb{E}[V(\cdot)] \rangle$ the \textit{static dual} game to $\langle \mathbf{D}, v \rangle$. 

We first prove that any general solution to the random cooperative game $\langle \mathbf{D}, v \rangle$ that satisfies the four robustified axioms, $\tau$, is a solution to its static dual game $\langle \dmax, \mathbb{E}[V(\cdot)] \rangle$ that satisfies the four original semivalue axioms in App. \ref{appendix:semivalue-axioms}: \begin{enumerate}
    \item Axiom \ref{ax:robust-linearity} (Robust Linearity) implies that the solution has to be a linear combination of $v(S)$ for all $S \subseteq \dmax$. Since the expectation operator $\mathbb{E}[\cdot]$ is linear and a linear combination of linear functions is linear, we have that $\tau$ is linear in $\langle \dmax, \mathbb{E}[V(\cdot)] \rangle$. In other words, $\tau$ is a linear combination of $\mathbb{E}[V(S)]$ for all $S \subseteq \dmax$.  
    \item Any dummy player $\mathtt{DP}$ in $\langle \mathbf{D}, v \rangle$ is also a dummy player in $\langle \dmax, \mathbb{E}[V(\cdot)] \rangle$ and vice versa, since for any coalition $S \subseteq \dmax \setminus \{\mathtt{DP}\}$ we have
    \begin{equation}\textstyle
        \begin{split}
            \mathbb{E}[V(S \cup \mathtt{DP})] - \mathbb{E}[V(S)] &= \mathbb{E}_{\mathbf{D} \sim P_{\mathbf{D}}}[v((S \cup \{\mathtt{DP}\}) \cap \mathbf{D})] - \mathbb{E}_{\mathbf{D} \sim P_{\mathbf{D}}}[v(S \cap \mathbf{D})] \\
            &= \mathbb{E}_{\mathbf{D} \sim P_{\mathbf{D}}}[v((S \cup \{\mathtt{DP}\}) \cap \mathbf{D}) - v(S \cap \mathbf{D})] \\
            &= \mathbb{E}_{\mathbf{D} \sim P_{\mathbf{D}}}[v(\{\mathtt{DP}\} \cap \mathbf{D})] \\
            &= \mathbb{E}[V(\{\mathtt{DP}\})].
        \end{split}
    \end{equation}
    Moreover, Axiom \ref{ax:robust-dummy-player} (Robust Dummy Player) ensures that $\tau_{\mathtt{DP}}(v) = \mathbb{E}[V(\{\mathtt{DP}\})]$.
    \item For any two robustly interchangeable data sources $d_i \cong d_j$ in $\langle \mathbf{D}, v \rangle$, they are also interchangeable in $\langle \dmax, \mathbb{E}[V(\cdot)] \rangle$ since for any $S$ \begin{equation}\textstyle
        \begin{split}
            \mathbb{E}[V(S \cup d_i)] &= \mathbb{E}_{\mathbf{D} \sim P_{\mathbf{D}}}[v((S \cup \{d_i\}) \cap \mathbf{D})] \\
            &= \mathbb{E}_{\mathbf{D} \sim P_{\mathbf{D}}}[v((S \cup \{d_j\}) \cap \mathbf{D})] \\
            &= \mathbb{E}[V(S \cup \{d_j\})].
        \end{split}
    \end{equation}
    Moreover, Axiom \ref{ax:robust-interchangeability} states that the valuation scores assigned to any two interchangeable players shall be equal.
    \item The monotonicity of $\tau$ is directly stated in Axiom \ref{ax:robust-monotonicity} (Robust Monotonicity).
\end{enumerate} 

Since semivalues are the only solutions to a static game given the four semivalue axioms and the particular choice of semivalues is restricted by the jointly-agreed semivalue $\phiopt$, the uniqueness of our solution thus follows from the uniqueness of semivalues and static duals.

\section{Efficient Approximation of DeRDaVa}\label{appendix:efficient-approximation-of-derdava}

\subsection{Approximation via Monte-Carlo Sampling} \label{appendix:mc-sampling}

As illustrated in the main paper, DeRDaVa scores can be efficiently approximated via Monte-Carlo sampling when it is easy to draw samples of staying set $D'$ from probability distribution $P_\mathbf{D}$. This method works precisely because DeRDaVa score is the expectation of marginal contributions over some suitable distribution of staying set $D'$ and coalition $S$.

The following theorem provides the approximation guarantee:

\begin{theorem}
    \textup{[Approximation guarantee for Monte-Carlo sampling]} Let $r$ be the range of model utility function $v$, $\tau = (\tau_1(v), \tau_2(v), \cdots, \tau_n(v))$ be the actual DeRDaVa scores, and $\Hat{\tau}$ be the estimator of $\tau$ via Monte-Carlo sampling. To ensure
    \begin{equation}
        \mathrm{Pr}\left[\|\Hat{\tau} - \tau\|_{\infty} \geq \epsilon\right] \leq \delta,
    \end{equation}
    we need at least $O\left(\frac{2r^2 n}{\epsilon^2} \log \frac{2n}{\delta}\right)$ samples.
\end{theorem}

\noindent
\textit{Proof.} Let $T$ be the number of samples used in Monte-Carlo sampling. Since each sample gives the marginal contribution of $1$ data source, there are $T/n$ samples of marginal contributions of each data source. By Hoeffding's inequality,
\begin{equation}
    \mathrm{Pr}\left[\|\Hat{\tau} - \tau\|_{\infty} \geq \epsilon\right] \leq 2n \exp{\left(- \frac{2(T/n) \epsilon^2}{4 r^2}\right)}.
\end{equation}
Let $2n \exp{\left(- \frac{2(T/n) \epsilon^2}{4 r^2}\right)} \leq \delta$, we have
\begin{equation}
    T \geq \frac{2r^2}{\epsilon^2} \log \frac{2n}{\delta}.
\end{equation}

\subsection{Approximation via 012-MCMC Algorithm} \label{appendix:012-mcmc}

We first justify 012-MCMC algorithm. We rewrite the formula for DeRDaVa in Definition \ref{def:urdava} as follows:
\begin{equation}
    \begin{split}
        \tau_i(v) & = \sum\limits_{D' \subseteq \dmax} \left(P_{\mathbf{D}}(\mathbf{D} = D') \cdot \mathbb{I}[d_i \in D'] \cdot \sum\limits_{S \subseteq D' \setminus \{d_i\}} w^{|D'|}_{|S|} [v(S \cup \{d_i\}) - v(S)]\right) \\
        & = \sum\limits_{D' \subseteq \dmax} \sum\limits_{S \subseteq D' \setminus \{d_i\}} \left(P_{\mathbf{D}}(\mathbf{D} = D') \cdot \mathbb{I}[d_i \in D'] \cdot w^{|D'|}_{|S|} \cdot [v(S \cup \{d_i\}) - v(S)]\right) \\
        & = \sum\limits_{D'' \subseteq \dmax \setminus \{d_i\}} \sum\limits_{S \subseteq D''} \left(P_{\mathbf{D}}(\mathbf{D} = D'' \cup \{d_i\}) \cdot w^{|D''| + 1}_{|S|} \cdot [v(S \cup \{d_i\}) - v(S)]\right).
    \end{split}
\end{equation}

Note that the last step is done by considering only the case where the indicator variable $\mathbb{I}[d_i \in D']$ equates to $1$. The validity of Eq. (\ref{eqn:urdava-sampling}) then follows from the rewritten formula through importance sampling from a uniform distribution of the pair $\{S, D''\}$. 

An $(\epsilon, \delta)$-approximation guarantee can be obtained by directly applying Hoeffding's inequality on Eq. (\ref{eqn:urdava-sampling}):

\begin{theorem}
    \textup{[Approximation guarantee for 012-MCMC algorithm]} Let $r$ be the range of model utility function $v$, $c$ be the maximum coefficient to the marginal contribution\footnote{This depends on the actual choice of probability distribution $P_{\mathbf{D}}$ and prior semivalue $\phi^n$.} (i.e., $c = \max\limits_{D'_{(t)}, S'_{(t)}} \frac{P_\mathbf{D}\left(\mathbf{D} = D'_{(t)} \right)}{1/3^{(n-1)}} \cdot w^{|D'_{(t)}|}_{|S_{(t)}|}$), $\tau = (\tau_1(v), \tau_2(v), \cdots, \tau_n(v))$ be the actual DeRDaVa scores, and $\Hat{\tau}$ be the estimator of $\tau$ via Monte-Carlo sampling. To ensure
    \begin{equation}
        \mathrm{Pr}\left[\|\Hat{\tau} - \tau\|_{\infty} \geq \epsilon\right] \leq \delta,
    \end{equation}
    we need at least $O\left(\frac{2c^2 r^2 n}{\epsilon^2} \log \frac{2n}{\delta}\right)$ samples.
\end{theorem}

Although this might be a larger bound than that of plain Monte-Carlo sampling (it might still be exponential but with a smaller base than exact computation), 012-MCMC algorithm allows us to take advantage of importance sampling, especially when the actual distribution of staying set $D'$ and coalition $S$ is hard to realize. Also note that when $n$ is large, the quantities $P_\mathbf{D}\left(\mathbf{D} = D'_{(t)}\right)$ and $w^{|D'_{(t)}|}_{|S_{(t)}|}$ are significantly small, thus 
$c$ is significantly smaller than $O(3^n)$. 

As described in Sec.~\ref{subsec:urdava-and-its-efficient-approximation}, the 012-MCMC algorithm consists of two parts: 012-sampling and MCMC-sampling. The pseudocode for each part is given in Algorithm \ref{algo:zero-one-two-sampling} and \ref{algo:mcmc}, respectively.

\begin{algorithm}[!ht]
\caption{012-sampling algorithm for efficient generation of samples.}
\label{algo:zero-one-two-sampling}
\hspace*{\algorithmicindent} \textbf{Input:} Set of all data sources $\dmax$. \\
\hspace*{\algorithmicindent} \textbf{Output:} Random sample of two sets of data sources $D''$ and $S$ where $S \subseteq D'' \subseteq \dmax \setminus \{d_i\}$. \vspace{2mm}
\begin{algorithmic}[1]
\State $D'' \gets \emptyset$
\State $S \gets \emptyset$
\For {$d_i$ \textbf{in} $D_{\text{max}}$} 
    \State $o \gets \text{randomly pick from $\{0, 1, 2\}$}$
    \If {$o == 0$}
        \State Do nothing
    \ElsIf {$o == 1$}
        \State Add $d_i$ to $D''$
    \ElsIf {$o == 2$}
        \State Add $d_i$ to both $D''$ and $S$
    \EndIf
\EndFor
\State \Return $D'', S$
\end{algorithmic}
\end{algorithm}

\begin{algorithm}
\caption{MCMC algorithm for efficient approximation of DeRDaVa scores.}
\hspace*{\algorithmicindent} \textbf{Input:} Set of all data sources $\dmax$; model utility function $v$; staying probability distribution \\
\hspace*{19mm} $P_{\mathbf{D}}$; batch size $b$; convergence threshold $\rho$. \\
\hspace*{\algorithmicindent} \textbf{Output:} Approximated DeRDaVa scores $\tau_i(v)$ of each data source $d_i, i = 1, 2, \ldots, n$. \vspace{2mm}
\begin{algorithmic}[1]
\State $\hat\rho \gets 2$
\State Initialize $M$ Markov chains in parallel
\While {$\hat\rho \geq \rho$}
    \For {$j = 1, 2, \ldots, b$}
        \State Generate $M$ random samples $\langle D''_m, S_m \rangle, m = 1, 2, \cdots, M$ with 012-sampling algorithm
        \For {each random sample $\langle D''_m, S_m \rangle$}
            \State Substitute $D''_m, S_m, v, P_{\mathbf{D}}$ into Eq. (\ref{eqn:urdava-sampling}) to get the terms of summation
            \State Append the value to Chain $m$
        \EndFor
    \EndFor
    \State $\hat\rho \gets$ Gelman-Rubin statistic of the current Markov chains
    \State Update DeRDaVa scores based on the new samples
\EndWhile
\State \Return DeRDaVa scores of each data source
\end{algorithmic}
\label{algo:mcmc}
\end{algorithm}

\section{C-CVaR and Risk-DeRDaVa} \label{appendix:non-additivity-of-risk-urdava}

\subsection{Formal Definition of C-CVaR}

The following is a formal definition of C-CVaR$^\mp$ at level $\alpha$. In particular, the fraction $\lambda$ is used to ``split'' the particular value where $\alpha$ falls inside, for instance the realization $V(S) = 3$ in Fig.~\ref{img:risk-averse}.

\begin{definition} \label{def:ccvar}
    \textup{[C-CVaR]} Given a random utility function $V$ and a coalition $S$, define for any $\alpha \in [0, 1]$
    \begin{equation}\textstyle
        \begin{split}
            z_\alpha[V(S)] = \sup\{z : \mathrm{Pr}[V(S) < z] \leq \alpha\}; \\
            E^-_{z_\alpha}[V(S)] = \mathbb{E}[V(S) | V(S) < z_\alpha[V(S)]].
        \end{split}
    \end{equation}
    Then the \textbf{Risk-Averse Coalitional Conditional Value-at-Risk at level $\pmb\alpha$} (C-CVaR$^-_\alpha$) is defined as \begin{equation}\textstyle
        \textup{C-CVaR}^-_\alpha[V(S)] = \lambda \cdot E^-_{z_\alpha}[V(S)] + (1 - \lambda) \cdot z_\alpha[V(S)],
    \end{equation}
    where \begin{equation}\textstyle
        \lambda = \frac{\mathrm{Pr}[V(S) < z_\alpha[V(S)]]}{\alpha}.
    \end{equation}
    The \textbf{Risk-Seeking Coalitional Conditional Value-at-Risk at level $\pmb\alpha$} (C-CVaR$^+_\alpha$) is defined as \begin{equation}\textstyle
        \textup{C-CVaR}^+_\alpha[V(S)] = -\textup{C-CVaR}^-_\alpha[-V(S)].
    \end{equation}
\end{definition}

\subsection{Non-Additivity of C-CVaR}

C-CVaR has the following property:

\begin{theorem}
    \textup{[Non-Additivity]} A function $f$ is said to be super-additive if $\forall x, y \quad  [f(x + y) \geq f(x) + f(y)]$ and it is said to be sub-additive if $\forall x, y \quad  [f(x + y) \leq f(x) + f(y)]$. \textup{C-CVaR}$^-$ is \textbf{super-additive} whereas \textup{C-CVaR}$^+$ is \textbf{sub-additive}. 
\end{theorem}

This theorem directly follows from the sub-additivity of CVaR \citep{yamout2007comparison}. Therefore, we cannot move the C-CVaR$_\alpha$ operator to the outside of the summation sign in Eq. (\ref{eqn:risk-urdava}), and hence cannot repeat the sampling procedures for DeRDaVa. We need to combine Monte-Carlo sampling of $S$ and Monte-Carlo CVaR algorithm \citep{hong2014monte} as described in the main paper.

\section{More Experimental Details and Results} \label{appendix:experimental-details}

We run our experiments on two machines: \begin{enumerate}
    \item Ubuntu 18.04.5 LTS, Intel(R) Xeon(R) Gold 6226R (2.90GHz); 
    \item Ubuntu 20.04.3 LTS, Intel(R) Xeon(R) Gold 6226R (2.90GHz).
\end{enumerate}
The use of GPUs is not required. All the programs are written in Python and executed using Anaconda. The program repository can be found at \texttt{https://github.com/snoidetx/derdava}. Refer to \texttt{requirements.txt} for the list of Python packages used and DeRDaVa's user guide for a detailed description of our codes. 

Tab.~\ref{tab:datasets} shows a summary of all datasets and respective models used in this research. The specific hyperparameters used in each model can be found in the codes.

\begin{table}[!b]
\begin{center}
\begin{tabular}{||c c c c c||} 
 \hline
 \textbf{Dataset} & \textbf{Size} & \textbf{Dimension} & \textbf{Source} & \textbf{Model(s)}\\ [0.5ex] 
 \hline\hline
 \texttt{CreditCard} & 30000 & 23 & \cite{yeh2009the} & Naïve Bayes, Support Vector Machine\\ 
 \hline
 \texttt{Diabetes} & 768 & 8 & \cite{carrion2022pima} & Naïve Bayes, Support Vector Machine \\
 \hline
 \texttt{Phoneme} & 5404 & 5 & \cite{grin2022phoneme}  & Logistic Regression, Support Vector Machine\\
 \hline
 \texttt{Pol} & 15000 & 48 & \cite{vanschoren2014pol} & Ridge Classifier \\
 \hline
 \texttt{Wind} & 6574 & 11 & \cite{vanschoren2014wind} &Naïve Bayes \\
 \hline
\end{tabular}
\end{center}
\caption{Summary of datasets and respective models used in this research.}
\label{tab:datasets}
\end{table}

\begin{figure}[!b]
    \centering
    \begin{subfigure}[b]{0.24\textwidth}
        \centering
        \includegraphics[width=\textwidth]{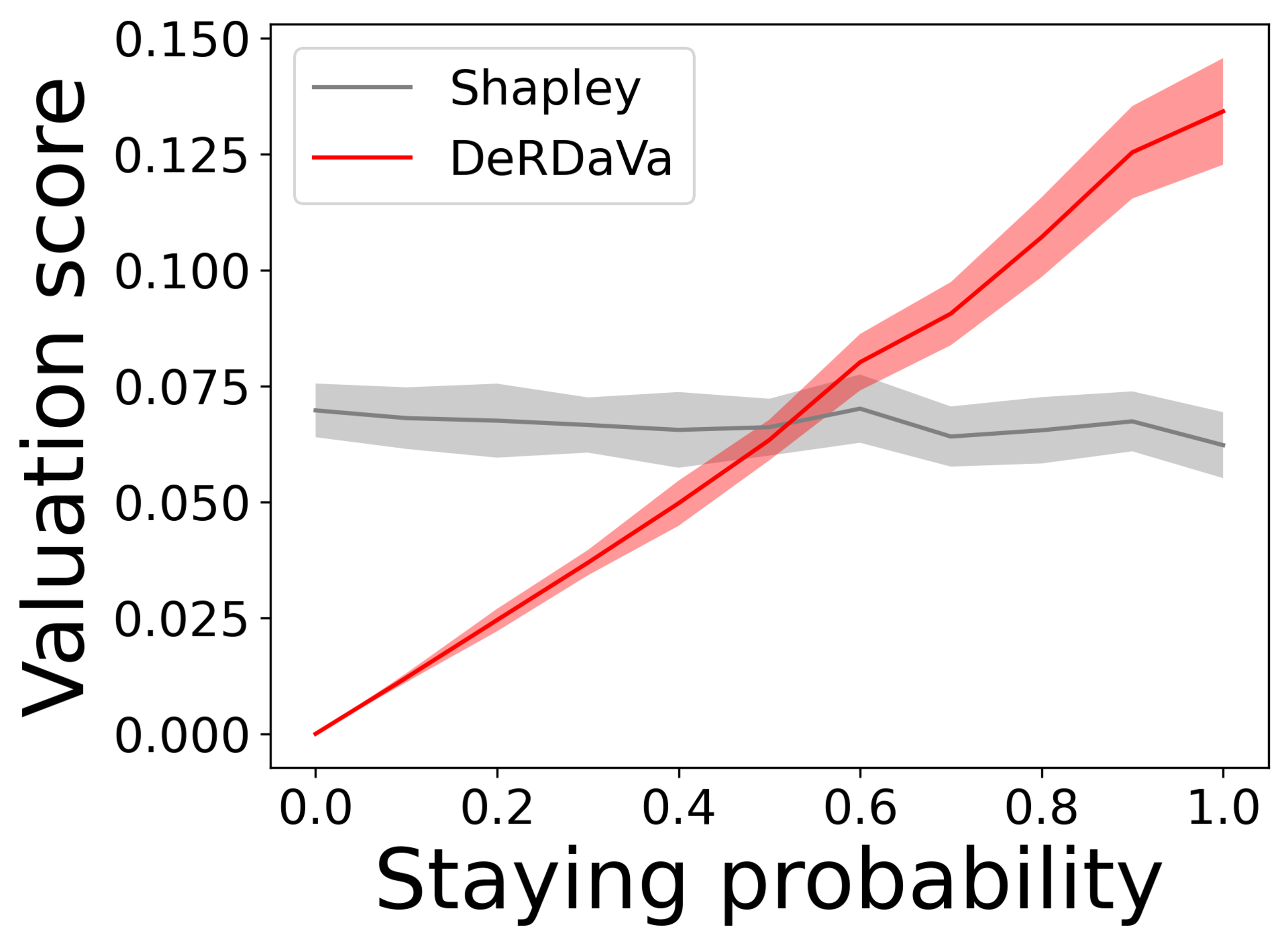}
        \caption{[LR-Pm].}\label{fig:sp-10-phoneme-lr-shapley}
    \end{subfigure}
    \begin{subfigure}[b]{0.24\textwidth}
        \centering
        \includegraphics[width=\textwidth]{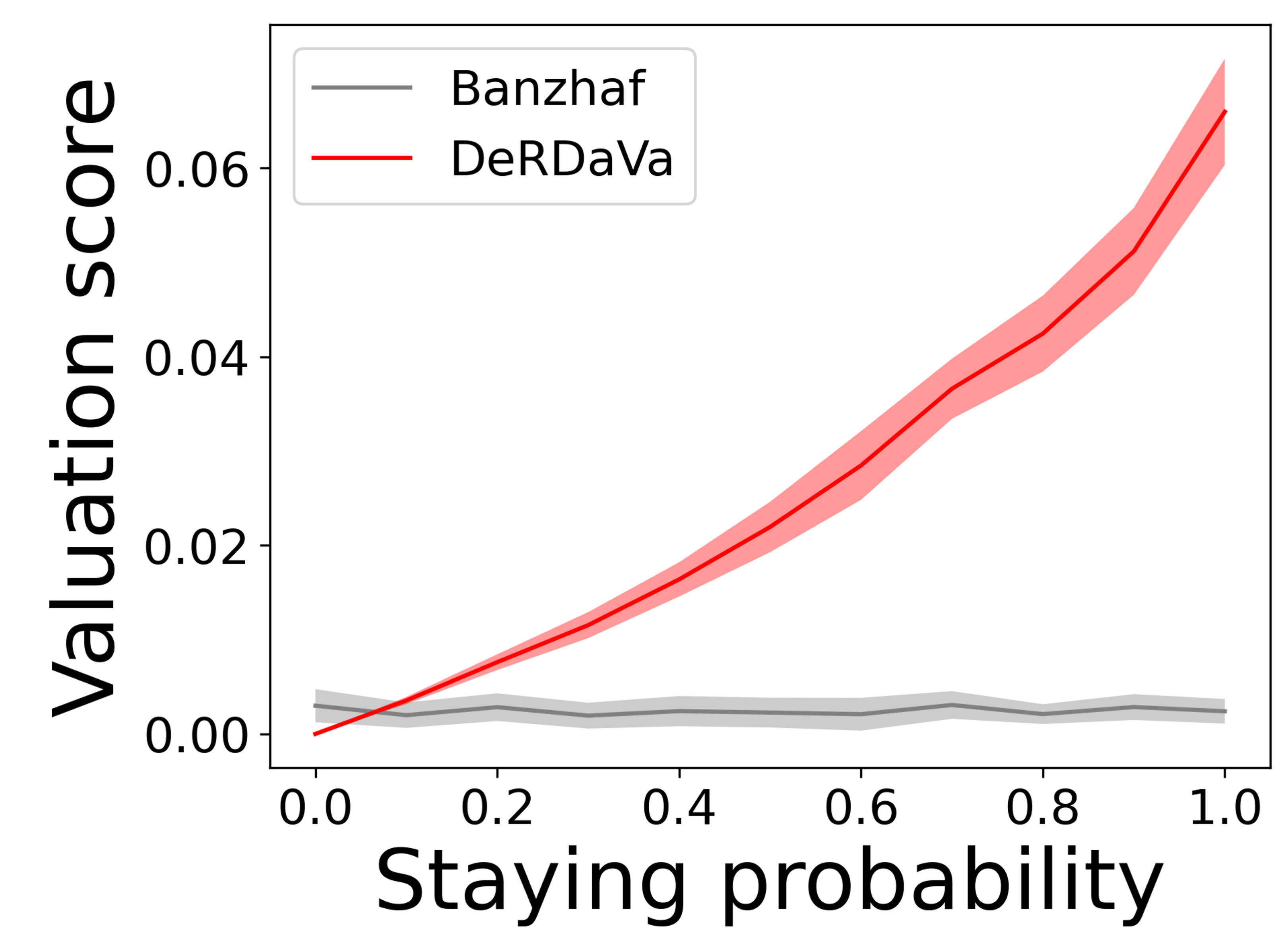}
        \caption{[NB-Wd].}\label{fig:sp-10-wind-nb-banzhaf}
    \end{subfigure}
    \begin{subfigure}[b]{0.24\textwidth}
        \centering
        \includegraphics[width=\textwidth]{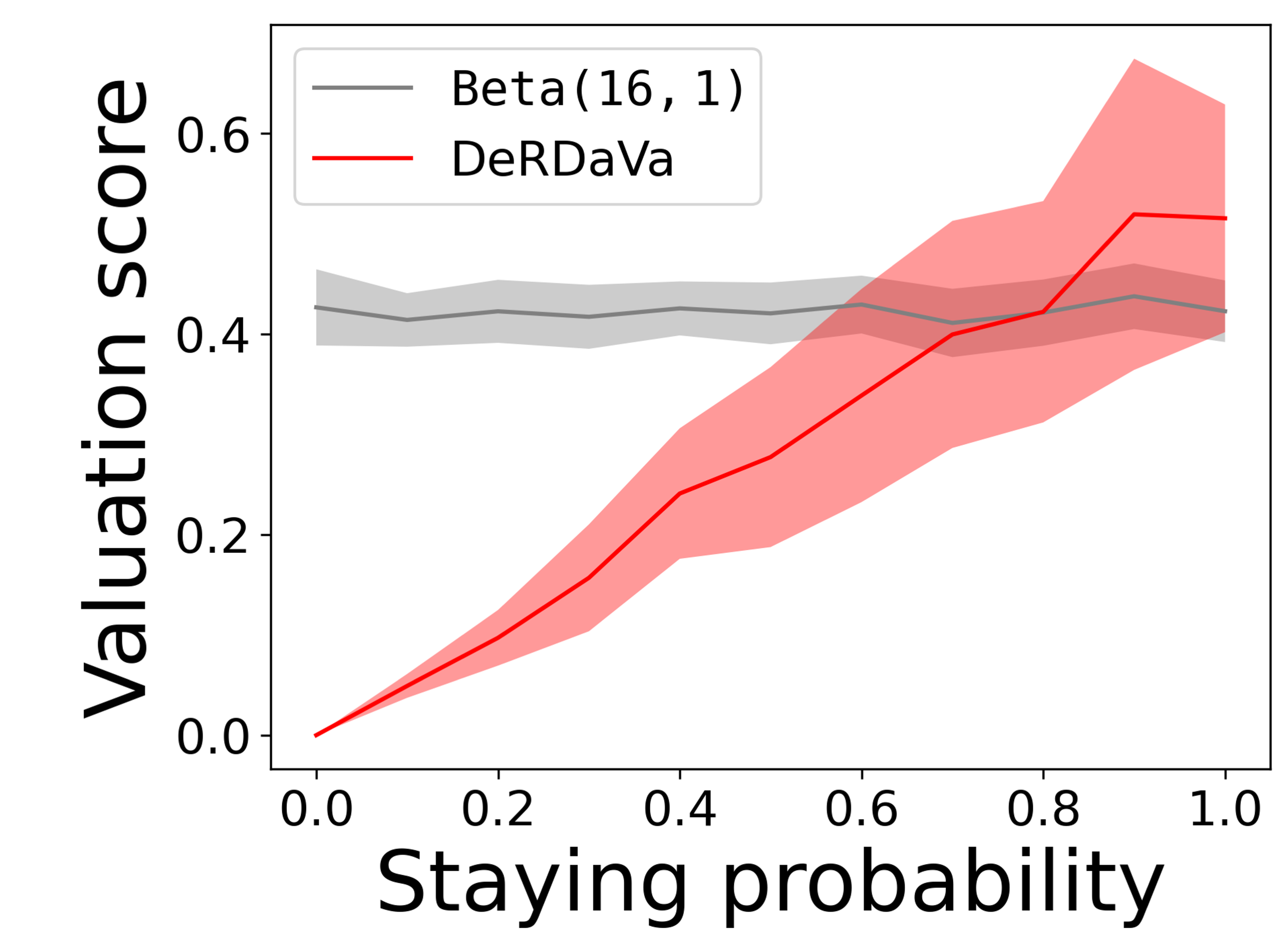}
        \caption{[SVM-Pm].}\label{fig:sp-10-phoneme-svm-beta-16-1}
    \end{subfigure}
    \begin{subfigure}[b]{0.24\textwidth}
        \centering
        \includegraphics[width=\textwidth]{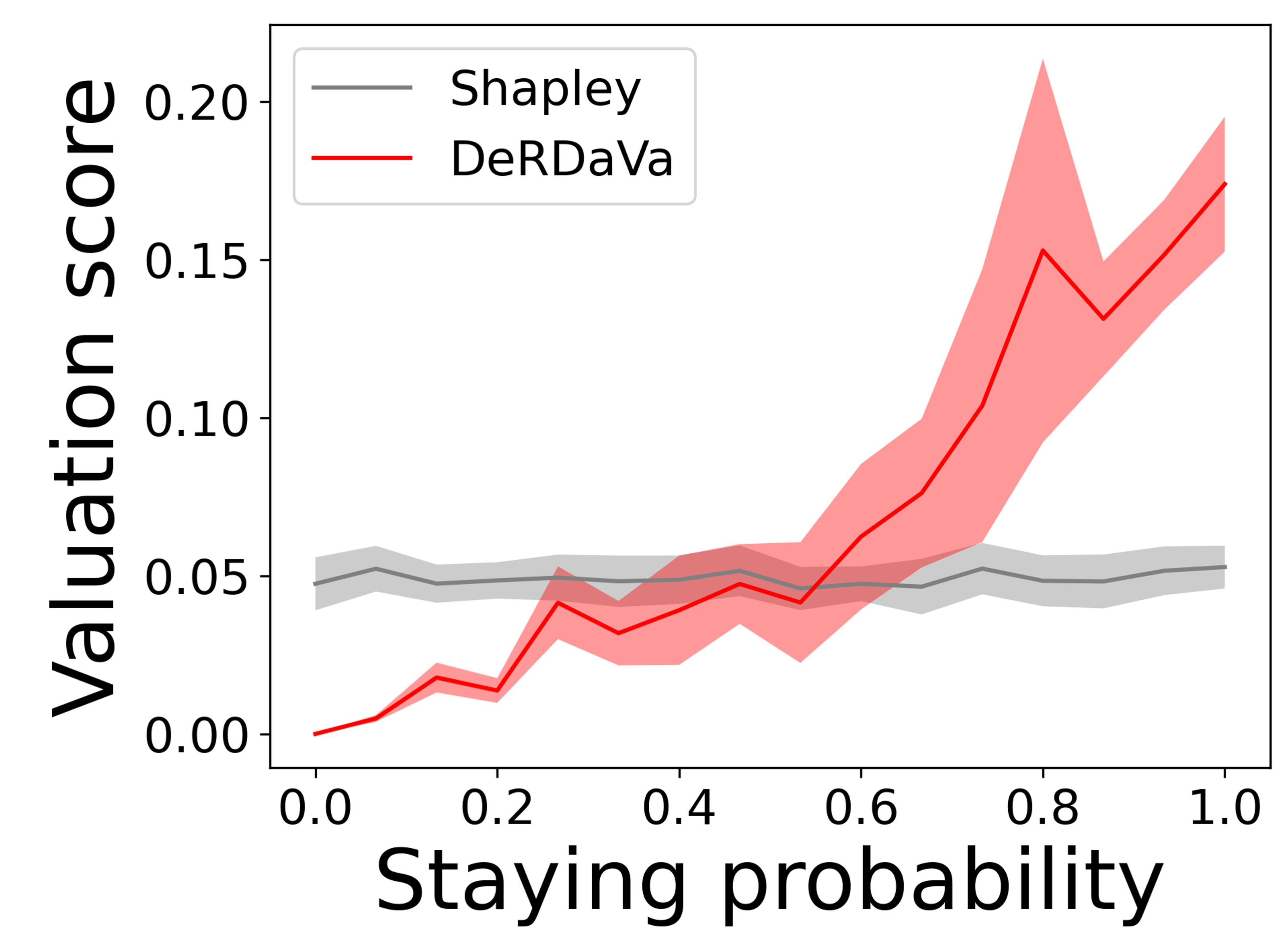}
        \caption{[RC-Po].}\label{fig:sp-15-pol-rc-shapley}
    \end{subfigure}
    \caption{Additional experiments on staying probability. (\ref{fig:sp-10-phoneme-lr-shapley}) uses Shapley prior and $11$ data sources; (\ref{fig:sp-10-wind-nb-banzhaf}) uses Banzhaf prior and $11$ data sources; (\ref{fig:sp-10-phoneme-svm-beta-16-1}) uses $\mathtt{Beta(16, 1)}$ prior and $11$ data sources; (\ref{fig:sp-15-pol-rc-shapley}) uses Ridge Classifier (RC) trained on Pol \citep{vanschoren2014pol}, Shapley prior and $16$ data sources.}
    \label{fig:appendix-contribution-to-model-performance-and-model-robustness}
\end{figure}

\subsection{Measure of Contribution to Model Performance and Deletion Robustness}

\paragraph{Staying Probability}

To ensure that each data source has a significantly large marginal contribution, we randomly allocate $5$ or $10$ training examples to each source. Fig.~\ref{fig:sp-10-diabetes-svm-beta-16-4} and \ref{fig:sp-20-creditcard-nb-banzhaf} use Beta Shapley ($\mathtt{Beta(16, 4)}$) and Data Banzhaf as prior semivalues respectively. More experimental results are included in Fig.~\ref{fig:appendix-contribution-to-model-performance-and-model-robustness}, where the details are described in the caption. Here we can observe that higher staying probabilities in general lead to higher DeRDaVa scores.

\begin{figure}
    \centering
    \begin{subfigure}[b]{0.24\textwidth}
        \centering
        \includegraphics[width=\textwidth]{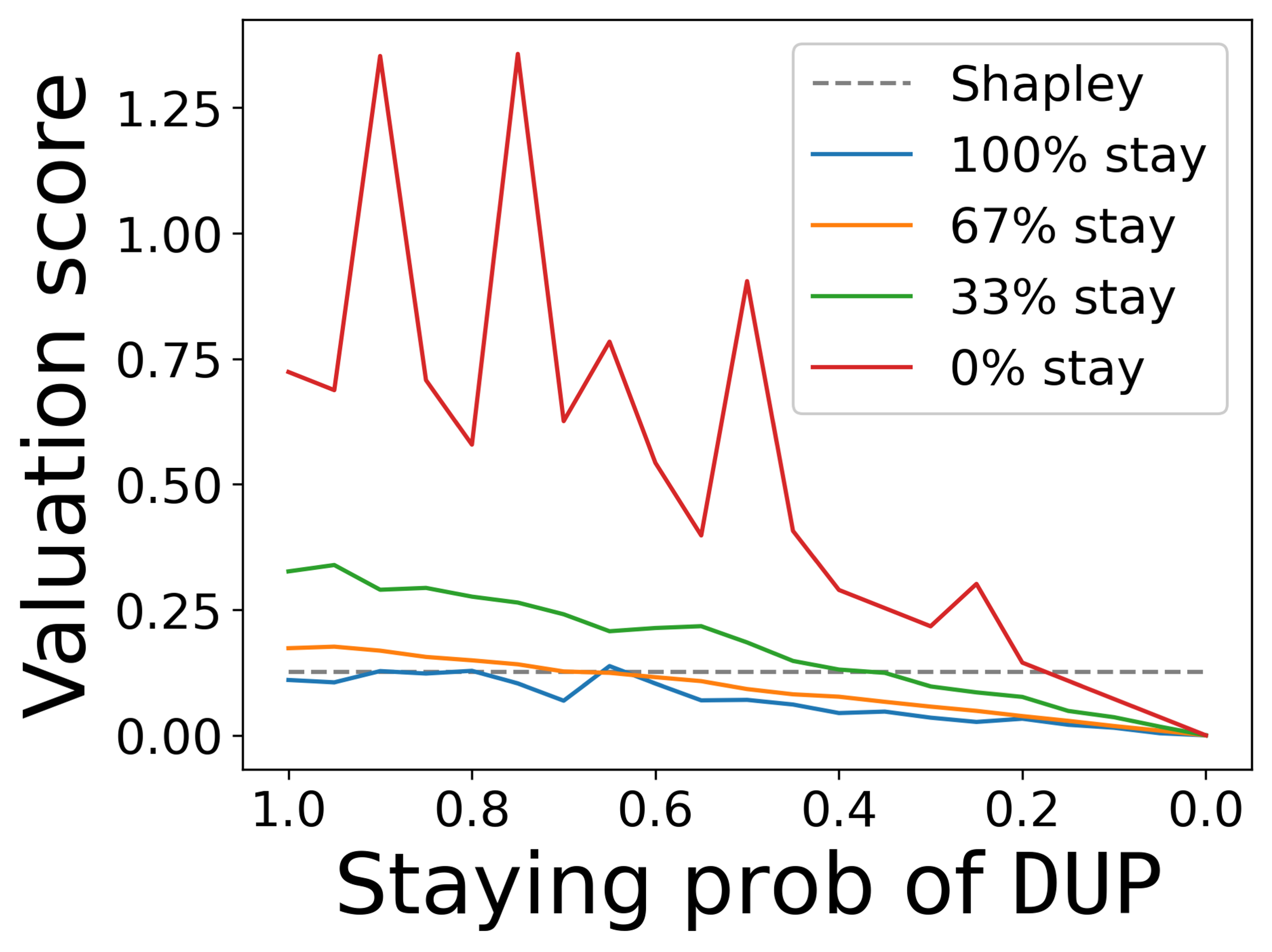}
        \caption{[LR-Pm].}\label{fig:ds-5-phoneme-lr-shapley}
    \end{subfigure}
    \begin{subfigure}[b]{0.24\textwidth}
        \centering
        \includegraphics[width=\textwidth]{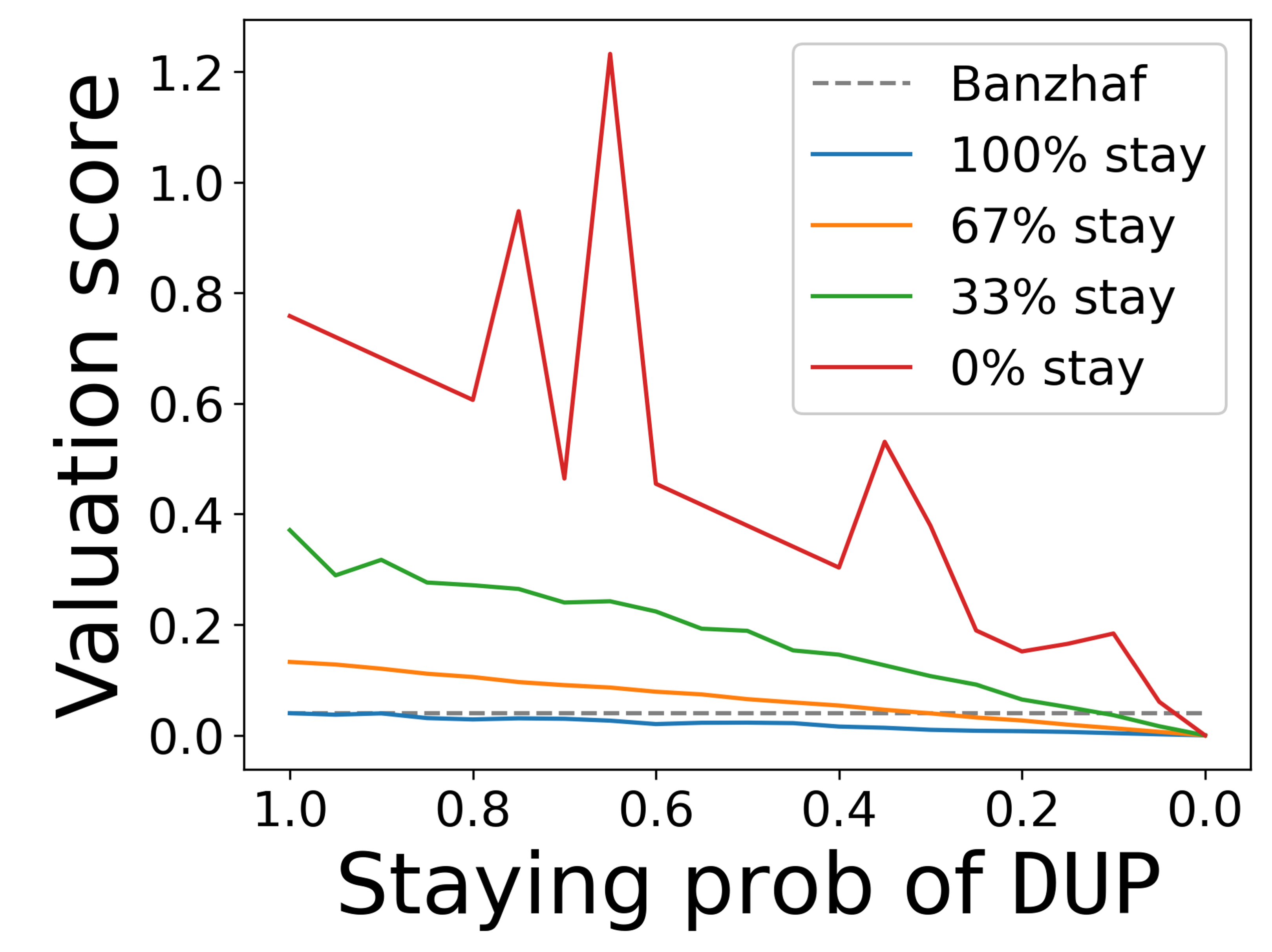}
        \caption{[NB-Wd].}\label{fig:ds-5-creditcard-svm-banzhaf}
    \end{subfigure}
    \begin{subfigure}[b]{0.24\textwidth}
        \centering
        \includegraphics[width=\textwidth]{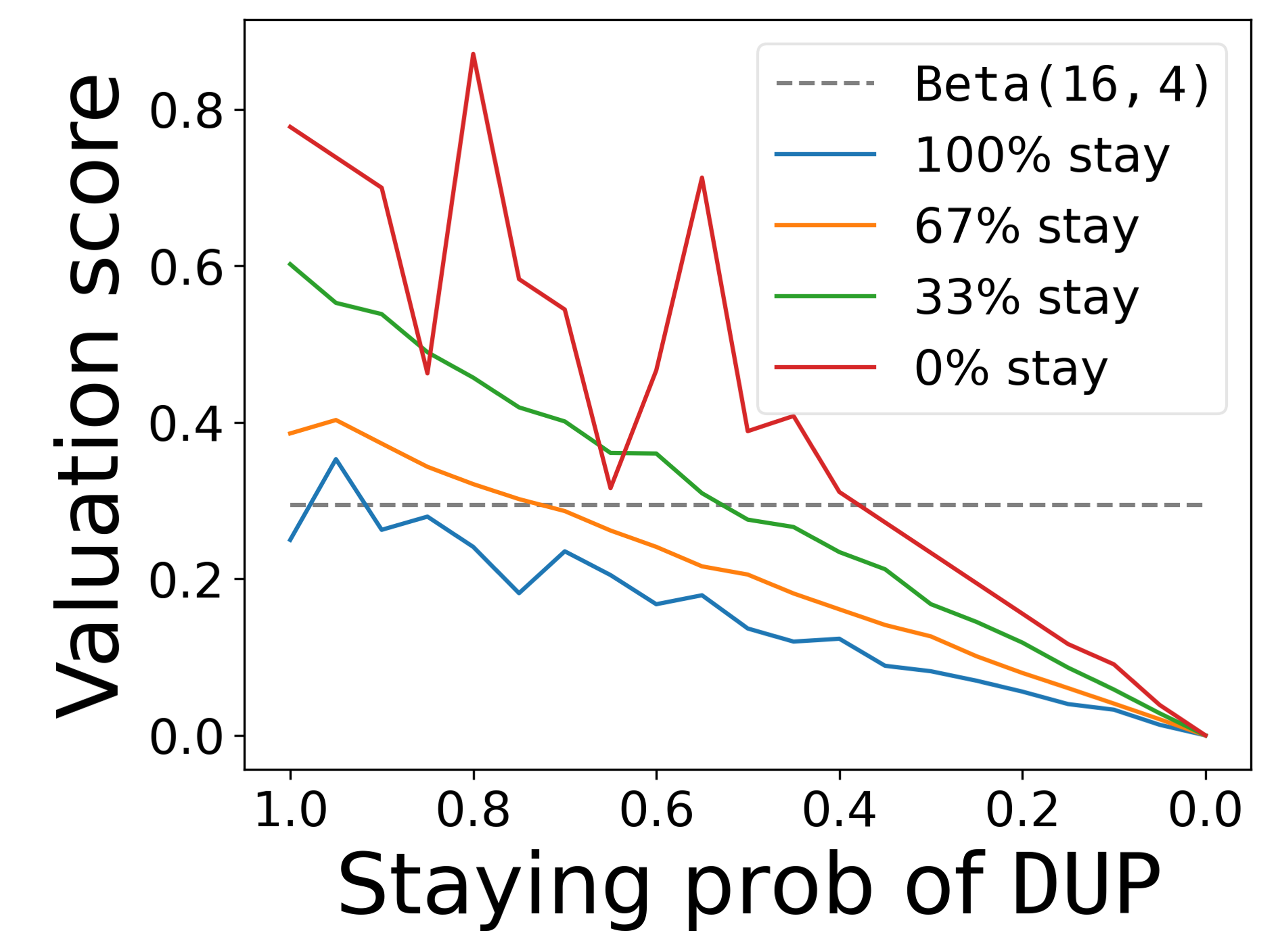}
        \caption{[SVM-Pm].}\label{fig:ds-5-wind-nb-beta-16-4}
    \end{subfigure}
    \begin{subfigure}[b]{0.24\textwidth}
        \centering
        \includegraphics[width=\textwidth]{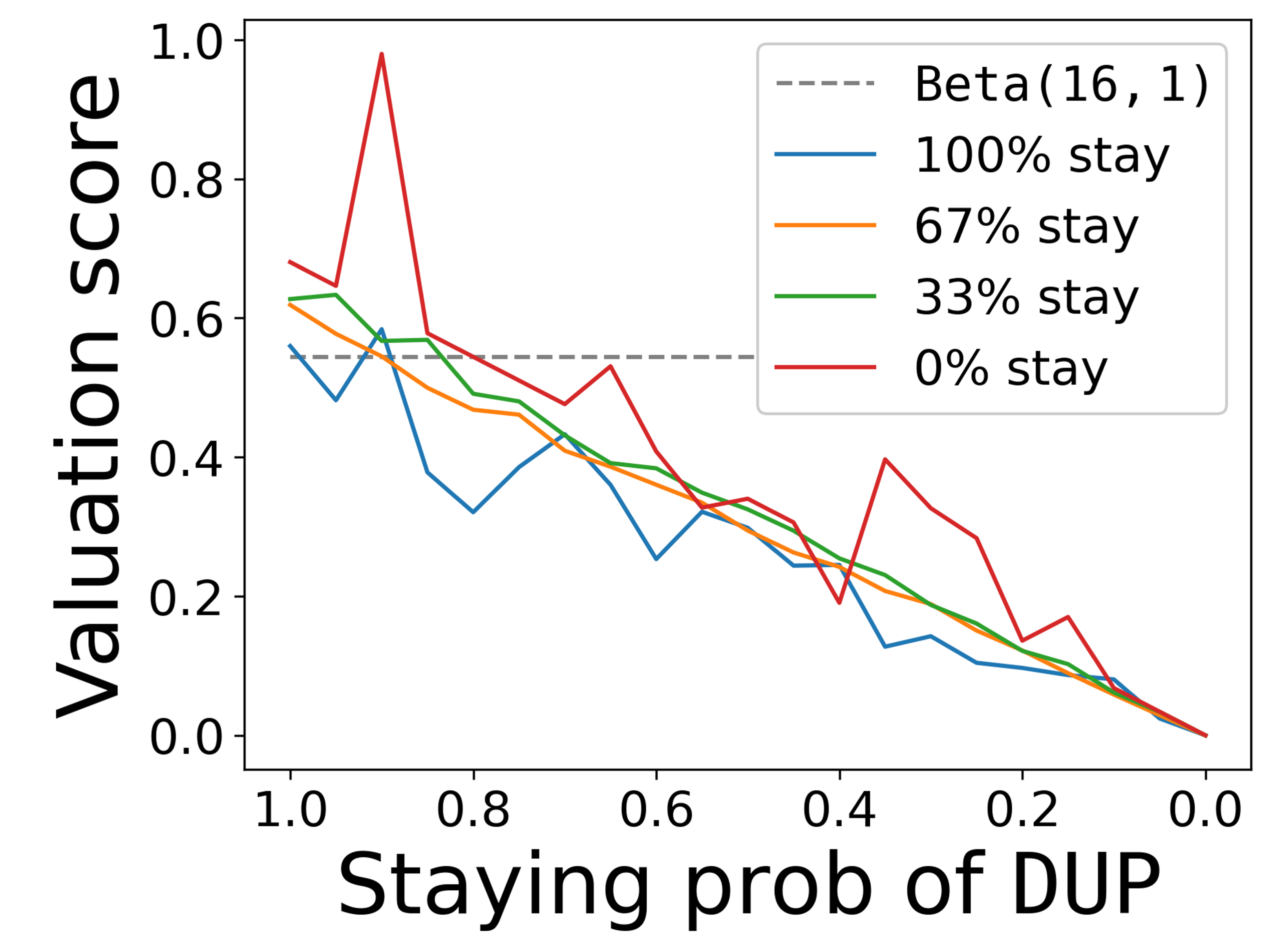}
        \caption{[RC-Po].}\label{fig:ds-5-pol-rc-beta-16-1}
    \end{subfigure}
    \caption{Additional experiments on data similarity. We measure the valuation score of the duplicated data source \texttt{DUP} when it has different staying probability, given that other data sources have $100\%$, $67\%$, $33\%$ and $0\%$ independent staying probability. (\ref{fig:ds-5-phoneme-lr-shapley}) uses Shapley prior; (\ref{fig:ds-5-creditcard-svm-banzhaf}) uses Banzhaf prior; (\ref{fig:ds-5-wind-nb-beta-16-4}) uses $\mathtt{Beta(16, 4)}$ prior; (\ref{fig:ds-5-pol-rc-beta-16-1}) uses $\mathtt{Beta(16, 1)}$ prior. Since $\mathtt{Beta(16, 1)}$ places a very large weight on smaller coalitions which are likely to remain even after data deletion, the deviations among the $4$ DeRDaVa lines with $\mathtt{Beta(16, 1)}$ prior is the smallest across all experimental runs.}
    \label{fig:appendix-data-similarity}
\end{figure}

\paragraph{Data Similarity} The amount of similarity or uniqueness of data sources is hard to quantify on real datasets. However, we try to replicate this experiment on real datasets by randomly splitting the whole dataset into $5$ smaller data sources, then adding an additional data source \texttt{DUP} that duplicates one of the existing data sources. We observe that the behaviours of DeRDaVa towards similar data sources which we discuss in the main paper still preserve (Fig.~\ref{fig:appendix-data-similarity}). Specifically, the data source \texttt{DUP} is given higher DeRDaVa scores than pre-deletion semivalue scores when its staying probability is high or when other data sources have low staying probabilities. This supports that DeRDaVa places higher importance on data that is more likely to stay than on its more deletable ``twins'', thus factoring in deletion-robustness.

\paragraph{Data Quality} 

\begin{figure}
    \centering
    \begin{subfigure}[b]{0.45\textwidth}
        \centering
        \includegraphics[width=\textwidth]{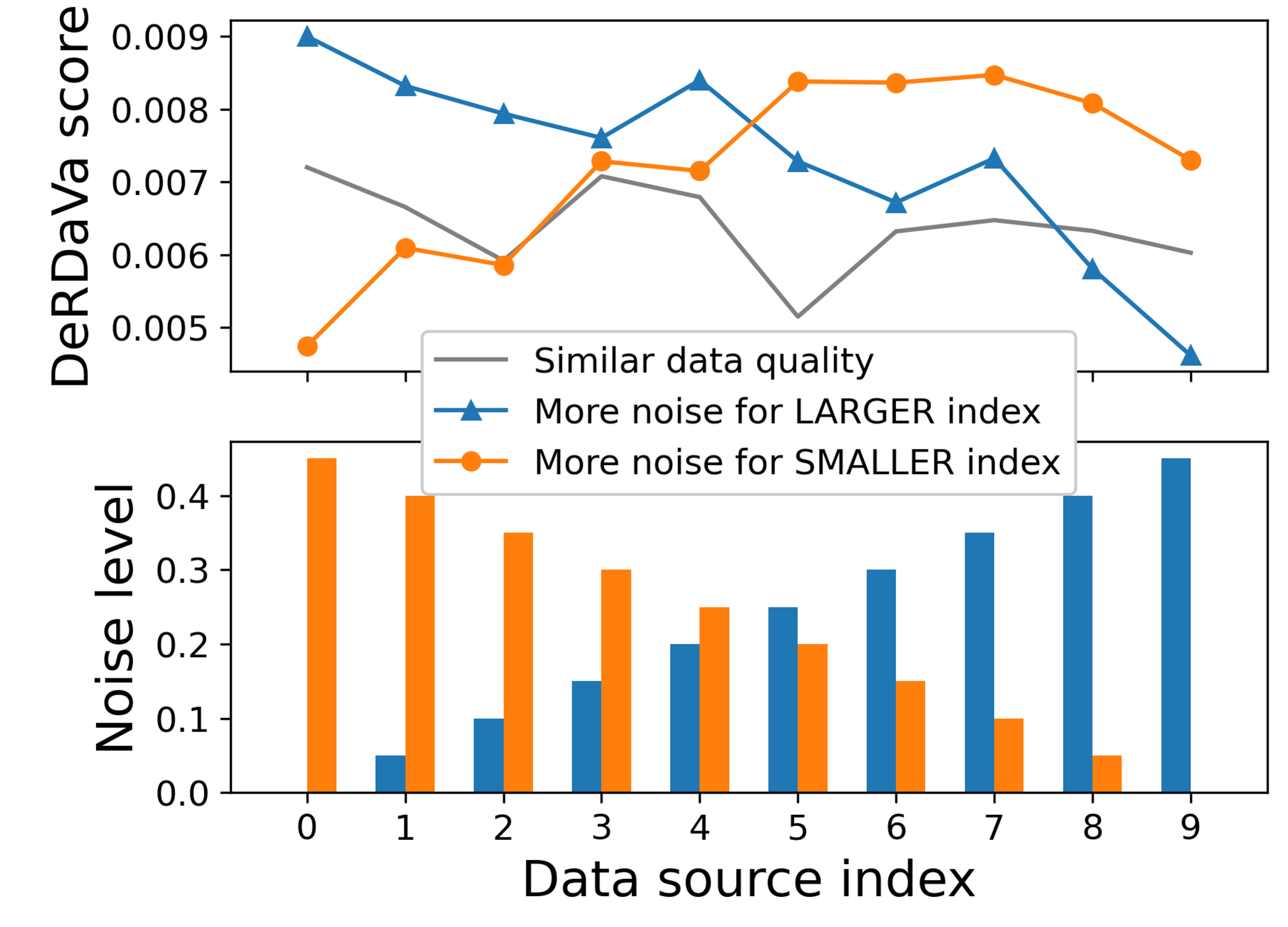}
        \caption{[LR-Pm].}\label{fig:dq-10-phoneme-lr-banzhaf}
    \end{subfigure}
    \begin{subfigure}[b]{0.45\textwidth}
        \centering
        \includegraphics[width=\textwidth]{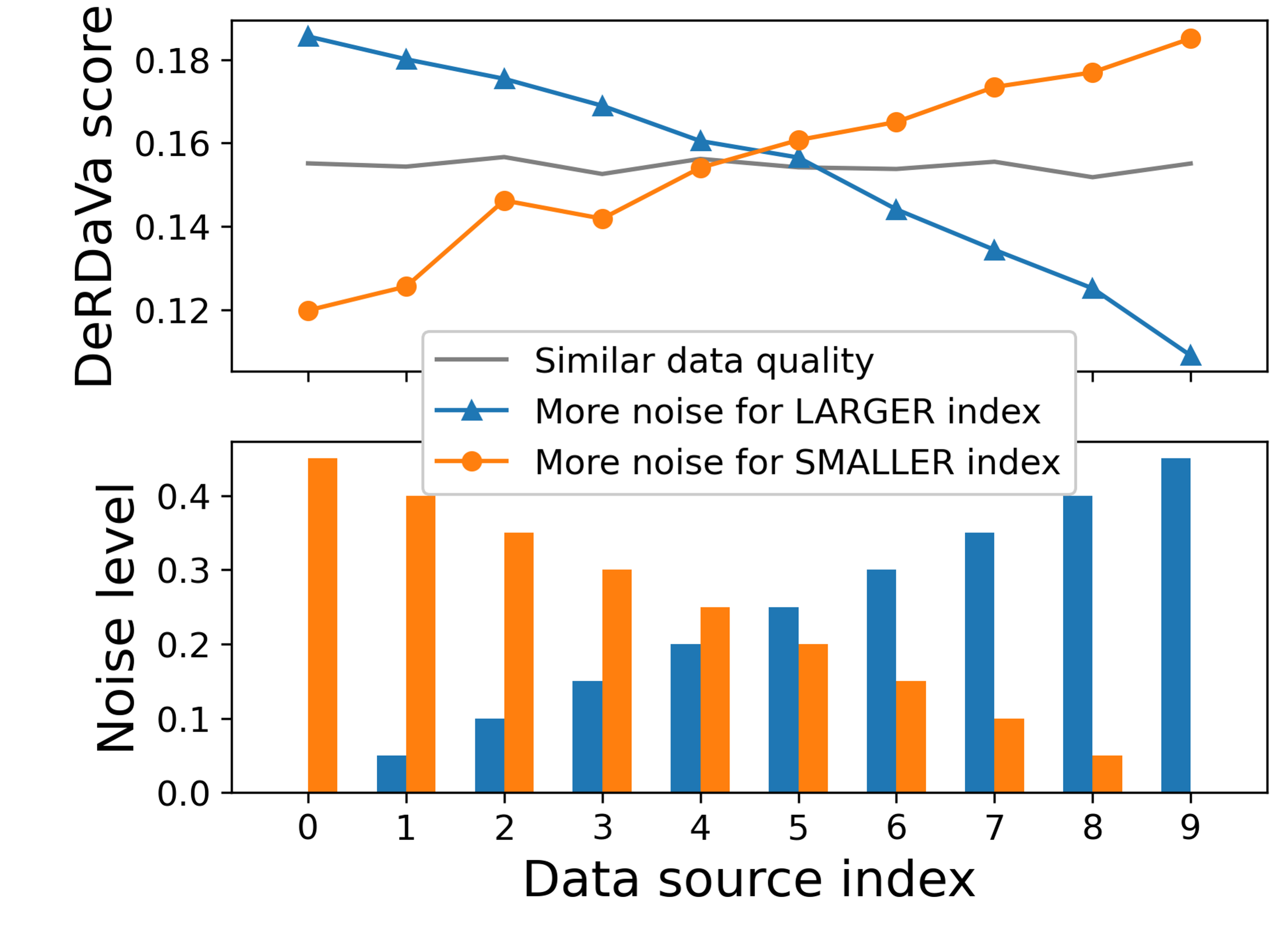}
        \caption{[SVM-Db].}\label{fig:dq-10-diabetes-svm-beta_16_4}
    \end{subfigure}
    \\
    \begin{subfigure}[b]{0.45\textwidth}
        \centering
        \includegraphics[width=\textwidth]{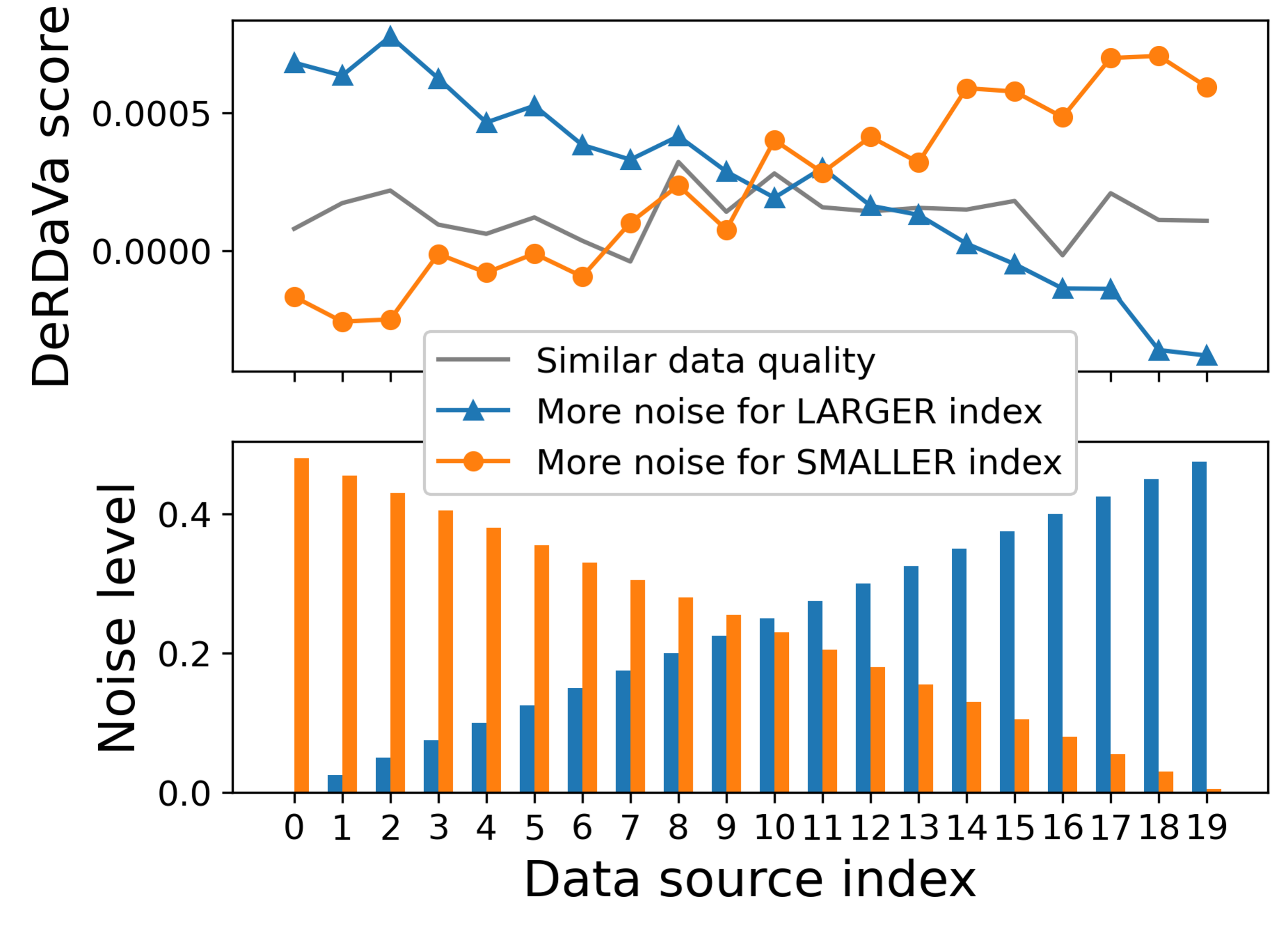}
        \caption{[NB-Wd].}\label{fig:dq-20-wind-nb-shapley}
    \end{subfigure}
    \begin{subfigure}[b]{0.45\textwidth}
        \centering
        \includegraphics[width=\textwidth]{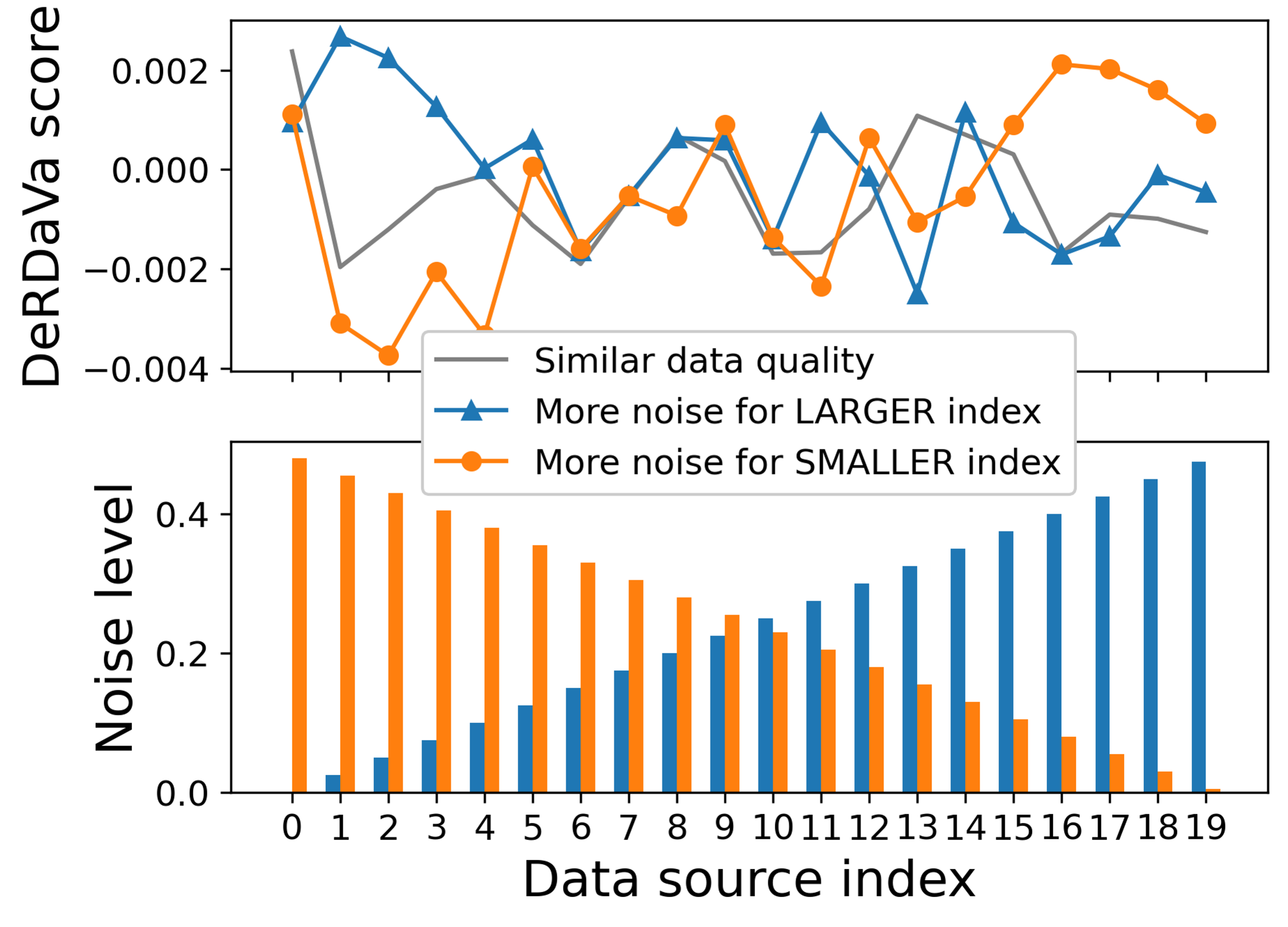}
        \caption{[NB-CC].}\label{fig:dq-20-creditcard-nb-beta_4_16}
    \end{subfigure}
    \caption{Experiments on data quality. (\ref{fig:dq-10-phoneme-lr-banzhaf}) uses $10$ data sources and Banzhaf prior; (\ref{fig:dq-10-diabetes-svm-beta_16_4}) uses $10$ data sources and $\mathtt{Beta(16, 4)}$ prior; (\ref{fig:dq-20-wind-nb-shapley}) uses $20$ data sources and Shapley prior; (\ref{fig:dq-20-creditcard-nb-beta_4_16}) uses $20$ data sources and $\mathtt{Beta(4, 16)}$ prior.}
    \label{fig:appendix-data-quality}
\end{figure}

In machine learning, data quality plays an important role in determining model performance. A data source with lower data quality contains higher proportion of noises, therefore making a small contribution to both model performance and deletion-robustness. We expect that DeRDaVa preserves the ability of semivalues to differentiate data sources with different data quality. In this experiment, we create data sources with varying data quality by first splitting a dataset into smaller subsets each representing a data source, and then adding different levels of synthetic noise to each data source. Moreover, we keep the staying probability of each data source constant at $0.9$ independently. The experimental results shown in Fig.~\ref{fig:appendix-data-quality} demonstrate that data sources with lower quality indeed receive lower DeRDaVa scores, and vice versa.

\subsection{Point Addition and Removal}

\begin{figure}
    \centering
    \begin{subfigure}[b]{0.24\textwidth}
        \centering
        \includegraphics[width=\textwidth]{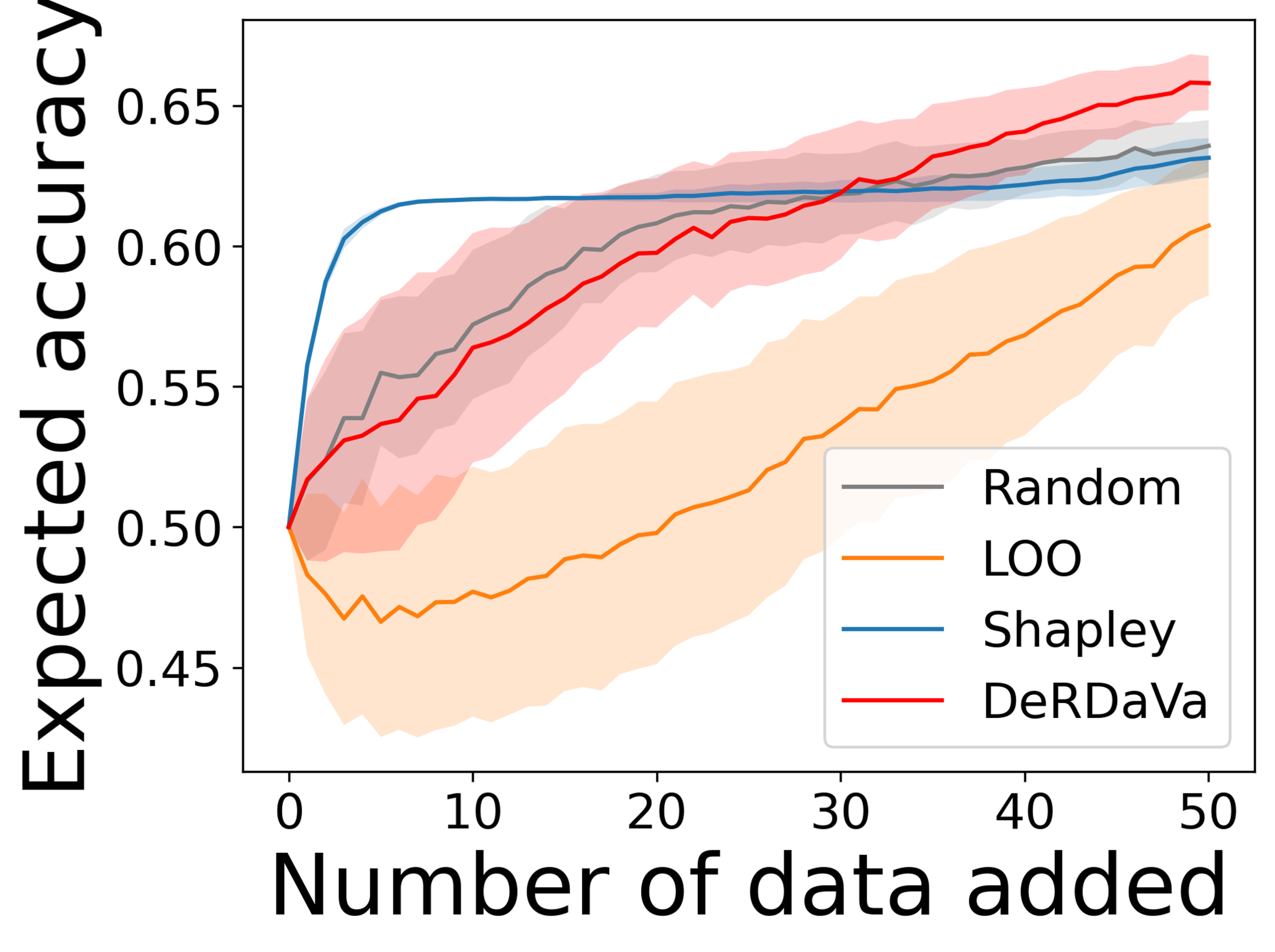}
        \caption{Add highest first.}\label{fig:par-add-highest}
    \end{subfigure}
    \begin{subfigure}[b]{0.24\textwidth}
        \centering
        \includegraphics[width=\textwidth]{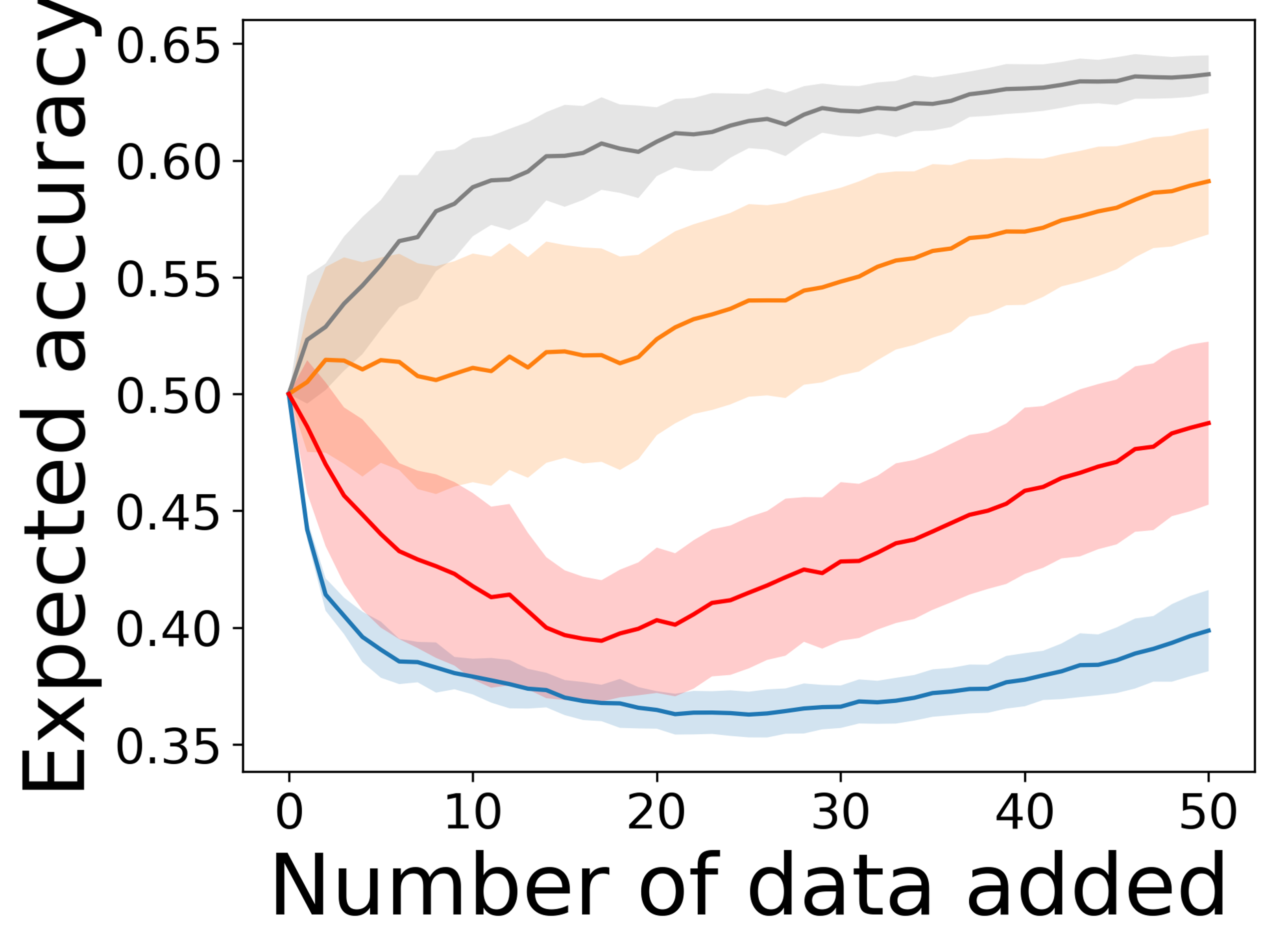}
        \caption{Add lowest first.}\label{fig:par-add-lowest}
    \end{subfigure}
    \begin{subfigure}[b]{0.24\textwidth}
        \centering
        \includegraphics[width=\textwidth]{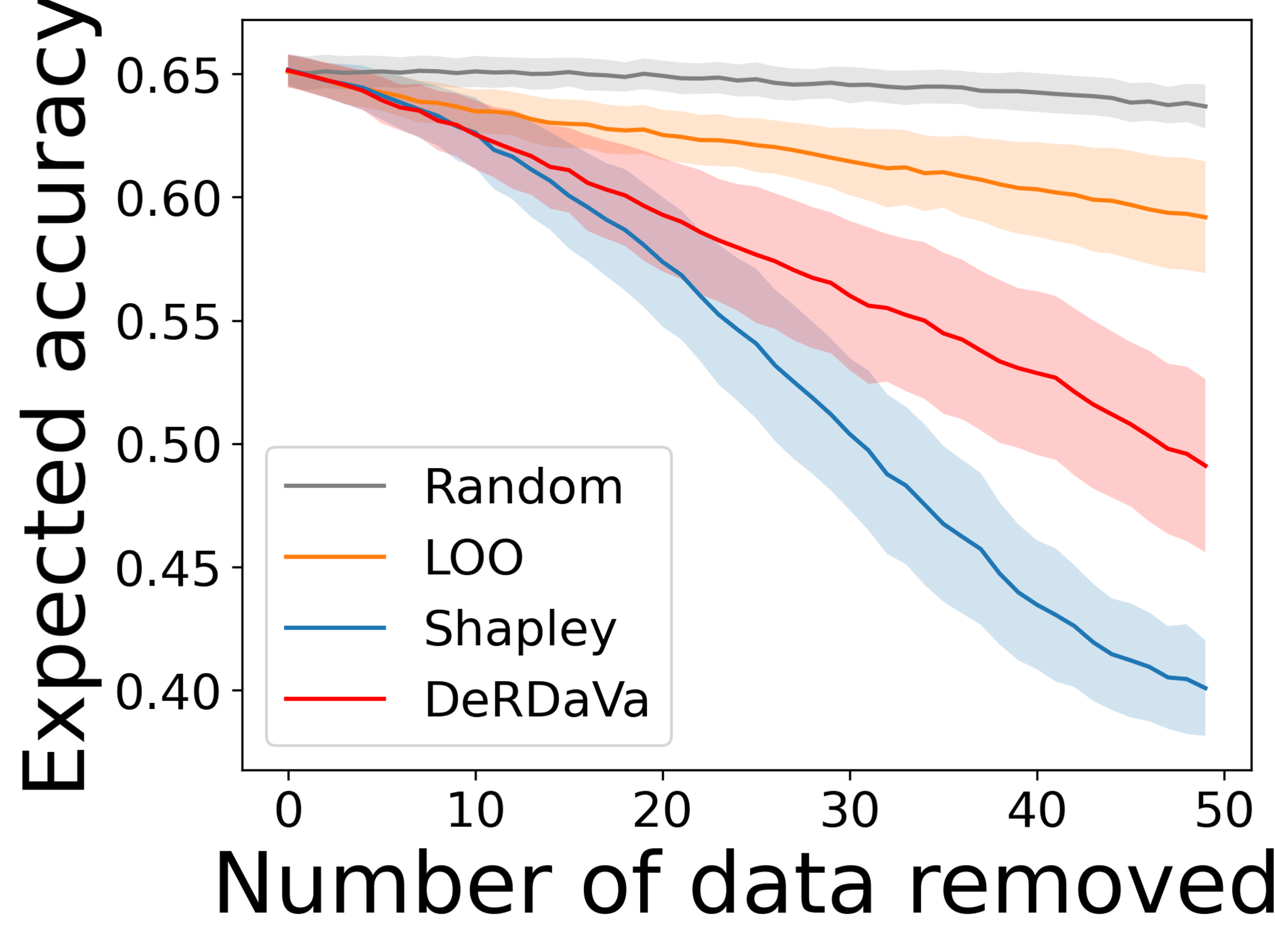}
        \caption{Remove highest first.}\label{fig:par-remove-highest}
    \end{subfigure}
    \begin{subfigure}[b]{0.24\textwidth}
        \centering
        \includegraphics[width=\textwidth]{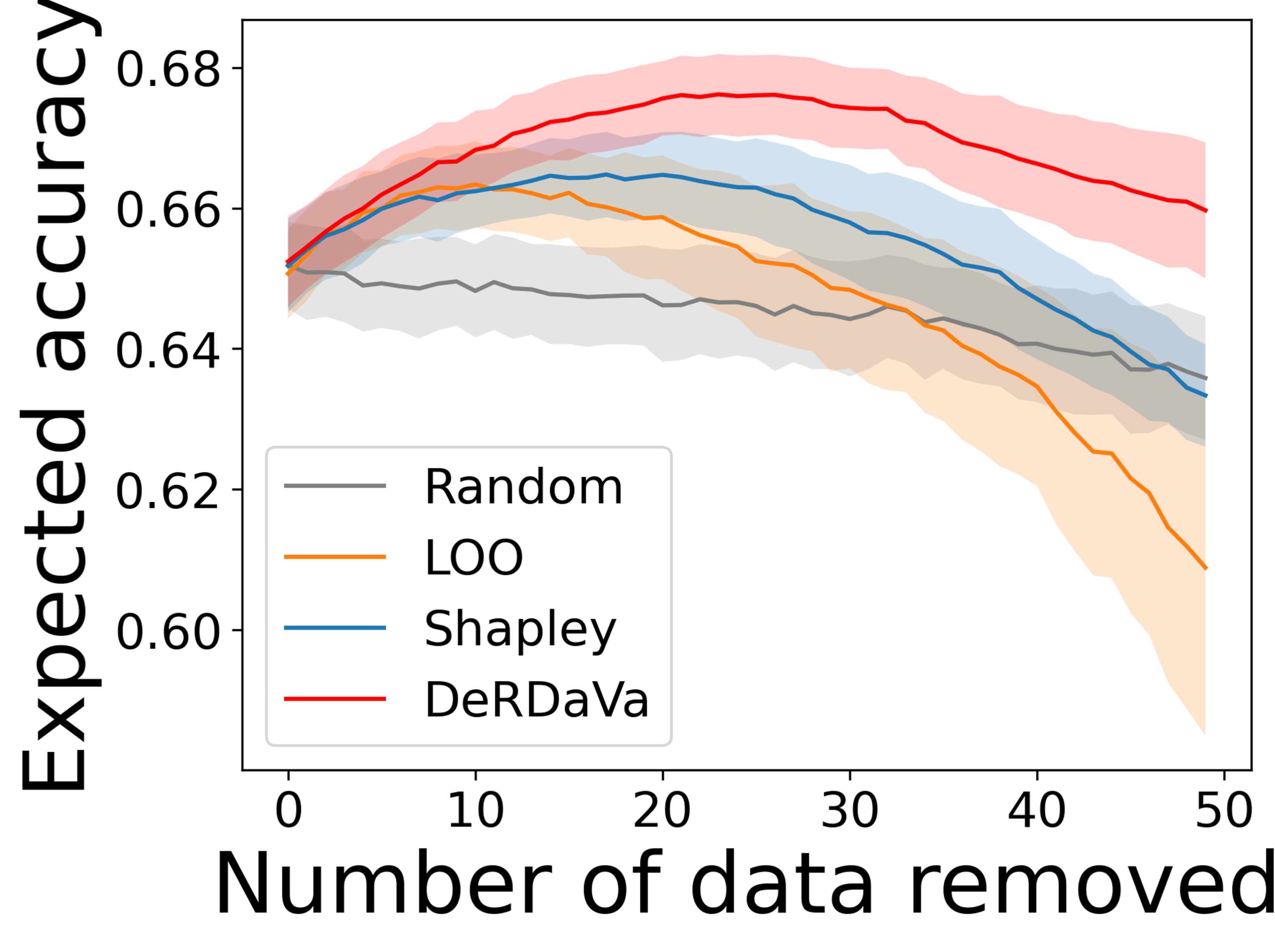}
        \caption{Remove lowest first.}\label{fig:par-remove-lowest}
    \end{subfigure}
    \\
    \begin{subfigure}[b]{0.24\textwidth}
        \centering
        \includegraphics[width=\textwidth]{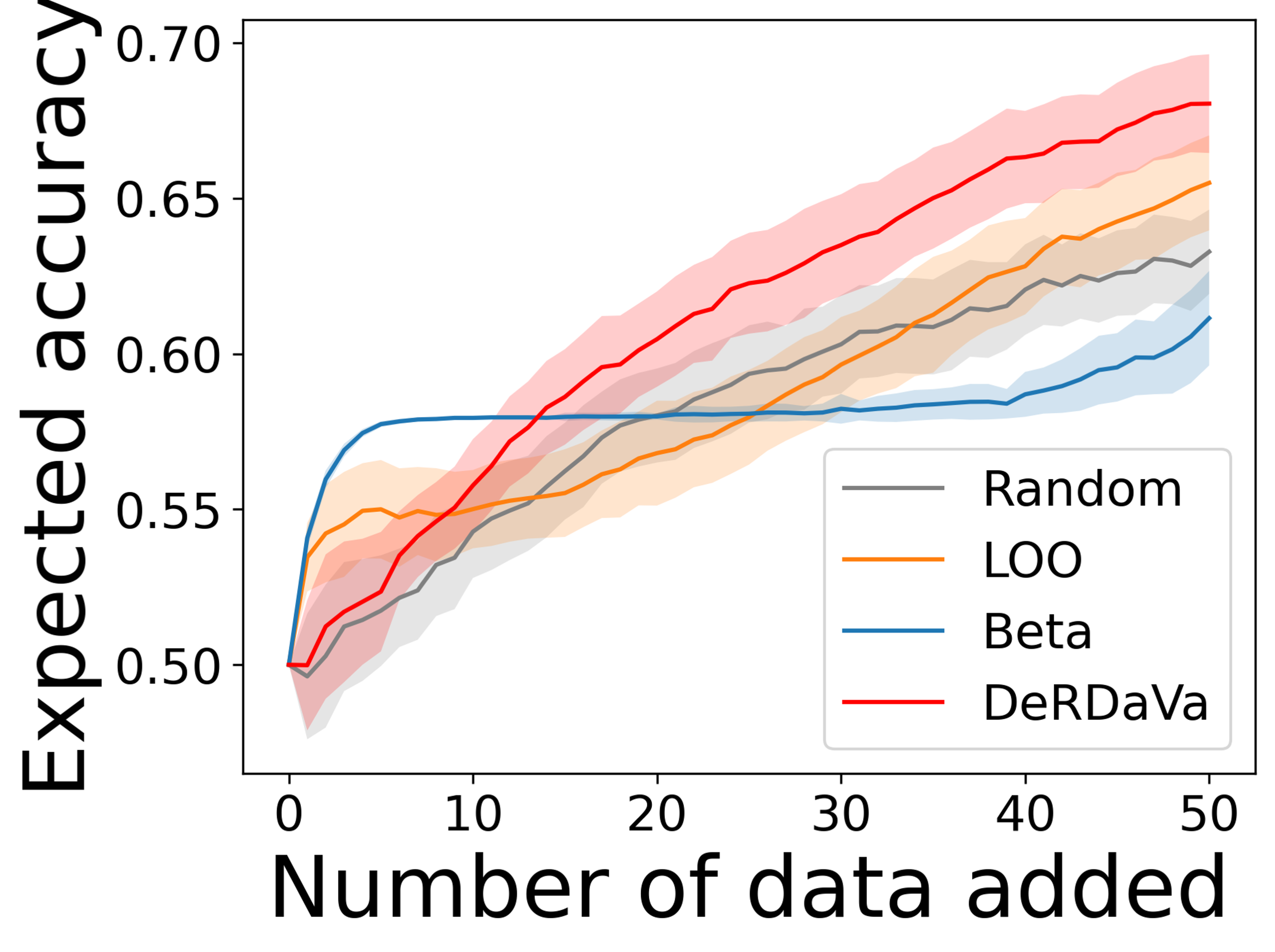}
        \caption{Add highest first.}\label{fig:par-add-highest-pol-rc-beta_16_1}
    \end{subfigure}
    \begin{subfigure}[b]{0.24\textwidth}
        \centering
        \includegraphics[width=\textwidth]{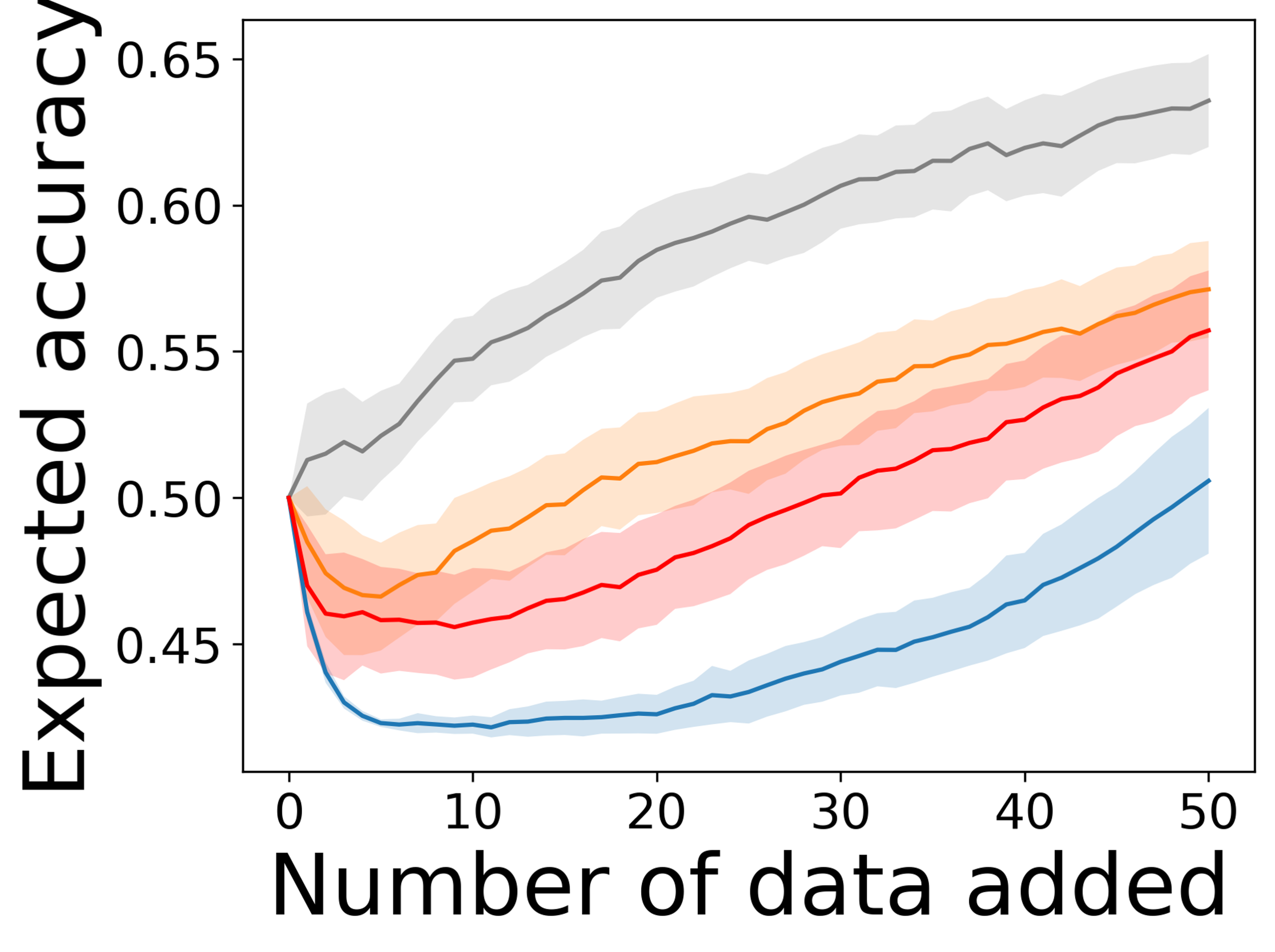}
        \caption{Add lowest first.}\label{fig:par-add-lowest-pol-rc-beta_16_1}
    \end{subfigure}
    \begin{subfigure}[b]{0.24\textwidth}
        \centering
        \includegraphics[width=\textwidth]{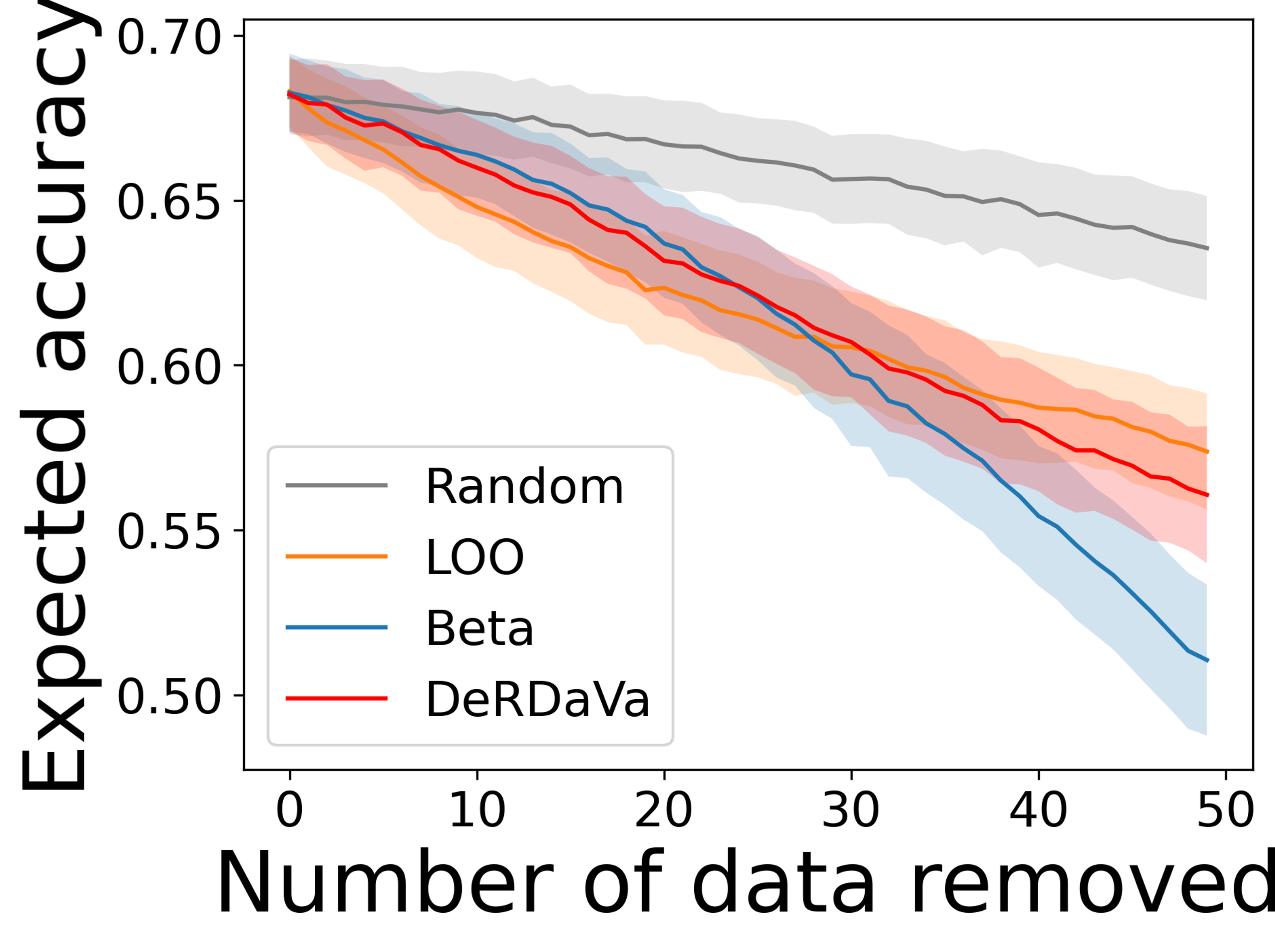}
        \caption{Remove highest first.}\label{fig:par-remove-highest-pol-rc-beta_16_1}
    \end{subfigure}
    \begin{subfigure}[b]{0.24\textwidth}
        \centering
        \includegraphics[width=\textwidth]{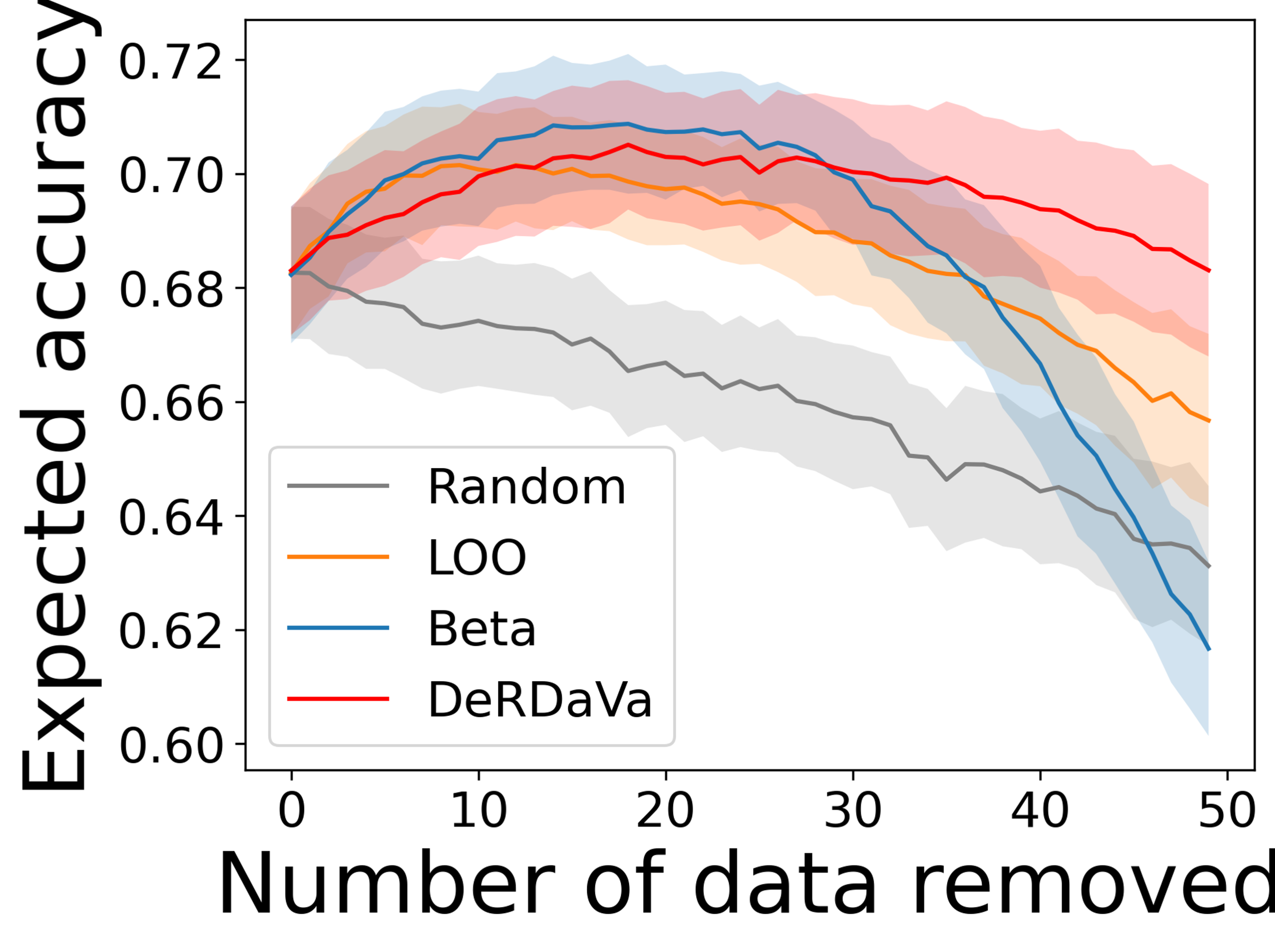}
        \caption{Remove lowest first.}\label{fig:par-remove-lowest-pol-rc-beta_16_1}
    \end{subfigure}
    \caption{Additional point addition and removal experiments. (\ref{fig:par-add-highest}) to (\ref{fig:par-remove-lowest}) use [LR-Pm] and Shapley prior; (\ref{fig:par-add-highest-pol-rc-beta_16_1}) to (\ref{fig:par-remove-lowest-pol-rc-beta_16_1}) use [RC-Po] and $\mathtt{Beta(16, 1)}$ prior.}
    \label{fig:appendix-point-addition-removal}
\end{figure}

The expected accuracy after each point addition or removal is measured by simulating the outcomes after data deletion for $50$ times and computing the means and variances. The results in Fig.~\ref{fig:appendix-point-addition-removal} show similar trends to the main paper. In general, when data with the highest scores are added first (\ref{fig:par-add-highest}, \ref{fig:par-add-highest-pol-rc-beta_16_1}), the increase in performance for DeRDaVa is relatively slower initially but higher when more data are added. When data with the lowest scores are added first (\ref{fig:par-add-lowest}, \ref{fig:par-add-lowest-pol-rc-beta_16_1}), DeRDaVa can identify noises at the beginning and reaches the turning point faster than Beta because Beta may penalize repetitiveness by giving lower scores. For Fig. \ref{fig:par-remove-lowest} and \ref{fig:par-remove-lowest-pol-rc-beta_16_1}, DeRDaVa exhibits rapid increase in performance initially and the slowest decrease later. This is because DeRDaVa assigns higher scores for data that contribute to deletion-robustness and these data are not removed first. This experiment further justifies the application of DeRDaVa in data interpretability.  

\subsection{Reflection of Long-Term Contribution}

\begin{figure}
    \centering
    \begin{subfigure}[b]{0.24\textwidth}
        \centering
        \includegraphics[width=\textwidth]{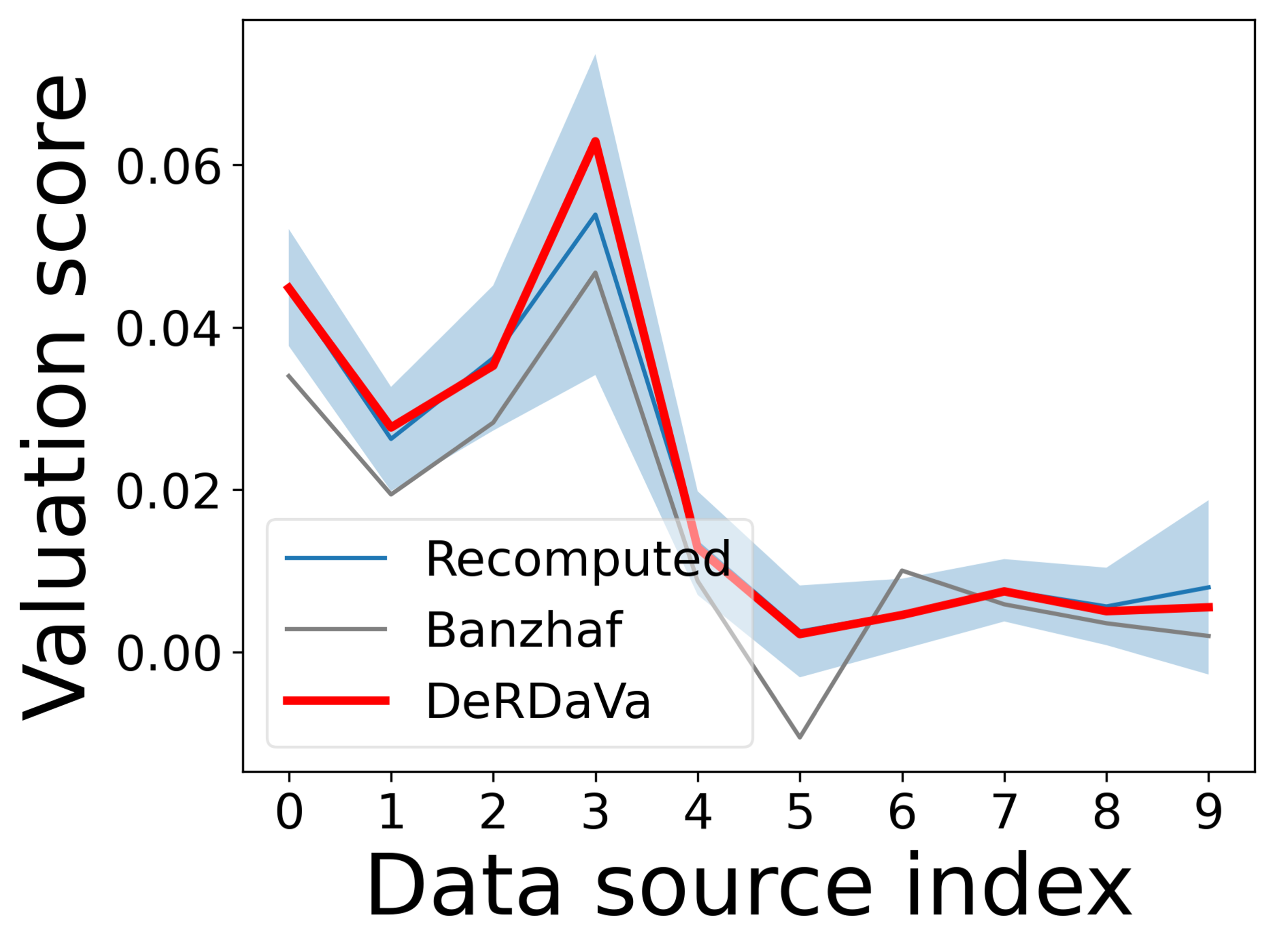}
        \caption{[SVM-CC].}\label{fig:dd-10-creditcard-svm-banzhaf-independent}
    \end{subfigure}
    \begin{subfigure}[b]{0.24\textwidth}
        \centering
        \includegraphics[width=\textwidth]{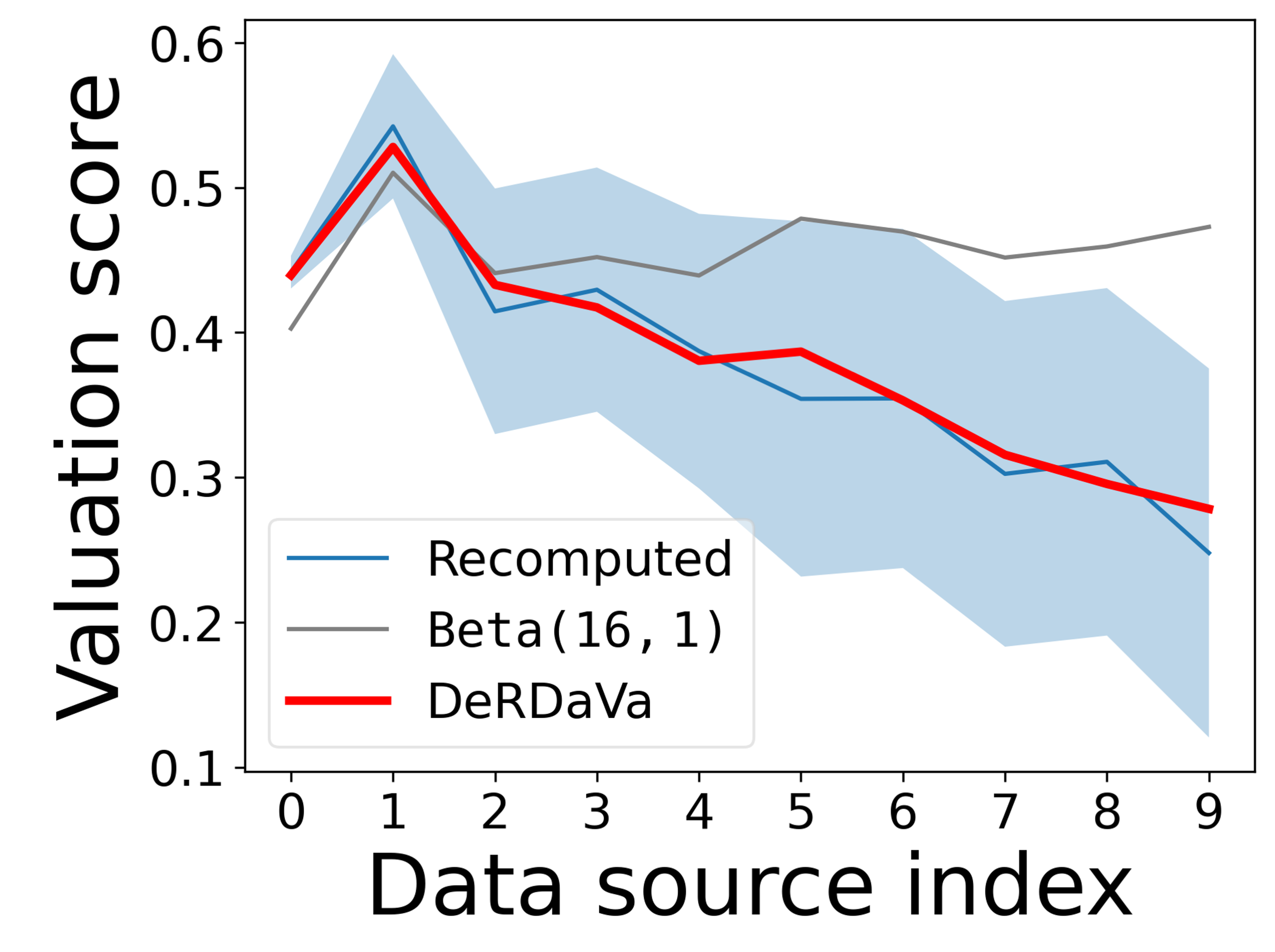}
        \caption{[RC-Po].}\label{fig:dd-10-pol-rc-beta_16_1_independent}
    \end{subfigure}
    \begin{subfigure}[b]{0.24\textwidth}
        \centering
        \includegraphics[width=\textwidth]{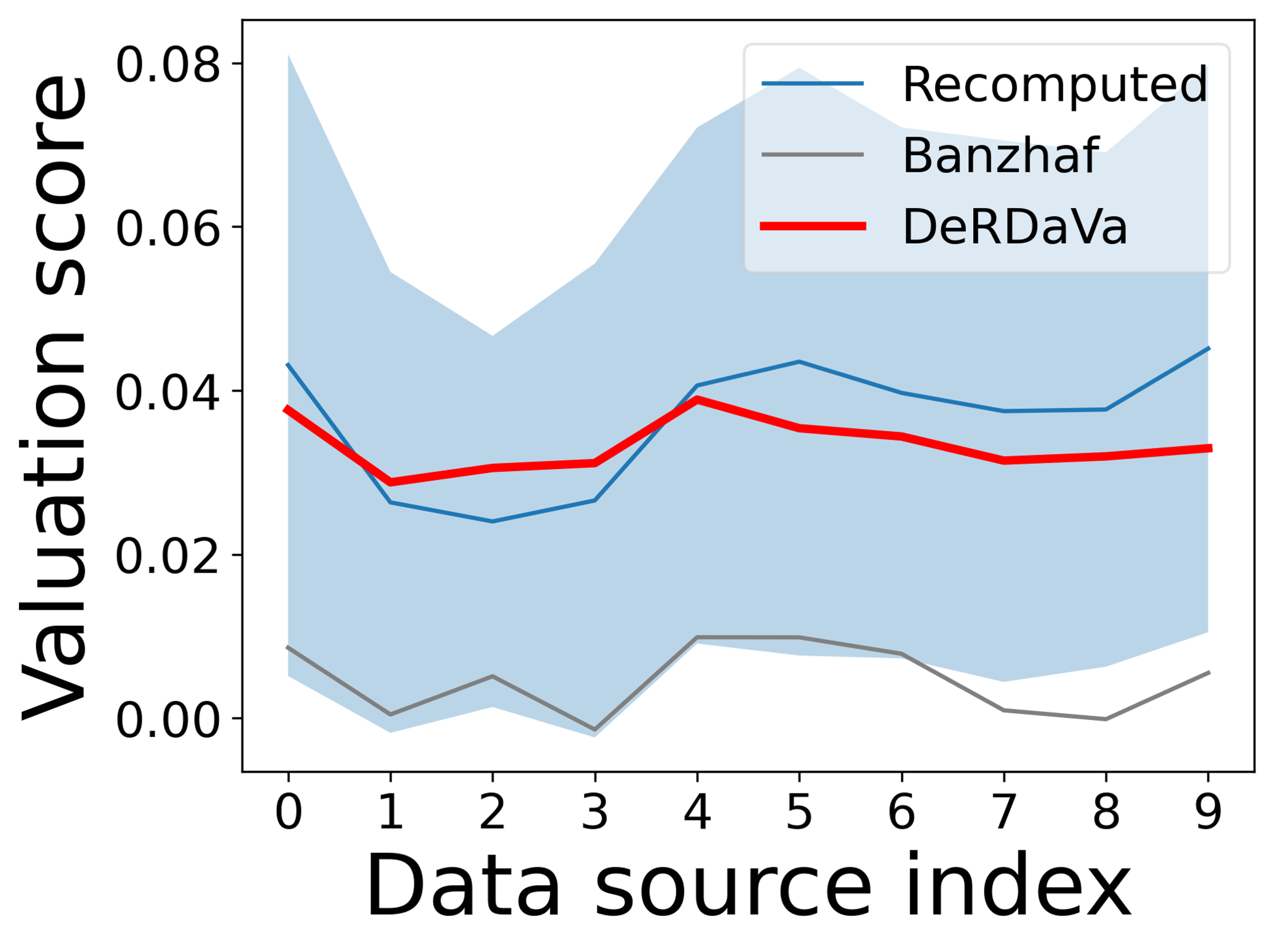}
        \caption{[SVM-CC].}\label{fig:dd-10-creditcard-svm-banzhaf_dependent}
    \end{subfigure}
    \begin{subfigure}[b]{0.24\textwidth}
        \centering
        \includegraphics[width=\textwidth]{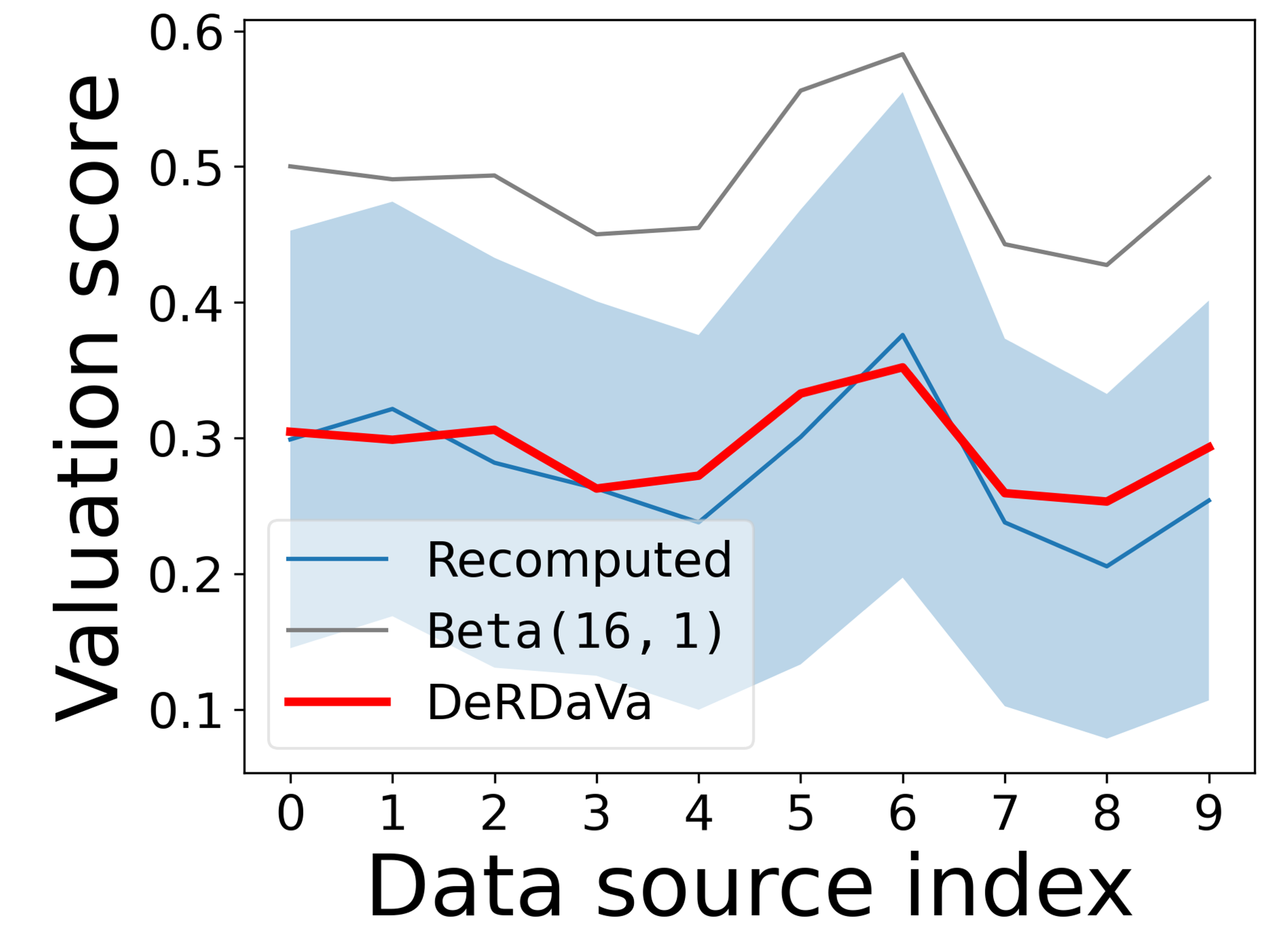}
        \caption{[RC-Po].}\label{fig:dd-10-pol-rc-beta_16_1_dependent}
    \end{subfigure}
    \caption{Additional experiments on reflection of long-term contribution. In (\ref{fig:dd-10-creditcard-svm-banzhaf-independent}) and (\ref{fig:dd-10-pol-rc-beta_16_1_independent}), data sources are assigned with independent staying probability which decreases with index; in (\ref{fig:dd-10-creditcard-svm-banzhaf_dependent}) and (\ref{fig:dd-10-pol-rc-beta_16_1_dependent}), we randomly generate the joint staying probability distribution such that data sources are dependent of each other.}
    \label{fig:appendix-dd}
\end{figure}

More experimental results are included in Fig.~\ref{fig:appendix-dd}. Particularly, we predetermine the staying probability distribution $P_{\mathbf{D}}$ and simulate the outcomes of data deletion for $100$ independent trials. We then recompute the semivalue scores for each trial and compare the average with DeRDaVa scores and semivalue scores. The trends shown are coherent with our analysis in the main paper: the average of recomputed semivalue scores approaches to the DeRDaVa score but deviates from the pre-deletion semivalue scores.

\subsection{Empirical Behaviours of Risk-DeRDaVa}

\begin{figure}
    \centering
    \begin{subfigure}[b]{0.24\textwidth}
        \centering
        \includegraphics[width=\textwidth]{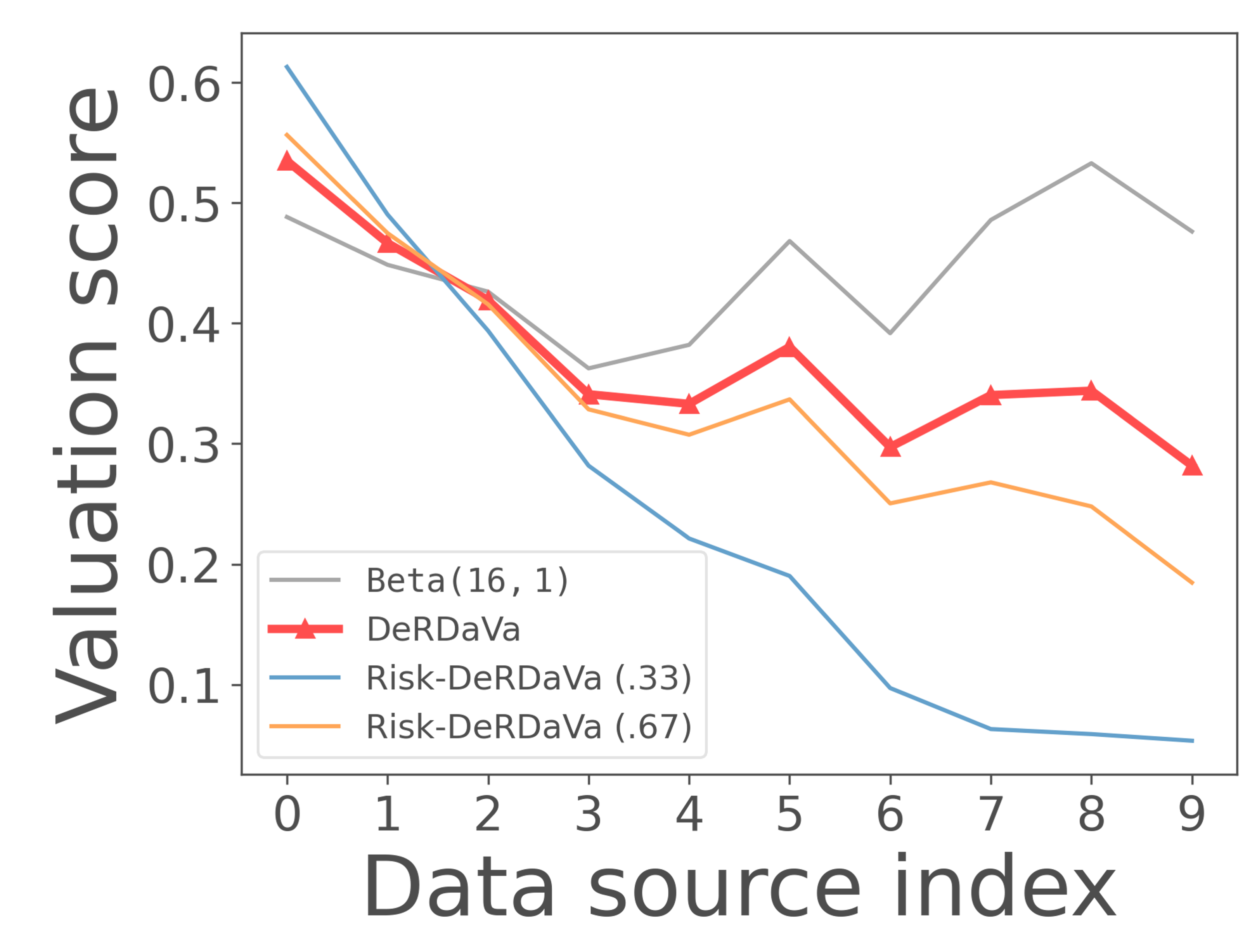}
        \caption{[SVM-CC].}\label{fig:rd-10-creditcard-svm-beta_16_1-averse}
    \end{subfigure}
    \begin{subfigure}[b]{0.24\textwidth}
        \centering
        \includegraphics[width=\textwidth]{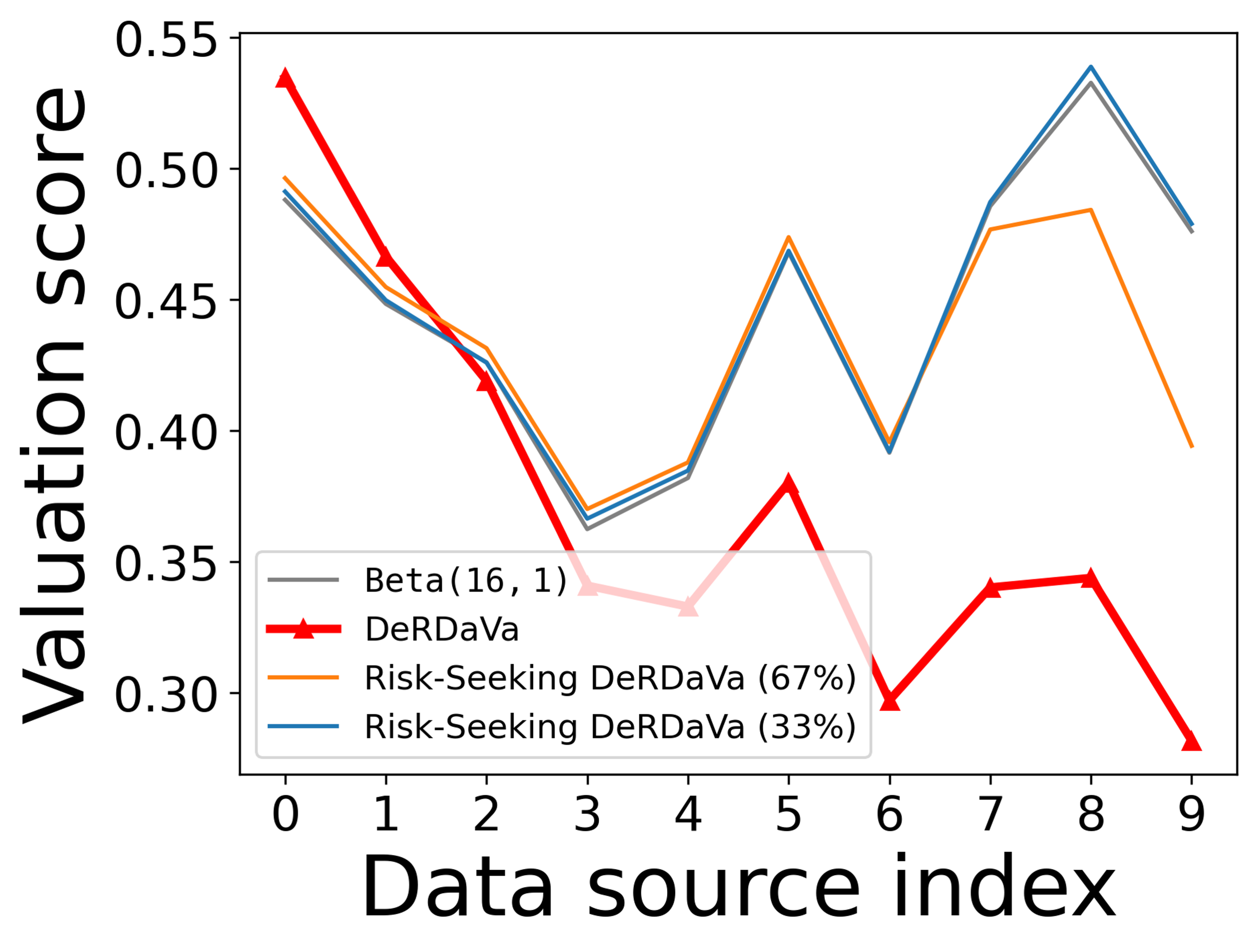}
        \caption{[SVM-CC].}\label{fig:rd-10-creditcard-svm-beta_16_1-seeking}
    \end{subfigure}
    \begin{subfigure}[b]{0.24\textwidth}
        \centering
        \includegraphics[width=\textwidth]{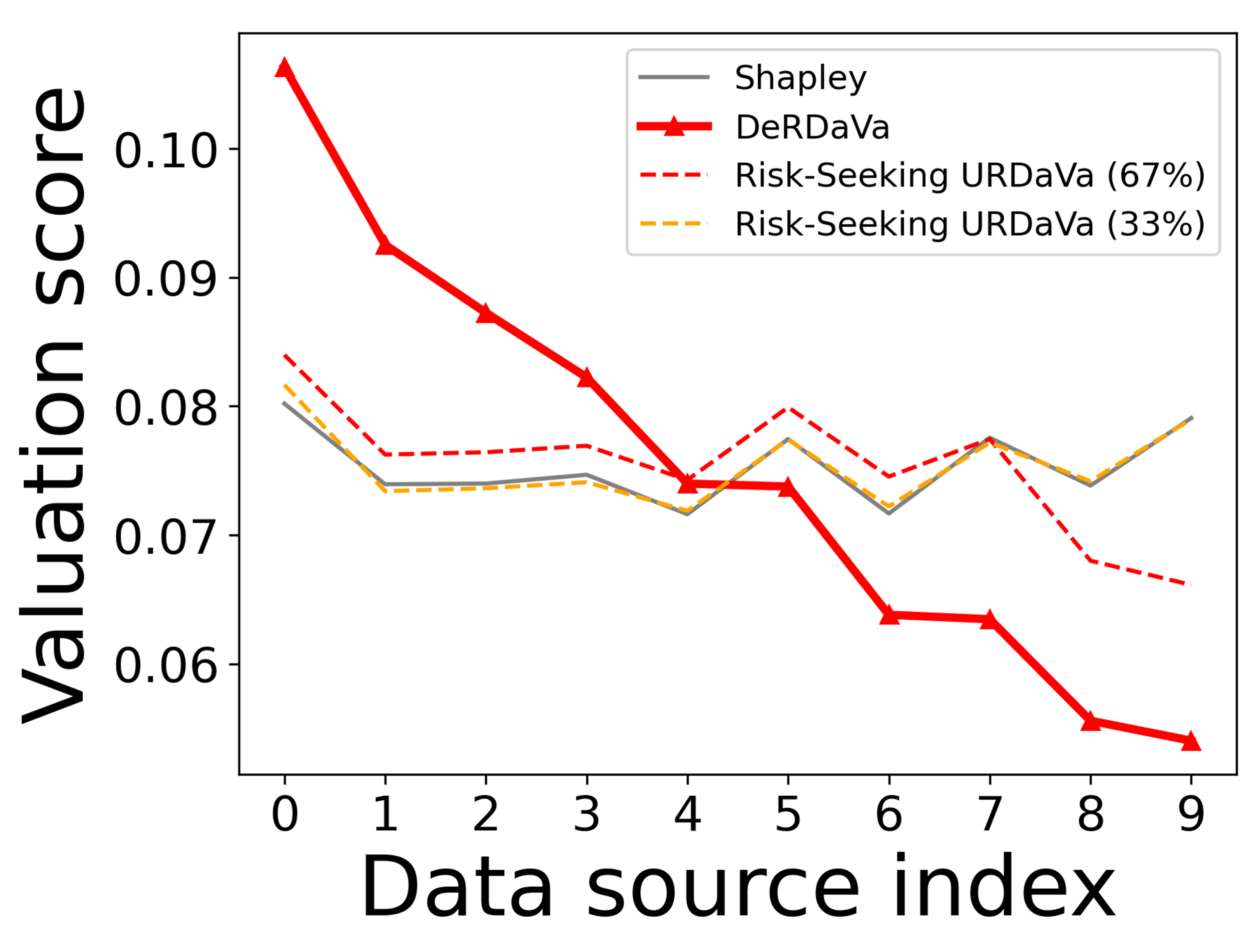}
        \caption{[LR-Pm].}\label{fig:rd-10-phoneme-lr-shapley-seeking}
    \end{subfigure}
    \begin{subfigure}[b]{0.24\textwidth}
        \centering
        \includegraphics[width=\textwidth]{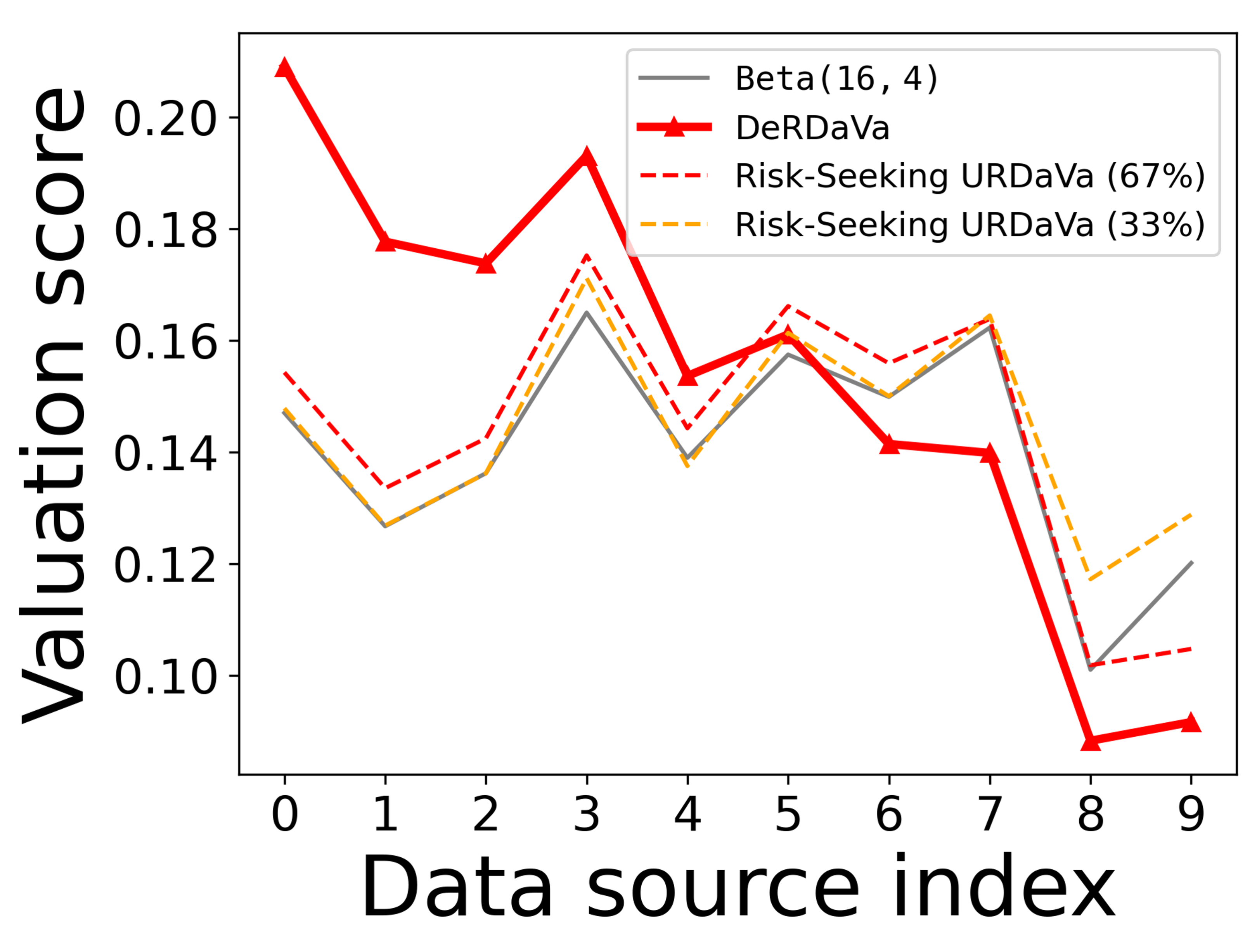}
        \caption{[NB-Db].}\label{fig:rd-10-diabetes-nb-beta_16_4-seeking}
    \end{subfigure}
    \caption{Additional experiments on Risk-DeRDaVa.}
    \label{fig:appendix-rd}
\end{figure}

Fig.~\ref{fig:rd-10-creditcard-svm-beta_16_1-averse} is done using similar procedures yet different dataset, model and prior semivalue from what is shown in the main paper. To better understand the behaviours of Risk-DeRDaVa, we also replicate the experiment for risk-seeking DeRDaVa using C-CVaR$^+$. Fig.~\ref{fig:rd-10-creditcard-svm-beta_16_1-seeking}, \ref{fig:rd-10-phoneme-lr-shapley-seeking} and \ref{fig:rd-10-diabetes-nb-beta_16_4-seeking} show the results. Converse to risk-averse DeRDaVa shown in the main paper, risk-seeking DeRDaVa penalizes data sources with a low staying probability by a smaller extent and the resulting valuation scores are closer to the pre-deletion semivalue scores.

\section{Other Discussions}

\subsection{Scaled Semivalue}\label{appendix:scaled-semivalue}

One may think of another way to account for deletion robustness in data valuation --- multiplying each data source $d_i$'s pre-deletion semivalue score $\phi^n_i(v)$ by its \textbf{individual} staying probability $\Pr[\mathbb{I}_{d_i} = 1]$. We call this \textit{scaled semivalue} (SS) score. For example, if two data sources $d_i$ and $d_j$ have pre-deletion Data Shapley scores $.6$ and $.4$ respectively and independent staying probabilities $.5$ and $.9$ respectively, then their SS scores are $.3$ and $.36$ respectively. 

\begin{figure}
    \centering
    \begin{subfigure}[b]{0.24\textwidth}
        \centering
        \includegraphics[width=\textwidth]{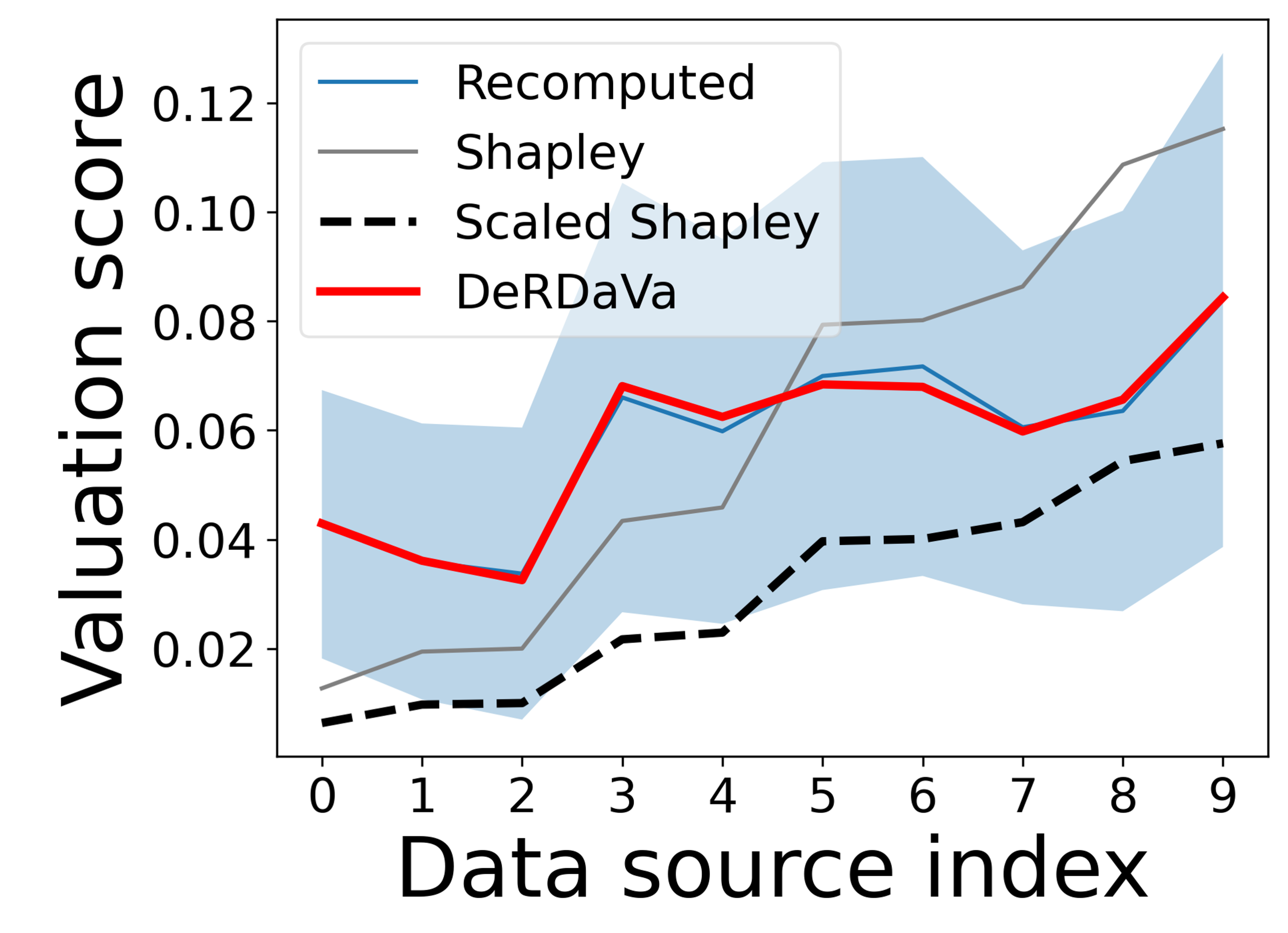}
        \caption{[RC-Db].}\label{fig:ext-diabetes-rc-shapley}
    \end{subfigure}
    \begin{subfigure}[b]{0.24\textwidth}
        \centering
        \includegraphics[width=\textwidth]{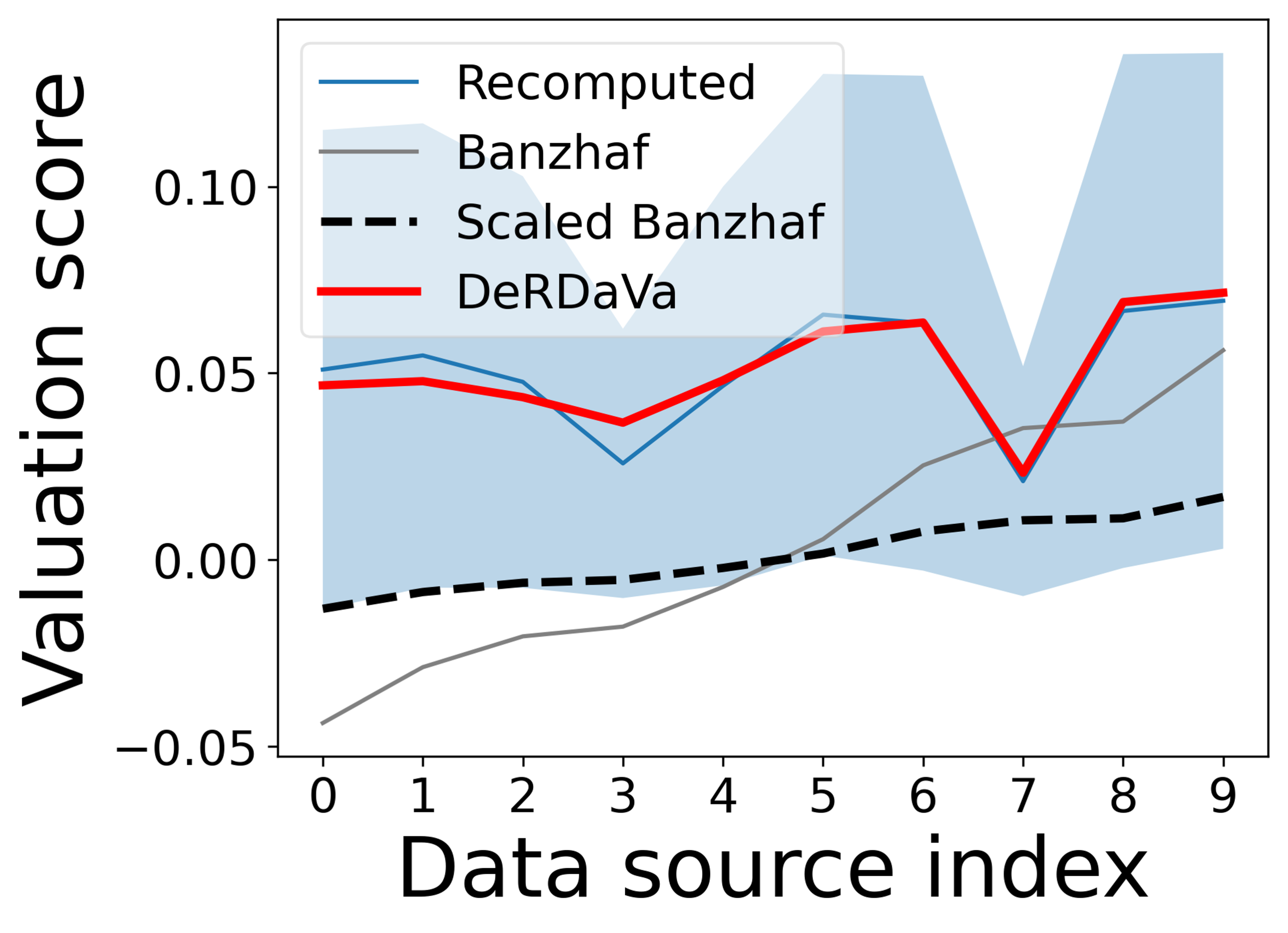}
        \caption{[LR-Pm].}\label{fig:ext-phoneme-lr-banzhaf}
    \end{subfigure}
    \caption{Comparison of DeRDaVa and scaled semivalue (SS). We repeat our Data Deletion Simulation experiments (Sec. 4.3). (\ref{fig:ext-diabetes-rc-shapley}) uses Shapley prior and assigns a $0.5$ independent staying probability to each data source; (\ref{fig:ext-phoneme-lr-banzhaf}) uses Banzhaf prior and assigns a $0.3$ independent staying probability to each data source. It can be seen that the orderings given by DeRDaVa and SS are different from each other. This is because SS does not account for the fact that the contribution of a data source changes when the support set becomes smaller. Instead, it only considers staying probability as a factor of deletion robustness.}
    \label{fig:derdava-vs-ss}
\end{figure}

Although SS can be computed more easily than DeRDaVa, it has several serious drawbacks. In our problem setting DeRDaVa has 4 major advantages over SS:

\begin{enumerate}
    \item SS does not work when each data source's staying probability depends on each other (i.e., the joint probability is not equal to the product of individual staying probability). For example, the model owner may find it more feasible to model the staying/quitting of data sources based on the number of remaining data sources (i.e., $P_\mathbf{D}(\mathbf{D} = D')$
 is a function of 
 $|D'|$).
\item SS fails to consider each data source's contribution to the smaller support set of remaining data sources (after deletion occurs), which can be different from their pre-deletion contribution. Hence it may not rank the data sources correctly. This can be observed in Fig. \ref{fig:derdava-vs-ss} --- SS ranking is merely a transformation of semivalues and differs from DeRDaVa. For example, in Fig. \ref{fig:ext-diabetes-rc-shapley}, despite its smaller Shapley value in the original set of data sources, DeRDaVa identifies that data source 3 is relatively more valuable to remaining data sources after deletions (as in the recomputed Shapley values).
\item The valuation scores given by SS are consistently lower than the original semivalues, and SS only penalizes data sources with low staying probabilities, but not rewards those with higher staying probabilities. However, when deletion actually occurs, the actual contribution of the staying data sources will increase (because model performance will increase more when there are fewer data sources, considering the training curve). Therefore, SS is undesirably undervaluing those with higher staying probabilities.
\item Fairness is essential to data valuation and collaborative machine learning, and the axiomatic approach provides a guideline/rubrics on what we should fulfil when designing data valuation techniques.
\end{enumerate}

\subsection{Discussions and Guidelines on Probability Distribution $P_{\mathbf{D}}$}\label{appendix:P_D}

As defined in Sec. \ref{subsec:random-cooperative-game-and-robustified-fairness-axioms}, $P_{\mathbf{D}}$ is the probability distribution of the random staying set $D'$. In the experiments, we set $P_\mathbf{D}$ by (a) assuming independent deletions across data sources and setting the independent staying probability of each data source, or (b) assuming dependent deletions and setting the probability of each case.

To complement Sec.~\ref{subsec:random-cooperative-game-and-robustified-fairness-axioms}, we provide some further guidelines for model owners to set $P_\mathbf{D}$ below.

\paragraph{Independent Deletions} In real life, the independent staying probability of each data source can be defined by surveying each data source or from past ``credit'' or ``reputation'' score: When $P_\mathbf{D}$ is interpreted as a ``credit'' score, data sources must maintain their credibility by keeping their quitting rate low, except for valid reasons such as a known privacy leakage accident. 
This approach protects data owners' right to be forgotten and model owners from malicious or abusive data deletions. On the other hand, data deletion can also occur in the future when data audits identify malicious data points.
To proactively account for future audits when performing data valuation, model owners can use DeRDaVa and interpret $P_\mathbf{D}$ as the ``reputation'' score of each data source and the likelihood that a source's data would not be deemed malicious in a future audit.

Note that although our staying probability is a value (e.g., $.5$), it still tolerates some uncertainty and misspecification as we do not assume a data source will stay/quit with certainty (unlike some of previous data valuation works that assume this information \textit{a priori}). Using a slightly misspecified $P_\mathbf{D}$ is better than assuming all data sources will stay with certainty.

Furthermore, the model owner can encode the uncertainty by using a distribution for the staying probability instead (e.g., $\mathtt{Beta(4, 4)}$ encodes an expected staying probability of $.5$). Then, $P_\mathbf{D}$ takes on a more complex form --- it is the product of the Beta and Bernoulli distribution. Sampling from $P_\mathbf{D}$ involves sampling the staying probability of each data source before sampling whether it stays/quits.

\paragraph{Dependent Deletions} In real life, the model owner may want to choose the weights on the different number of deletions instead, i.e., set $P_\mathbf{D}(\mathbf{D} = D')$ based on the size of the staying set $D'$. As the probabilities must sum to $1$, data owners' decisions to stay/quit become dependent events.

\subsection{Limitations and Social Impacts of DeRDaVa}\label{appendix:limitations-and-social-impacts}

From Sec. \ref{subsec:urdava-and-its-efficient-approximation} and App.~\ref{appendix:efficient-approximation-of-derdava}, 012-MCMC algorithm needs a larger bound on the number of samples than plain Monte-Carlo sampling to ensure that the absolute error in the DeRDaVa values exceeds $\epsilon$ on less than $\delta$ of the time. This occurs only when the proposed uniform distribution is a poor estimate for $P_{\mathbf{D}}$.
However, this limitation is justifiable --- it may be impossible for the model owners to sample from $P_{\mathbf{D}}$ if they cannot afford to compute $P_{\mathbf{D}}$ beforehand for all possible coalition $D'$. Using 012-sampling would enable the model owner to only query the auditor (e.g., the auditor audits coalitions to detect collusion) or survey the data owners (e.g., surveys a group of data sources together) upon sampling and get a scaled version of $P_{\mathbf{D}}(\mathbf{D} = D')$. 

Next, we discuss that the model owner may have difficulties setting $P_{\mathbf{D}}$ extensively in Sec.~\ref{subsec:random-cooperative-game-and-robustified-fairness-axioms}. We discuss how to address this limitation in App.~\ref{appendix:P_D}. This limitation is justifiable as DeRDaVa does not assume a data source will stay/quit with certainty (unlike some of previous data valuation works that assume this information \textit{a priori}) and can tolerate some uncertainty and misspecification. Using a slightly misspecified $P_\mathbf{D}$ is better than assuming all data sources will stay with certainty.

Another potential societal concern is that data sources may misreport their staying probability in order to groundlessly gain a higher valuation score and reward. This truthfulness concern is shared by existing works (e.g., \citet{sim2020collaborative} relies on data owners truthfully reporting the location of data points they collected). Such concerns might be mitigated by existing works on incentivizing truthfulness \citep{estornell2021incentivizing}. 

Lastly, model owners may also purposely set relatively low staying probabilities for data sources, to prevent them from quitting (hence threatening their \textit{right to be forgotten}), or to distribute less rewards.
To address this, a third-party inspection authority is necessary such that there is no conflict of interest with either model owner or data sources.

\end{document}